\let\mypdfximage\pdfximage
\def\pdfximage{\immediate\mypdfximage}

\documentclass[journal,onecolumn,10pt]{IEEEtran1}

\makeatletter
\def\markboth#1#2{\def\leftmark{\@IEEEcompsoconly{\sffamily}\MakeUppercase{\protect#1}}%
\def\rightmark{\@IEEEcompsoconly{\sffamily}\MakeUppercase{\protect#2}}}
\makeatother

\let\labelindent\relax

\usepackage{times}
\usepackage{marvosym}
\usepackage{longtable}
\usepackage{graphics}
\usepackage{graphicx}
\usepackage{caption}
\usepackage[english]{babel}
\usepackage{epsfig}
\usepackage{color}
\usepackage{multirow}
\usepackage{mathrsfs}
\usepackage{textcomp}
\usepackage{amsfonts}
\usepackage{amsmath}
\usepackage{amssymb}
\usepackage{nccmath}
\usepackage[numbers,square,sort&compress,comma]{natbib}
\usepackage{chapterbib}
\usepackage[ruled,vlined,linesnumbered]{algorithm2e}
\usepackage{setspace}
\usepackage{verbatim}
\usepackage{wrapfig}
\usepackage{dblfloatfix}
\usepackage{comment}
\usepackage[strict]{changepage}
\usepackage{ragged2e}
\usepackage{microtype}
\usepackage{enumitem}
\usepackage{nccmath}
\usepackage{hyperref}
\usepackage{doi}
\usepackage{afterpage}
\usepackage{multirow,bigdelim}
\usepackage{booktabs}
\usepackage{tikz}
\usepackage{color}
\usepackage[outline]{contour}
\usepackage{xcolor}
\usepackage[nameinlink,noabbrev]{cleveref}
\usepackage{bibunits}

\setlist{parsep=0pt,listparindent=\parindent}

{\fontsize{9.75}{10}\selectfont \defaultbibliographystyle{Sledge-TGARS-2020-1col-1} \defaultbibliographystyle{IEEEtran}}

\DeclareCaptionFont{singlespacing}{\setstretch{1}}
\numberwithin{figure}{section}
\renewcommand{\thefigure}{\arabic{section}.\arabic{figure}}
\renewcommand{\dblfloatpagefraction}{0.999}
\renewcommand{\dbltopfraction}{0.999}

\renewcommand \thesection{\arabic{section}}
\numberwithin{equation}{section}

\hypersetup{bookmarks=false,bookmarksopen=false,pdfpagemode=empty,pdfstartview=}

\let\oldnl\nl
\newcommand{\nonl}{\renewcommand{\nl}{\let\nl\oldnl}}

\title{\singlespacing\sf\huge Weakly-Supervised Semantic Segmentation of Circular-Scan, Synthetic-Aperture-Sonar Imagery}
\author{Isaac J. Sledge, \emph{Member, IEEE}, Dominic M. Byrne, \emph{Member, IEEE}\\ Jonathan L. King, \emph{Member, IEEE}, Steven H. Ostertag, \emph{Member, IEEE},\vspace{-0.01cm}\\ Denton L. Woods, \emph{Member, IEEE}, James L. Prater, \emph{Member, IEEE},\vspace{-0.01cm}\\ Jermaine L. Kennedy, \emph{Member, IEEE}, Timothy M. Marston, \emph{Member, IEEE},\vspace{-0.01cm}\\$\;\;\;\;\;\;\;\;\;\;\;$and Jos\'{e} C. Pr\'{i}ncipe, \emph{Life\! Fellow, IEEE}%
\thanks{\fontdimen2\font=1.55pt Isaac J. Sledge is the Senior Machine Learning Scientist with the Advanced Signal Processing and Automated Target Recognition Branch, Naval Surface Warfare Center, Panama City, FL, USA (email: isaac.j.sledge.civ@us.navy.mil).  He is also the Principal Machine Learning Scientist with the Machine Intelligence Defense (MIND) lab at the Naval Sea Systems Command.}
\thanks{\fontdimen2\font=1.55pt Dominic M. Byrne is an Engineer with the Advanced Signal Processing and Automated Target Recognition Branch, Naval Surface Warfare Center, Panama City, FL, USA (email: dominic.m.byrne3.civ@us.navy.mil).}
\thanks{\fontdimen2\font=1.55pt Jonathan L. King is an Engineer with the Applied Sensing and Processing Branch, Naval Surface Warfare Center, Panama City, FL, USA (email: jonathan.l.king1.civ@us.navy.mil).}
\thanks{\fontdimen2\font=1.55pt Steven H. Ostertag is an Engineer with the Applied Sensing and Processing Branch, Naval Surface Warfare Center, Panama City, FL, USA (email: steven.h.ostertag.civ@us.navy.mil).}
\thanks{\fontdimen2\font=1.55pt Denton L. Woods is an Engineer with the Littoral Acoustics and Target Physics Branch, Naval Surface Warfare Center, Panama City, FL, USA (email: denton.l.woods.civ@us.navy.mil).}
\thanks{\fontdimen2\font=1.55pt James L. Prater is the Senior Acoustic Sensing Scientist with the Advanced Signal Processing and Automated Target Recognition Branch, Naval Surface Warfare Center, Panama City, FL, USA (email: james.l.prater.civ@us.navy.mil).}
\thanks{\fontdimen2\font=1.55pt Jermaine L. Kennedy is the Division Head of the Sensing Sciences and Systems Division, Naval Surface Warfare Center, Panama City, FL, USA (email: jermaine.l.kennedy.civ@us.navy.mil).}
\thanks{\fontdimen2\font=1.55pt Timothy M. Marston is a Principal Engineer with the Applied Physics Laboratory, University of Washington, Seattle, WA, USA (email:\newline\noindent marston@apl.washington.edu).}
\thanks{\fontdimen2\font=1.55pt Jos\'{e} C. Pr\'{i}ncipe is the Don D. and Ruth S. Eckis Chair and Distinguished Professor with both the Department of Electrical and Computer Engineering and the Department of Biomedical Engineering, University of Florida, Gainesville, FL, USA (email: principe@cnel.ufl.edu).  He is the director of the Computational NeuroEngineering Laboratory (CNEL) at the University of Florida.\vspace{0.1cm}}
\thanks{The first and last authors were funded by grants N00014-19-WX-00636 (Marc Steinberg), N00014-21-WX-00476 (J. Tory Cobb), N00014-21-WX-00525 (Thomas McKenna), and N00014-21-WX-01348 (Marc Steinberg) from the US Office of Naval Research.  The first author was additionally supported by in-house laboratory independent research (ILIR) grant N00014-19-WX-00687 (Frank Crosby) from the US Office of Naval Research and a Naval Innovation in Science and Engineering (NISE) grant from NAVSEA.}%
}
\begin{document}
\bstctlcite{IEEEexample:BSTcontrol}

\maketitle
\RaggedRight\parindent=1.5em
\fontdimen2\font=2.1pt
\vspace{-1.25cm}\begin{abstract}\normalsize\setstretch{0.9}
\vspace{-0.25cm}{\small{\sf{\textbf{Abstract}}}}---We propose a weakly-supervised framework for the semantic segmentation of circular-scan synthetic-aperture-sonar (CSAS) imagery.  The first part of our framework is trained in a supervised manner, on image-level labels, to uncover a set of semi-sparse, spatially-discriminative regions in each image.  The classification uncertainty of each region is then evaluated.  Those areas with the lowest uncertainties are then chosen to be weakly labeled segmentation seeds, at the pixel level, for the second part of the framework.  Each of the seed extents are progressively resized according to an unsupervised, information-theoretic loss with structured-prediction regularizers.  This reshaping process uses multi-scale, adaptively-weighted features to delineate class-specific transitions in local image content.  Content-addressable memories are inserted at various parts of our framework so that it can leverage features from previously seen images to improve segmentation performance for related images.

We evaluate our weakly-supervised framework using real-world CSAS imagery that contains over ten seafloor classes and ten target classes.  We show that our framework performs comparably to nine fully-supervised deep networks.  Our framework also outperforms eleven of the best weakly-supervised deep networks.  We achieve state-of-the-art performance when pre-training on natural imagery.  The average absolute performance gap to the next-best weakly-supervised network is well over ten percent for both natural imagery and sonar imagery.  This gap is found to be statistically significant.
\end{abstract}%
\begin{IEEEkeywords}\normalsize\singlespacing
\vspace{-1.35cm}{{\small{\sf{\textbf{Index Terms}}}}---Seabed segmentation, seabed classification, semantic segmentation, imaging sonar, convolutional networks, deep learning}
\end{IEEEkeywords}
\IEEEpeerreviewmaketitle
\allowdisplaybreaks
\setstretch{0.9}

\begin{bibunit}
\bstctlcite{IEEEexample:BSTcontrol}

\vspace{-0.4cm}\subsection*{\small{\sf{\textbf{1.$\;\;\;$Introduction}}}}\addtocounter{section}{1}

The localization and classification of targets in sonar-derived imagery is crucial in many problems.  Both processes can, however, be challenging to automate in a robust manner.

Automation challenges arise because there is a strong functional relationship between seabed characteristics and the ability to both detect and label targets on the seafloor.  Underwater environments with seafloor features like kelp fields, rocky outcroppings, and coral reefs complicate sonar-based target-recognition systems.  This is because targets may be either partly or completely obscured and hence difficult to detect in the presence of distracting clutter.  Conversely, environments with simplistic seafloor features, like flat, sandy bottoms, simplify target analysis.  Proud targets are often easily located for this case.

As a consequence of this relationship, knowledge of the seafloor conditions can promote more accurate automated target analyses \cite{WilliamsDP-jour2014a}.  Such knowledge can also aid in quantifying the uncertainty in the target analysis predictions and determine if the underlying target analysis models should be believed in certain circumstances.  Moreover, seafloor segmentations can aid in the performance estimation of automated target analysis models.

In this paper, we propose a deep-network framework for the semantic segmentation of targets and seafloor types.  Our aim is to infer dense segmentation masks from sonar imagery that can help realize the aforementioned benefits.

It is important to note that we are not the first to address seafloor segmentation for imaging sonar.  Many approaches exist \cite{PaceNG-jour1988a,MalinvernoA-jour1989a,AlexandrouD-jour1993a,ChakrabortyB-jour2003a,AmiriSimkooeiAR-jour2011a,ChakrabortyB-jour2001a} (see \hyperref[sec2]{Section 2}).  A commonality amongst them is that they operate on bathymetry measurements from hydroacoustic echo sounders \cite{MichalopoulouZH-jour1995a,HellequinL-jour2003a,ChakrabortyB-jour2004a}.  Others process either side-scan, real-aperture-sonar (LRAS) imagery or side-scan, synthetic-aperture-sonar (LSAS) imagery \cite{StewartWK-jour1994a,HuangSW-jour2018a} collected in a linear, strip-map search mode \cite{HayesMP-jour2009a}.  The remainder tend to act on either topographic or natural-image modalities.

In contrast to existing works, our framework acts on circular-scan, synthetic-aperture-sonar (CSAS) imagery \cite{FriedmanAD-conf2005a,CallowHJ-conf2009a,CallowHJ-conf2010a,MarstonT-conf2011a,MarstonT-conf2012a,MarstonTM-jour2016a,KennedyJL-jour2014a}.  We do this since such imagery contains a great amount of information that is effective for automated seafloor segmentation.  For instance, CSAS imagery commonly exhibits greatly improved shape resolvability \cite{MitchellSK-conf2002a,FergusonBG-jour2009a} compared both LRAS- and LSAS-derived imagery.  Small-scale seabed features can thus be resolved, yielding a more physically accurate understanding of the seafloor.  CSAS imagery also provides complete-aspect coverage of underwater environments within a small circular region and partial-aspect coverage outside of that region.  LRAS and LSAS images collected using a linear search pattern often offer either a single view or limited set of views, in contrast \cite{MignotteM-jour2000a,MignotteM-jour2000b}, and hence have extremely limited aspect coverage.  Non-complete aspect coverage can significantly increase segmentation uncertainty due to the presence of acoustic shadows that obscure the surveyed environment.  We refer readers to the online appendices for an overview of our chosen imaging modality and illustrations of its advantages (see \hyperref[secA]{Appendix A} and \hyperref[secA]{Appendix B}).

While CSAS imagery offers several benefits, it can be challenging to annotate a sufficient amount of it for semantic segmentation.  The number of classes that can be resolved in a CSAS image is often much higher than either LRAS or LSAS imagery, which burdens human annotators.  The incredibly high resolution of the imagery compounds the issue.  Similar difficulties are encountered when handling natural imagery \cite{CordtsM-conf2016a,NeuholdG-conf2017a,ZhuY-jour2021a}.  

To substantially reduce per-image annotation times, we constrain our framework to operate in a weakly-supervised way (see \hyperref[sec3]{Section 3}).  We train part of our framework in a supervised fashion using global, image-level class labels.  We then form class-activation maps.  We use low-uncertainty regions in those maps in another part of the framework to infer local, pixel-level semantic segmentation masks in an unsupervised manner.  At no point does our framework leverage human-derived, pixel-level annotations for training.

More specifically, our framework relies on multiple networks to iteratively extract details for constructing and refining segmentation masks.  One network returns image-level labels of the classes that it deems to be present in an image.  Class-activation maps \cite{ZhouB-conf2016a,SelvarajuRR-conf2017a} are then constructed from these global labels to recover some spatial class information about the scene content.  Such maps highlight discriminative image regions that can serve as localized seed cues for each observed seafloor and target class.  Not all of the cues may be reliable, though.  We thus quantify the classification uncertainty of each cue to select only those that are related to image content seen in the training samples.  The chosen cues are then passed to another network that iteratively adjusts a pixel-level segmentation map in response to local image content.  We re-weight the features within this network to encourage large-margin feature separability for each class, which improves segmentation quality.  A third network forms adaptive superpixels to ensure that the uncovered segmentation extents align well with class boundaries observed in the imagery.  Each of these networks can be either trained separately or joined and trained in an end-to-end manner.

Our framework can operate on a single image at a time.  This would, however, ignore shared information across images that may enhance segmentation quality \cite{HuangZ-conf2019a,SunG-conf2020a}.  To leverage such details, we incorporate convolutional, content-addressable memories into each of the aforementioned networks.  The memories store class-specific contexts extracted from multiple images.  Contexts are recalled, as necessary, for new images presented to the networks.  Those contexts are subsequently transformed and then synthesized into the intermediate feature representations of the networks.  We additionally permit simultaneously segmenting multiple CSAS images.  This is because certain regions may lose full-aperture coverage and hence become difficult to annotate in different images of the same general area.  We do this by a fourth network that uncovers dense scene correspondences and warps candidate segmentation masks for further refinement.  Propagating and utilizing segmentations from one sonar image typically improves solutions for other images captured in the same area.  Analogous improvements in multi-image classification rates have been demonstrated in \cite{WilliamsDP-conf2016a} for sonar imagery and in \cite{RotherC-conf2006a,QuanR-conf2016a,LiB-conf2019a} for natural imagery.

In this paper, we empirically assess our weakly-supervised segmentation framework using real-world CSAS imagery.  We illustrate that it reliably isolates and labels seafloor types when processing either one or more CSAS images (see \hyperref[sec4]{Section 4}).  When pre-training on benchmark datasets of natural imagery, our framework yields state-of-the-art performance (see \hyperref[secE]{Appendix~E}).  

We are the first to consider the problem of semantic segmentation for CSAS imagery.  There are hence no existing sonar-focused models available for comparison.  To evaluate our work, we take supervised deep networks for natural-image semantic segmentation and then train them on CSAS imagery.  We highlight that our weakly-supervised framework performs almost on par with these networks for the more than ten bottom types and ten target types that we consider.  Ample extended results are included in online appendices to emphasize certain claims (see \hyperref[secC]{Appendix~C}, \hyperref[secD]{Appendix~D}, \hyperref[secE]{Appendix~E}, and \hyperref[secF]{Appendix~F}).  Moreover, we show that our framework additionally outperforms state-of-the-art segmentation networks that are semi-supervised and that we have adapted to sonar imagery.  We offer justifications for this behavior, which extend from discussions of our framework's novelties (see \hyperref[sec2]{Section 2}).  We also demonstrate that the performance gaps are statistically significant with incredibly high probability of rejecting the null hypothesis.

\phantomsection\label{sec2}
\subsection*{\small{\sf{\textbf{2.$\;\;\;$Literature Review}}}}\addtocounter{section}{1}

In this section, we survey the literature on the segmentation of acoustic imagery (see \hyperref[sec2.1]{Section 2.1}).  We then provide overviews of the literature on supervised and weakly supervised semantic segmentation for natural imagery (see \hyperref[sec2.2]{Section 2.2}).  For both topics, we compare our framework to the cited research.

\phantomsection\label{sec2.1}
\subsection*{\small{\sf{\textbf{2.1.$\;\;\;$Semantic Segmentation of Acoustic Images}}}}

A great amount of research has been done for seafloor characterization, segmentation, and annotation.  Much of the research can be divided into one of two categories, those that process side-scan sonar images and those that process digital terrain models produced by hydroacoustic echo sounders.

Conventional sonar-image-based approaches \cite{PaceNG-jour1988a,StewartWK-jour1994a,SchockSG-jour1994a,PearceSK-jour2013a,HuangSW-jour2018a,WilliamsDP-jour2009b,WilliamsDP-jour2015b} for these tasks typically rely on shallow decision-making architectures applied to manually extracted features.  In \cite{StewartWK-jour1994a}, Stewart et al. outline over forty features to quantify spectral properties and image-based geometric primitive attributes that describe texture content.  A shallow network topology was used to select informative features for distinguishing between three mid-ocean-ridge terrain types.  In \cite{WilliamsDP-jour2009b}, Williams proposes a Bayesian fusion approach for seabed classification in multi-view LSAS imagery.  He employs wavelet-based features to assess local image frequency content and models the distributions of such features for each of the four possible classes with a mixture of Gaussians.  Williams also proposes applying the concept of lacunarity, or intensity variation assessment, to the task of seafloor segmentation \cite{WilliamsDP-jour2015b}.  He showed that using this feature with a simple maximum-likelihood classifier could separate well three seafloor classes.  Cobb et al. \cite{CobbJT-jour2010a} develop a supervised-trainable sonar texture model that is derived from autocorrelation functions of the LSAS imaging point-spread functions and the autocorrelation functions of the seabed texture sonar cross sections.  They demonstrate that it can characterize well four different bottom types.  More recently, Huang et al. \cite{HuangSW-jour2018a} rely on statistical and information-theoretic features obtained from side-scan-sonar scanlines to recognize and segment three seafloor categories with a na\"{i}ve Bayes classifier.

Many seafloor annotation approaches operate on backscatter and bathymetry data from single-beam and multi-beam echosounders \cite{PaceNG-jour1988a,AlexandrouD-jour1993a,MichalopoulouZH-jour1995a,ClarkeJH-jour1994a,HellequinL-jour2003a,ChakrabortyB-jour2000a,ChakrabortyB-jour2004a,ChakrabortyB-jour2015a,MalinvernoA-jour1989a,StewartWK-jour1994b,LandmarkK-jour2014a,SnellenM-jour2019a}.  Such modalities provide insights into crucial segmentation properties like seafloor height and roughness, along with sediment bulk density and layering \cite{SimonsDG-jour2007a}.  Model-based strategies are widely employed.  An early work, from Pace and Gao \cite{PaceNG-jour1988a}, extracts power-spectrum features from backscattered returns to distinguish six different classes using a shallow network.  Stewart et al. \cite{StewartWK-jour1994b} show that multi-modal statistical models can be fit to histograms of echo amplitude responses, among other features, to distinguish between three geological regions.  Alexandrou and his colleagues \cite{AlexandrouD-jour1993a,MichalopoulouZH-jour1995a} propose applying parametric models to the raw and transformed acoustic signals to recognize three seafloor classes.  Chakraborty et al. \cite{ChakrabortyB-jour2001a,ChakrabortyB-jour2003a,ChakrabortyB-jour2015a} advocate using nonparametric frameworks for handling backscatter data in segmentation and labeling tasks, as they avoid the need for computationally intensive pre-processors so that physics-based frameworks can be employed \cite{ChakrabortyB-jour2000a}.  More recently, both Landmark et al. \cite{LandmarkK-jour2014a} and Snellen at al. \cite{SnellenM-jour2019a} offered parametric Bayesian models for assessing differences in backscattering strength for various sediment types.  Image-based schemes have also been researched.  Clarke \cite{ClarkeJH-jour1994a} highlights that it is helpful to consider the angular response of backscattering if certain classification ambiguities are to be resolved when processing texture content.  Dugelay et al. \cite{DugelayS-conf1996a,AugustinJM-conf1997a} provide a random-field model, within a Bayesian framework, for partitioning echosounder image mosaics into homogeneous seafloor regions.

To our knowledge, there has been no published work on CSAS-based semantic segmentation.  While the aforementioned annotation approaches for LRAS or LSAS could be applied to CSAS imagery, many would be poorly suited for the latter modality.  It is unlikely that manually-specified features would capture a range of characteristics needed to detect and label the more than ten bottom types that we are interested in recognizing in this study.  They would have immense difficulties with our multi-aspect characterization of acoustic signatures.  The models used by the above authors also would likely have difficulties scaling to the more than ten target classes that we consider.  Similar claims apply to the shallow models that have been developed for echosounder backscatter signatures.

We believe that deep-network architectures, like our framework, would be more appropriate for these tasks.  Multi-aspect representations of the seabed can be learned in a data-driven fashion when using such networks.  Robust segmentation strategies and region classifiers can also be concurrently constructed.  Moreover, deep networks have been shown to handle thousands of classes well in the presence of distractors, provided that sufficient training samples are available.  The performance of some shallow models has been known to saturate, after only a few classes, when relying on handcrafted features.

\phantomsection\label{sec2.2}
\subsection*{\small{\sf{\textbf{2.2.$\;\;\;$Semantic Segmentation of Natural Images}}}}

There are a variety of deep-network architectures for natural images that could be adapted for the problem of automated seafloor segmentation.  For instance, several convolutional networks can be trained in a weakly-supervised manner to localize targets \cite{OquabM-conf2015a,HoffmanJ-conf2015a,BilenH-conf2016a,ZhouB-conf2016a,LiD-conf2016a,DibaA-conf2017a,LinCH-conf2017a,ShenY-conf2018a,WangC-conf2014a,ShiM-conf2016a,KantorovV-conf2016a,CinbisRG-jour2017a,SanginetoE-jour2019a}.  Early works in this area rely on separately-trained components.  For instance, Wang et al. \cite{WangC-conf2014a} develop a semantic clustering method to process pre-trained convolutional features and detect object extents.  Cinbis et al. \cite{CinbisRG-jour2017a} combine multi-fold, multiple-instance learning with convolutional features to bias against prematurely looking for certain object types in erroneous locations.

While methods like these produce promising results, they cannot always be trained end-to-end.  Their performance can sometimes suffer due to not letting the network training guide the entire feature extraction, transformation, and inference process.  Work on end-to-end architectures has therefore intensified in recent years.  Oquab et al. \cite{OquabM-conf2015a}, for example, propose a convolutional architecture to determine targets' positions.  Zhou et al. \cite{ZhouB-conf2016a} extend this network design to encourage the network to cover the full extent of the targets, not just discern their positions in the scene.  Bilen et al. \cite{BilenH-conf2016a} build upon region-based convolutional networks \cite{GirshickR-conf2015a}.  They suggest using dual processing streams, one focusing on classification and the other on localization.  Lin and Lucey \cite{LinCH-conf2017a} design a network with spatial-transformer layers that automatically warp feature maps to align objects to common reference frames, which facilitates localization.

Several other classes of weakly-supervised approaches exist.  These often rely on either bounding boxes \cite{DaiJ-conf2015a}, contours and scribbles \cite{LinD-conf2016a,VernazaP-conf2017a}, points \cite{BearmanA-conf2016a,PapadopoulosDP-conf2017a}, or image-level labels \cite{WeiY-conf2017a,FanR-conf2018a,LeeJ-conf2019a,FanJ-conf2020a,FanJ-conf2020b} to provide a minimal amount of training supervision while significantly reducing per-image human annotation times.

For the latter type of methods, it is common to first train a deep-network classifier on global, image-level tags.  The classifiers are then modified to return an activation heatmap for each class.  Such heatmaps codify the correlations between the classes that the network observes and their spatial locations in the imagery.  The heatmaps are, however, not spatially exhaustive.  They often highlight only those regions that contain discriminative textures and patterns, not necessarily all of the regions that belong to the class.  They may also be rather imprecise.  The aim of these weakly-supervised methods is to iteratively modify the activation mappings to better encompass the full spatial extent of the classes.  This is done without relying on local, pixel-level information to guide the transformation.  For some works, this can entail estimating and merging multiple activation mappings to expand their combined spatial extent, as Wei et al. \cite{WeiY-conf2018a}, Huang et al. \cite{HuangZ-conf2018a}, and several others did.  As Jiang et al. show, it can also involve accumulating activation maps and assessing how they change, across different samples, to obtain a more thorough understanding of the spatial class structure \cite{JiangP-T-conf2019a}.  A second strategy involves iteratively erasing images to encourage the classifiers to learn about secondary discriminative details \cite{WeiY-conf2017a,ZhangX-conf2018b,SunK-conf2021a}.  This can yield more spatially complete activation maps not only for foreground targets, but also for backgrounds.  Other works introduce regularizers that provide pseudo-supervision \cite{ShimodaW-conf2019a,AhnJ-conf2019a}.  Wang et al. \cite{WangY-conf2020a} advocate using consistency constraints, in the form of pixel correlations combined with affinity attention maps, to bias against over- and under-activation in the class heatmaps.  Chang et al. \cite{ChangYT-conf2020a} rely on clustering to generate quasi-sub-category labels that are used to learn more spatially-complete activations.

Our framework is related to approaches like \cite{WeiY-conf2017a,FanR-conf2018a,LeeJ-conf2019a,FanJ-conf2020a,FanJ-conf2020b}.  This is because it also infers and leverages class-activation mappings.  There are several differences, though.  First, our framework extracts multi-scale, local-global spectral features \cite{ChiL-coll2020a}.  These help to maintain uniformity of the class activations across affine transformations, yielding more reliable and more complete maps.  Our use of other heuristics \cite{JungH-conf2021a,HartleyT-conf2021a} also improves map quality.  Second, we assess and utilize the pixel-level classification uncertainty of the activation maps.  In doing so, we can find those regions in new images that most mimic properties of the training samples \cite{SinghR-conf2020a,SinghR-jour2021a} and select them as seed points for spatial classification.  Such points tend to be highly discriminative.  Even a single point per observed class can provide enough supervision to promote semi-accurate segmentations.  We further improve the segmentation accuracy by imposing a large-margin constraint on features from different classes.  As a byproduct of this constraint, we obtain a feature-space organization that preserves semantic relationships and tends to mitigate many types of segmentation errors.  Third, we develop and incorporate a deep-superpixel approach to promote strong coherence with observable class boundaries.

One of the biggest differences between our framework and previous ones is that it exploits cross-image information.  Our framework does this in two ways.  The first way is by merging segmentation masks derived from images captured in nearby locations.  We do this via dense, pixel-level flow fields.  After such fields are inferred, the corresponding segmentation maps can be warped along them and then combined to correct segmentation errors that occur in partial-aspect regions of the sonar imagery.  This approach stems from earlier ideas developed for co-segmentation \cite{QuanR-conf2016a,TaniaiT-conf2016a,ShenT-conf2017a,LiG-conf2017a,HsuKJ-conf2019a} but tends to yield better performance, as we show in our experiments.  The second way is through the use of convolutional, content-addressable memories.  We scatter memory layers throughout the various framework components and use them to store complementary, class-sensitive details.  When new images are presented, appropriate details are recalled from memory and merged with the currently extracted features to enhance the information content.  This synthesis occurs at multiple feature scales to not only guide classification at the earliest parts of the framework, but also aid in the formation of accurate segmentation maps in the latter stages.  

Several authors have developed multi-image strategies for feature mining \cite{HuangZ-conf2019a,ZhangF-coll2021a,LiuY-jour2022a,FanJ-jour2023a}.  However, they typically require constructing pixel-level affinity maps \cite{SunG-conf2020a}, which can be quite prone to errors in weakly-supervised-trained settings.  These maps are also usually intractable to compute when considering more than two images.  The remaining authors, like \cite{FanJ-jour2023a}, use non-content-addressable memories.  Their choice of memories can cause performance to saturate after only a few entries, though.  Segmentation performance can even degrade, since there is no reliable way to recall relevant content.  Here, we consider highly general memories that do not possess a memory-depth versus precision trade-off.  Growing the memory-bank size hence yields a graceful, nearly-monotonic increase in segmentation performance.

Much work has also been done on fully-supervised methods for semantic segmentation \cite{GirshickR-conf2014a,HariharanB-conf2015a,ChengY-conf2017a,LongJ-conf2015a,DaiJ-conf2015a,EigenD-conf2015a,NohH-conf2015a,ZhengS-conf2015a,HariharanB-conf2014a,PinheiroPO-conf2016a,BadrinarayananV-jour2017a,ChenLC-jour2018a}.  Some early methods \cite{GirshickR-conf2014a,HariharanB-conf2014a} rely on classifying region proposals to generate segmentation results.  Such works are commonly trained in a piecewise manner, however, which does not always yield good generalization performance.  Recently, end-to-end-trained, fully-convolutional networks \cite{LongJ-conf2015a,DaiJ-conf2015a} have shown to be effective at extracting good segmentation features and have thus become a popular option.  Fully-convolutional-based methods have the limitation of yielding low-resolution predictions.  Many researchers have since extended these networks to generate higher-fidelity segmentation maps.  The {\sc DeepLab-CRF} architecture \cite{ChenLC-jour2018a}, for example, constructs coarse score maps and then applies bilinear upsampling and conditional-random-field models to refine segmentation boundaries.  {\sc CRF-RNNs} \cite{ZhengS-conf2015a} extend this network by implementing recurrent layers for end-to-end learning.  Transposed-convolution methods \cite{ChengY-conf2017a,NohH-conf2015a,BadrinarayananV-jour2017a} learn filters that upsample low-resolution predictions while generally preserving hard segmentation boundaries.  Eigen and Fergus \cite{EigenD-conf2015a} perform coarse-to-fine learning of multiple networks with different resolution outputs for refining the coarse prediction.  Long et al. and Pinheiro et al. \cite{LongJ-conf2015a,PinheiroPO-conf2016a} add prediction layers to middle portions of the fully-convolutional networks to generate prediction scores at multiple resolutions.

\addtocounter{section}{1}
\begin{wrapfigure}{r}{0.7\textwidth}
   \vspace{-0.125cm}\hspace{-0.05cm}\includegraphics[width=4.65in]{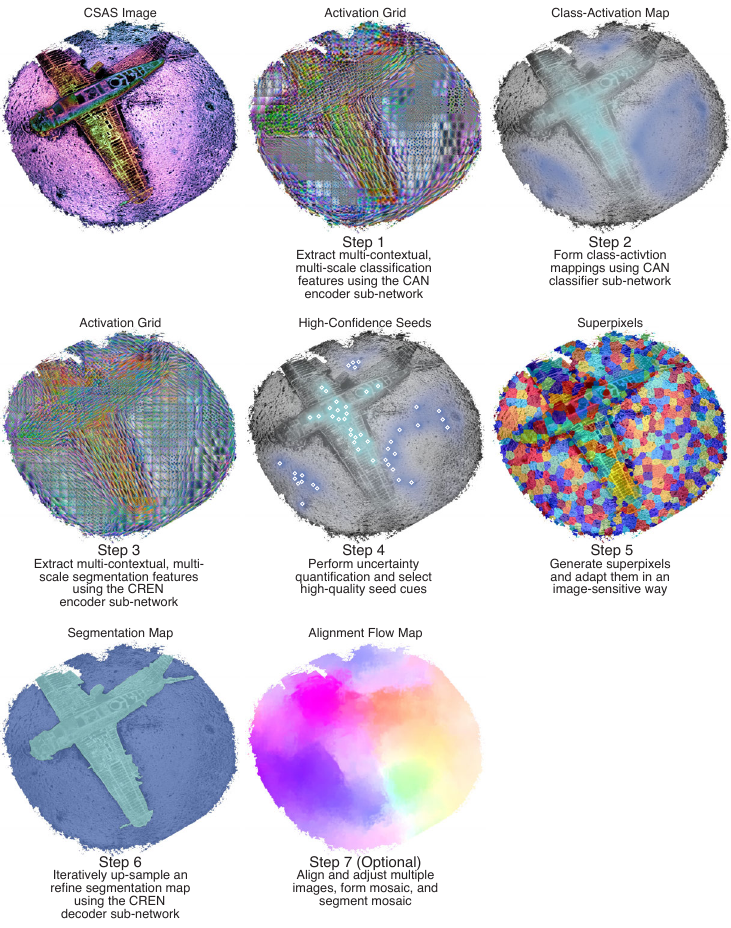}\vspace{-0.1cm}
   \caption[]{\fontdimen2\font=1.55pt\selectfont A summary of the major steps in our framework.  Steps 1 and 2 are addressed by the network in \cref{fig:can-network}.  Steps 3, 4, 5, and 6 are addressed by the networks in \cref{fig:cren-network} and \cref{fig:usn-network}.  Step 7 is addressed by the network in \cref{fig:sfn-network}.  Note that bathymetric details are shown in the above CSAS image.  Depth cues are hence available for this scene.  We do not utilize such cues in the current study to help with segmentation.  This is because interferometry data was only captured for a small subset of the scenes in the dataset that we use.  However, bathymetry can be leveraged by our framework without any design changes.\vspace{-0.2cm}}
   \label{fig:framework}
\end{wrapfigure}
\addtocounter{section}{-1}

Akin to \cite{LongJ-conf2015a,DaiJ-conf2015a}, our framework is predominantly convolutional.  It can also be trained in an end-to-end way, leading to feature representations that leverage all of the available supervisory signals.  Our framework has several practical advantages, though, compared to the above works.  It does not suffer from low-resolution prediction issues found in some networks \cite{LongJ-conf2015a,ZhengS-conf2015a,ChengY-conf2017a,NohH-conf2015a,BadrinarayananV-jour2017a,PinheiroPO-conf2016a}.  This is because our framework uses spectral convolution and spectral transposed convolution to create filters with large receptive fields, which help yield high-resolution maps.  Spectral unpooling is also employed to effectively upsample the maps, further enhancing their resolution.  Our framework additionally constrains the segmentation maps to adhere to observed class boundaries, yielding accurate annotations.  Such behavior stems from our use of deep-superpixel-based structured prediction.  Perhaps the biggest benefit of our framework is that it only requires sparsely annotated samples to do well.  Unlike fully-supervised networks, which need dense, pixel-level labels, ours requires only global, image-level labels.  The former can take several minutes to supply per image, while the latter require only a few seconds.  Significantly less upfront human effort is therefore needed to create a meaningful training set than for fully-supervised frameworks.  While there is a performance trade-off that comes with using less supervision, we show that our framework substantially narrows the gap to fully-supervised networks.

\phantomsection\label{sec3}
\subsection*{\small{\sf{\textbf{3.$\;\;\;$Methodology}}}}\addtocounter{section}{1}

Our aim is to realize robust semantic segmentation capabilities for imaging sonar.  We simultaneously want to reduce human annotation efforts compared to the conventional setting where pixel-level masks are used for training.

There many ways that we can do this.  We have opted to exploit, as much as possible, the weak supervision provided by image-level labels of the classes that appear in the CSAS imagery.  Through a multi-network framework, we seek to construct a set of dense, pixel-level segmentation maps for these images.  A high-level flowchart of our framework is given in \cref{fig:framework}.

In this section, we outline our framework.  We first present our convolutional, class-activation-mapping network ({\sc CAN}) (see \hyperref[sec3.1]{Section 3.1}).  This network is presented in \cref{fig:can-network}.  It is trained in a supervised way to report the classes present in a given sonar image.  We couple the {\sc CAN} with a class-activation-mapping approach, Lift-CAM, to recover spatial information correlated with the predicted labels.  An example is provided in \cref{fig:can-cam-examples}.

The pseudo-segmentation masks returned by Lift-CAM may, unfortunately, not be spatially exhaustive.  Only highly discriminative regions in the images will usually be highlighted.  In some cases, spurious locations may be selected.  The masks therefore must be further transformed to be effective for segmentation.  While processes like structured prediction can help constrain the class affinities to observed boundaries, they are not designed to reject incorrect spatial class assignments.  We hence require a more principled way of amending the maps.

We develop a convolutional region-expansion network ({\sc CREN}) to iteratively broaden the class structure observed in the Lift-CAM-inferred class-activation mappings and correct some of the over- and under-activations observed in them (see \hyperref[sec3.1]{Section 3.1}).  This network is shown in \cref{fig:cren-network}.  Our {\sc CREN} takes as input a sonar image and its corresponding class-activation mapping returned by the {\sc CAN}.  From these dual inputs, the {\sc CREN} extracts, in an unsupervised manner, a series of multi-scale features that are used to deduce an initial segmentation map.  This map is then progressively upscaled and refined.

Due to a lack of strong supervision, a naive implementation of the {\sc CREN} may make significant errors.  We introduce several regularizers in an attempt to preempt this from happening (see \hyperref[sec3.2]{Section 3.2}).  We first apply information potential fields to analyze the prediction uncertainty within the class-activation mappings.  We automatically remove the most uncertain spatial locations from the affinity mapping.  The features associated with the remaining locations are used as supervisory cues to impose large-margin, between-class separability.  We simultaneously enforce within-class compactness.  The interplay of both constraints yields a feature representation that mitigates many types of segmentation mistakes.  To reduce the potential for other mistakes, we leverage deep superpixels formed by an unsupervised superpixel network ({\sc USN}) (see \hyperref[sec3.1]{Section 3.1}).  This network is provided in \cref{fig:usn-network}.  The {\sc USN} helps to not only ensure a locally consistent class assignment, but also preserves observable class boundaries.  This is illustrated in \cref{fig:cren-examples}.

Due to how we survey underwater scenes, we sometimes have batches of CSAS images that come from nearby regions.  Seafloor characteristics in a subset of these images may be better resolved than in others.  As an optional step in our framework, we directly exploit this improved resolvability.  We first spatially co-register pairs of CSAS images using a small-scale flow network ({\sc SFN}) (see \hyperref[secC]{Appendix C}).  We then overlap and merge the CSAS images, which yields a mosaic.  We process this mosaic with the other networks in our framework.

All of these networks can be trained separately.  It is also possible to train them in an end-to-end manner.

\phantomsection\label{sec3.1}
\subsection*{\small{\sf{\textbf{3.1.$\;\;\;$Network Architectures}}}}

We rely on three networks to perform weakly-supervised semantic segmentation of CSAS imagery.  These are the {\sc CAN}, {\sc CREN}, and {\sc USN}.  A fourth network, the {\sc SFN} is an optional part of our processing pipeline since not all scenes may possess multiple CSAS images.

\begin{figure*}
   \hspace{-0.15cm}\includegraphics[]{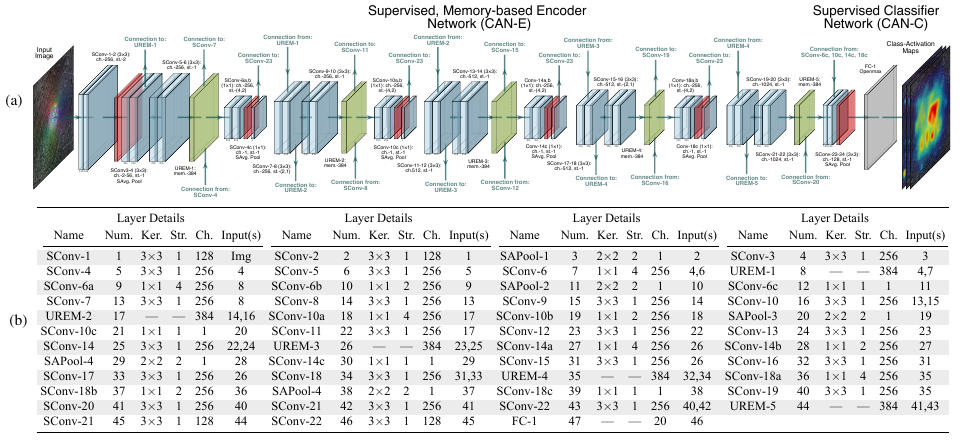}\vspace{-0.15cm}
   \caption[]{\fontdimen2\font=1.55pt\selectfont An overview of our supervised network for assessing class-activation mappings. (a) A network diagram of our class-activation network ({\sc CAN}) and the resulting class-activation outputs for a CSAS image containing a variety of seafloor types.  The encoder branch of our network ({\sc CAN-E}) relies on a series of multi-scale convolution banks that extract local-global spectral features.  These banks enable our network can discern high-quality semantic features with both small and large receptive fields despite using only a few filters.  Universal, recurrent memory layers are inserted within the branch to store and recall multi-context details to further improve the feature quality in an efficient way.  All of these features are aggregated in the deepest part of the encoder before being transformed into a probabilistic, open-set classification response in the classifier branch ({\sc CAN-C}).  Lift-CAM is then used to infer a class-activation mapping from the {\sc CAN} classification response.  For this diagram, spectral convolutional layers (SConv) are denoted via light blue blocks.  Darker blue bands on these blocks are used to signify that rectified-linear-unit activations are applied.  Green blocks correspond to the content-addressable, universal-recurrent memory cells (UREM).  Spectral average pooling layers (SAPool) are denoted using red blocks.  The fully-connected, openmax aggregation layer (FC) is denoted using a gray block.  (b) A tabular summary of the major network layers of the {\sc CAN}.  For each layer, we list its name, its numerical order in the network, the kernel size, the stride either the number of channels or the number of elements, and the index of the layer that feeds into it.  We recommend that readers consult the electronic version of this paper to see the full image details.\vspace{-0.4cm}}
    \label{fig:can-network}
\end{figure*}

Our {\sc CAN} possesses a straightforward architecture (see \cref{fig:can-network}).  It is composed of a five-block feature encoder (SConv-1 through SConv-22).  Each of these blocks progressively transforms a CSAS image into a mixture of local and global features with high semantic content.  A small convolutional decoder (SConv23 through FC-1) converts the features into open-set classification probabilities \cite{BendaleA-conf2016a}.  After training, the network is modified to enable the formation of boundary-preserving class-activation mappings.

\begin{figure*}
   \hspace{-0.15cm}\includegraphics[]{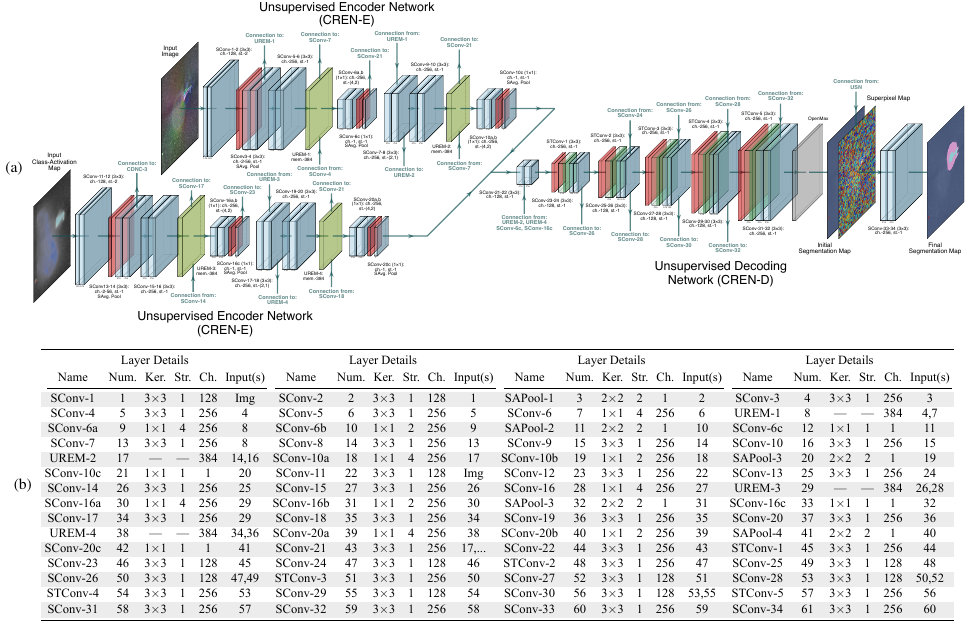}\vspace{-0.15cm}
   \caption[]{\fontdimen2\font=1.55pt\selectfont An overview of our unsupervised network for semantic segmentation. (a) A network diagram of our convolutional region-expansion network ({\sc CREN}) the resulting segmentation outputs for a CSAS image containing a crashed helicopter.  The encoder branches of our network ({\sc CREN-E}) rely on a series of multi-scale convolution banks that extract local-global spectral features.  These features are obtained both for the input image and from the class-activation mappings produced by the {\sc CAN} in \cref{fig:can-network}.  Dual inputs are used to provide both high- and low-level cues so that the network can good semantic features that aid in classification.  Using only a single input, such as the class-activation mapping, can impede segmentation, since local image content is obscured by the pixel affinities.  Universal, recurrent memory layers are inserted within the branch to store and recall multi-context details to further improve the feature quality.  The features are aggregated in the deepest part of the encoders.  We apply an adaptive, large-margin regularization to re-organize the features.  This helps mitigate segmentation errors.  An initial segmentation map is formed in the decoder stage ({\sc CREN-D}) and is progressively upsampled and refined before an openmax activation is applied to yield a probabilistic classification response.  Superpixels, generated from our unsupervised superpixel network ({\sc USN}) in \cref{fig:usn-network}, are then employed to enforce consistent spatial labeling.  The block coloring scheme used in \cref{fig:can-network} is reused in this diagram.  We use red blocks followed by green blocks to denote upsampling by spectral, transposed local-global convolution layers (STConv).  (b) A tabular summary of the major network layers of the {\sc CREN}.  We recommend that readers consult the electronic version of this paper to see the full image details.\vspace{-0.4cm}}
   \label{fig:cren-network}
\end{figure*}

Our {\sc CREN} has a more complex design than the {\sc CAN} (see \cref{fig:cren-network}).  The initial portion of the network contains dual convolutional encoders (SConv-1 through SConv-20) that operate on the class-activation mappings and the input sonar image.  We process both inputs, not just the former, so as to reintroduce low-level image features that can provide additional cues about the scene for segmentation.  The encoders feed into a decoder that progressively forms and refines a segmentation map in a weakly-supervised way (SConv-21 through SConv-32).  The segmentation map is systematically regularized throughout the decoder to improve its quality.  Near the end of the decoder, the map is adjusted to better obey observed class boundaries (SConv-33 and SConv-34).

\begin{figure*}
   \hspace{-0.15cm}\includegraphics[]{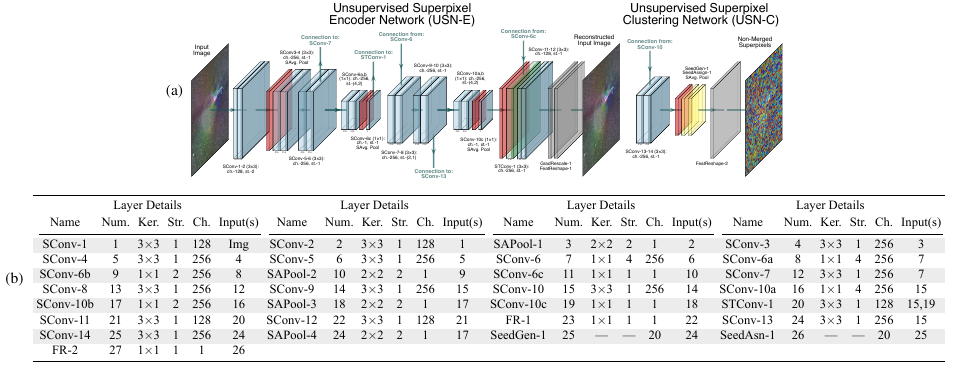}\vspace{-0.15cm}
   \caption[]{\fontdimen2\font=1.55pt\selectfont An overview of our unsupervised network for generating superpixels. (a) A network diagram of our unsupervised superpixel network ({\sc USN}) the resulting segmentation outputs for a CSAS image containing a crashed helicopter.  The encoder branch of our network ({\sc USN-E}) relies on dual multi-scale convolution banks that extract local-global spectral features from the CSAS image.  These features are aggregated and then upscaled to approximately reconstruct the CSAS image.  This yields an incredibly compact network that eschews feature redundancy, which facilitates robust superpixel generation.  Seed cues are selected from the embedded features in the decoder stage ({\sc USN-C}).  Non-iterative region merging occurs to yield a spatial grouping of pixels.  The block coloring scheme used in \cref{fig:can-network} is reused in this diagram.  Yellow-colored blocks denote non-iterative clustering layers for either generating the initial superpixels or assigning pixel affinity.  (b) A tabular summary of the major network layers of the {\sc USN}.  We recommend that readers consult the electronic version of this paper to see the full image details.\vspace{-0.4cm}}
   \label{fig:usn-network}
\end{figure*}

The map adjustment at the end of the {\sc CREN} is facilitated by the {\sc USN} (see \cref{fig:usn-network}).  This network minimal transformations the sonar image (SConv-1 through SConv-10) to learn a set of features.  Few layers are used to avoid features that severely overfit.  The {\sc USN} then uses those features to approximately reconstruct the input (STConv-1 through FeatReshape-1).  This ensures that the features capture relevant characteristics about the scene content.  Those features are simultaneously employed to infer a set of seed cues for non-iteratively grouping spatial regions (SConv-13 through FeatReshape-2).  The result is a set of superpixels.  Since we tend to over-segment underwater scenes, nearby superpixels are merged in a post-processing step.

We include several standard processes within these networks.  For each convolutional block in the encoder, we insert skip connections \cite{SrivastavaRK-coll2015a} between the first and last layers.  Skip connections help shorten training times due to avoiding singularities caused by model non-identifiability \cite{OrhanAE-conf2018a}.  Such connections also routinely preempt the occurrence of vanishing gradients \cite{HuangG-conf2017a}.  Batch normalization \cite{IoffeS-conf2015a} is applied after all of the convolutional layers in the network.  Its role is to improve the training speed and stability of the network by smoothing the optimization landscape \cite{SanturkarS-coll2018a}.  Leaky-rectified-linear-unit activations are cascaded after batch normalization to introduce non-linearities into the feature transformation process \cite{NairV-conf2010a}.  The activations help boost performance, which is a byproduct of their sparsity-promoting behaviors \cite{MehtaD-conf2019a}.

There are several aspects of our networks that make them unique compared others that infer class-activation mappings.  One of these is the type of feature-extraction layers that we use, which are spectral, local-global convolutional layers \cite{ChiL-coll2020a}.  Such layers efficiently implement mixed-size receptive fields, which aid in the analysis of underwater scenes that have classes with vastly different physical extents.  Conventional deep networks, in comparison, rely on convolutional layers that focus on contrast features extracted at extremely local scales and must stack layers to iteratively widen the receptive field.  Many layers can be required, however, to achieve a sufficiently wide field for completing a given task.  Even processes like deformable convolution \cite{DaiJ-conf2017a}, non-local convolution \cite{WangX-conf2018a}, and \`{a}-trous convolution \cite{ChenLC-jour2018a} may not always facilitate lowering the overall layer count as effectively as spectral convolution. 

Although spectral convolutions possess non-local receptive fields, large amounts of filters can be required, per layer, to adequately characterize differently-sized classes.  To reduce the filter count, we augment the network topologies so that they occasionally extracts multi-scale contrast features.  We do this by connecting sets of four additional layers to the initial blocks in the encoder portions of the networks.  The first two layers for each set use spectral, local-global convolutions to non-linearly transform the feature maps.  Spectral average pooling \cite{RippelO-coll2015a} is then applied to encourage the networks to focus on higher-level concepts that aid in robust classification.  The final layer in each set resizes the feature maps from each block so that they can be combined at the end of the decoder.  Despite the produced feature maps being the same size, they are computed for different receptive fields.  They hence represent contextual features at multiple scales.

Another change is the type of pooling that is performed.  Typically, max pooling is used for classifier networks.  Such a pooling operation impedes inferring correlations between global, image-level labels and local, pixel-level content, though \cite{OquabM-conf2015a}.  We instead use spectral average pooling \cite{RippelO-coll2015a}.  There are two reasons for this.  First, it better preserves spatial information than max pooling.  It therefore helps to identify the complete extent of an observed class \cite{ZhouB-conf2016a}.  The second reason is that max pooling has a tendency to propagate sonar speckle \cite{WilliamsDP-jour2020a}, a type of random noise.  If speckle noise is not adequately filtered, then useful features about the classes can be inadvertently ignored in the initial layers and hence remain missing throughout the rest of the network.

The final difference is the inclusion of universal-recurrent memory cells, a type of content-addressable memory \cite{DouR-jour2023a}.  Such recurrent layers permit efficiently storing and recalling multi-image contextual features.  That is, the network, in conjunction with a linear controller, discerns what information it has seen in existing samples that it should explicitly encode and subsequently combine with new samples to help boost classification performance.  These features, we find, are often associated with samples from under-represented classes.  They may also be from those samples that are difficult to correctly classify for the current set of convolutional features.  As we show in an associated online appendix, our use of this type of memory has distinct performance advantages over existing recurrent architectures (see \hyperref[secF]{Appendix~F}).  It has been previously shown to resolve the memory-depth versus resolution trade-off dilemma.  Classification rates can thus readily increase as a function of the memory size.  Achieving similar functionality has been difficult with other memories and has thus limited the generalization capabilities of existing cross-image segmentation networks \cite{WeiY-conf2017a,FanR-conf2018a,LeeJ-conf2019a,FanJ-conf2020a,FanJ-conf2020b}.

In an online appendix, we provide a comprehensive ablation study (see \hyperref[secF]{Appendix~F}).  We show that each of these contributions has a non-negligible improvement over the baseline where only convolutional layers are used.

\phantomsection\label{sec3.2}
\subsection*{\small{\sf{\textbf{3.2.$\;\;\;$Network Training}}}}

{\small{\sf{\textbf{CAN Training and Inference.}}}} We train the {\sc CAN} on image-level labels using mini-batch-based back-propagation with a categorical cross-entropy loss
\begin{equation*}
\mathcal{L}_\textnormal{CAN} = \sum_{i = 1}^n\sum_{j = 1}^d \Bigg(\frac{\omega_{i,j}}{n}\textnormal{log}(\psi_{i,j}^\theta) \!+\! \frac{(\!1 \!-\! \omega_{i,j})}{n}\textnormal{log}(1 \!-\! \psi_{i,j}^\theta) \Bigg) \!+ \lambda \|\theta\|_2^2.
\end{equation*}
\noindent This loss function handles noisy labels well \cite{ZhangZ-coll2018a}, including those that are incorrectly specified by human annotators.  Here, the mini-batch size is given by $n$, while $d$ is the total number of classes.  The non-negative parameter $\lambda$ provides the weighting for an $L_2$-ridge factor, which helps prevent overfitting.  The term $\omega_{i,j}$ represents the ground-truth, image-level label, while $\psi_{i,j}^\theta$ is the predicted network response, the latter of which depends on the network parameters $\theta$.

\begin{figure*}
   \hspace{0.65cm}\includegraphics[width=5.7in]{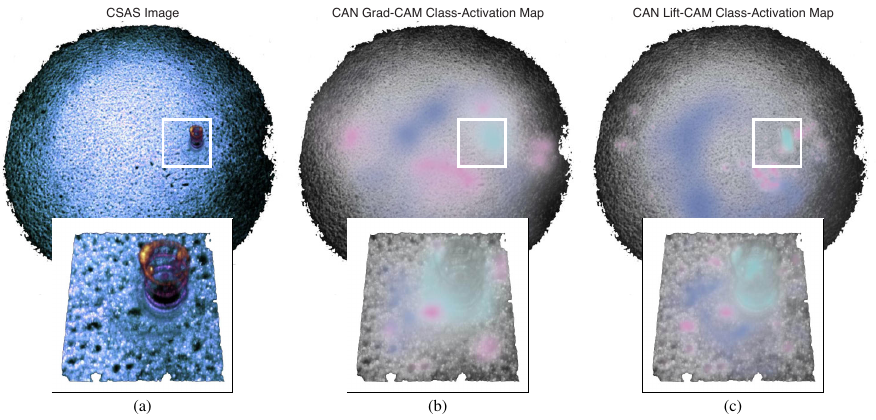}\vspace{-0.1cm}
   \caption[]{\fontdimen2\font=1.55pt\selectfont Examples of assessing feature importance as a pre-processor for weakly-supervised segmentation.  (a) A bathymetric CSAS image of an underwater scene that contains a plastic barrel.  A non-color-by-aspect encoding is used here.  (b) A Grad-CAM-inferred class-activation mapping from our {\sc CAN} \cite{ZhouB-conf2016a,SelvarajuRR-conf2017a}.  (c) A Lift-CAM-inferred class-activation mapping from our {\sc CAN} \cite{JungH-conf2021a}.  The former would provide poor seed cues for semantically segmenting this scene.  The latter, however, would not for this scene.  This is largely because Lift-CAM scores quantify how the expected model performance changes when conditioning on a particular feature.  Such scores weight features based on their importance and often emphasize well class-specific regions in the samples.  In this scene, they highlight a majority of the plastic barrel.  They also identify well the flat sand and indented sand regions.\vspace{-0.35cm}}
   \label{fig:can-cam-examples}
\end{figure*}

After training, we remove the final spectral average pooling layer in the {\sc CAN} and use Lift-CAM \cite{JungH-conf2021a} to form a set of class-activation mappings for each class $j$.  We do this by first iterating over the following SHAP-based expression \cite{LundbergSM-coll2017a} for all possible feature orderings,
\begin{equation*}
\varphi_{k,q} = \sum_{\gamma'_k \subset \Gamma}\frac{(g_{k} \!-\! |\gamma'_k|)!(|\gamma'_k| \!-\! 1)!}{g_{k}!}\Bigg(\omega_j(h_\alpha(\gamma'_k)) \!-\! \omega_j(h_\alpha(\gamma'_k\backslash q))\Bigg).
\end{equation*}
\noindent Here, $\omega_j$ denotes the target output the network.  As well, $\gamma'_k$ is a binary vector obtained from $\alpha_{k,q}$, the $q$th feature map at the $k$th layer.  This vector codifies if $\alpha_{k,q}$ maintains its original activation values, which occurs if $\gamma_{k,q} \!=\! 1$, or does not, which occurs if $\gamma_{k,q} \!=\! 0$.  The number of non-zero entries in $\gamma'_k$ is specified by $|\gamma'_k|$.  Additionally, $\gamma'_k\backslash q$ means that $\gamma_{k,q}' \!=\! 0$.  The number of feature maps at a given layer is expressed by $g_{k}$.  We also have that $h_\alpha$ is a mapping that converts $\gamma'_k$ into the activation embedding space.  That is, $\alpha_{k,q} \!=\! h_\alpha(\Gamma)$, where $\Gamma$ is a vector of ones, which implies that $\gamma_{k,q}' \!=\! 1$ is mapped to $\alpha_{k,q}$ and $\gamma_{k,q}' \!=\! 0$ to the zero vector that has the same dimension as $\alpha_{k,q}$.  Due to the\\ \noindent computational intensity of considering all possible feature orderings, we use a DeepLift approximation \cite{ShrikumarA-conf2017a} to find the coefficients $\varphi_{k,q}$ for all $k$ and $q$.

After iterating, we form a class-activation map via $f(\sum_{q=1}^r \varphi_k' \alpha_{k,q})$, with $\varphi_k' \!=\! \varphi_{k,0} \!+\! \sum_{q=1}^r \gamma_{k,q}'\varphi_{k,q}$ and $f$ being some activation function.  Here, $\varphi_{k,0}$ can be viewed as a baseline explanation, while $\varphi_{k,q}$ denotes the importance of the $q$th feature map at the $k$th layer.  The entries of the binary vector $\gamma'_k$ specify either the presence or absence of a given feature.  These class-activation mappings are thus affinity heatmaps that describe how much certain spatial image content is predictive for a given class (see \cref{fig:can-cam-examples}).

Lift-CAM includes several enhancements over traditional class-activation-mapping methods that help it infer state-of-the-art visual explanations.  The most important of these is a quantitative ranking of feature importance that satisfies notions of missingness, consistency, and localized accuracy, which is implicitly provided by the SHAP-based expression.  In leveraging this ranking, Lift-CAM can often mitigate either over-activating or under-activating.  Our use of DeepLift allows Lift-CAM to avoid a gradient-saturation problem encountered in other class-activation methods.  We further explore the benefits of Lift-CAM in our experimental discussions (see \hyperref[sec4]{Section 4}).

Lift-CAM may not always produce boundary-sensitive mappings.  It can thus yield visual explanations that can become mixed and hence impede segmentation in later stages of our framework.  To minimize this chance, we post-process the class-activation mappings with structured prediction \cite{BuloSR-conf2012a}.

\vspace{0.15cm}{\small{\sf{\textbf{CREN/USN Training and Inference.}}}} The class-activation mappings from the {\sc CAN} may contain erroneous affinities even after applying structured prediction.  The affinity magnitudes may also not be completely correlated with the observed spatial image content and its influence on the classification response.  Solely processing the raw class-activation mappings with our {\sc CREN} can hence yield poor segmentation maps.

\begin{figure*}
   \hspace{0.65cm}\includegraphics[width=5.7in]{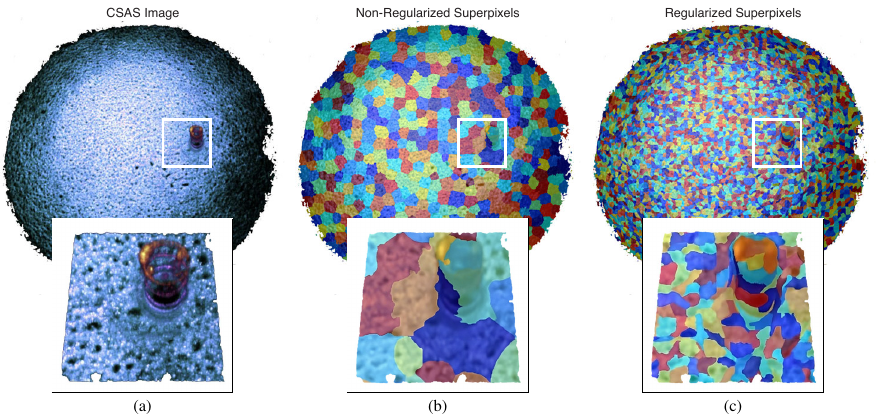}\\
   \hspace{0.65cm}\includegraphics[width=5.7in]{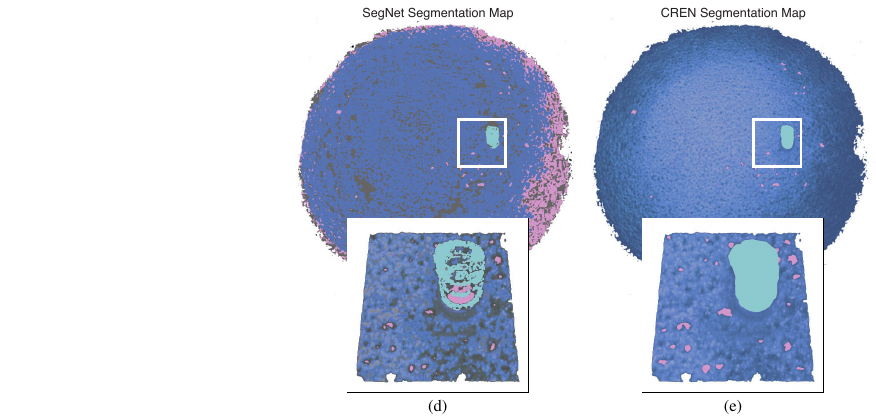}\vspace{-0.1cm}
   \caption[]{\fontdimen2\font=1.55pt\selectfont Examples of the benefits of learning a regularized representation.  (a) A bathymetric CSAS image of an underwater scene that contains a plastic barrel.  A non-color-by-aspect encoding is used here.  (b) An SLIC superpixel segmentation \cite{AchantaR-jour2012a}.  (c) The segmentation from our adaptive deep network before entropy-based regularization and superpixel group reduction.  Since SLIC relies on hand-crafted features, it may not always be suitable for complicated imaging modalities like the one we consider here.  Unless a large number of superpixels are used, then class boundaries may be incorrectly identified, yielding poor semantic segmentation performance.  In this instance, the true extents of the barrel are not correctly identified in (b) while they are in (c).  However, even a poor regularization is still more effective than eschewing regularization entirely.  As shown in (d), a {\sc SegNet} without any local spatial constraints returns poor segmentation masks.  Our {\sc CREN} returns a result, given in (e), that better aligns with the ground truth.\vspace{-0.35cm}}
   \label{fig:cren-examples}
\end{figure*}

We correct for some of these mistakes via a series of regularizers.  The first is local label homogeneity via inferred superpixels.  We use our {\sc USN} for generating these superpixels.  This network is trained with the following loss,
\begin{multline*}
\mathcal{L}_\textnormal{USN} = \sum_{i=1}^n\sum_{j=1}^c\sum_{p=1}^q \Bigg( u_{i,j,p}^\theta\textnormal{log}(u^\theta_{i,j,p}) \!-\! u^\theta_{i,j,p}\textnormal{log}(\tau^\theta_{i,j,p}) \!+\! \frac{\tau_{i,j,p}^\theta\delta_{\|t_p - s_j\|_1 \leq \epsilon_p}}{\tau_{i,j,p}^\theta(1 \!-\! \delta_{\|t_p - s_j\|_1 \leq \epsilon_p})}\Bigg) \,+ \\
\eta\Bigg(\sum_{p \notin Q}\Bigg(\frac{\omega_p}{m}\textnormal{log}(\psi_p^\theta) \!+\! \frac{\pi}{m}\textnormal{log}(\textnormal{cosh}(\phi_p \!-\! \phi_p')) \Bigg) \!+\sum_{p \in Q}\Bigg(\frac{\omega_p}{m'}\textnormal{log}(\psi_p^\theta) \!-\! \frac{\pi}{m'}T_p\textnormal{log}(\textnormal{cosh}(\phi_p \!-\! \phi_p'))\Bigg)\!\Bigg).
\end{multline*}
Here, the minibatch size is given by $n$.  The initial number of superpixels is given by $c$.  Both $\psi^\theta_p$ and $\omega_p$ are taken to be, respectively, the {\sc USN} image reconstruction and the expected response for the $p$th pixel.  The term $\phi_p$ corresponds to the spatial pixel indices predicted by the {\sc USN}, while the term $\phi_p'$ are the anticipated indices.  The set $Q$ contains the indices of pixels that are near the contours between superpixels.  The number of pixels outside and within this set are given by $m$ and $m'$, respectively.  The contour map is denoted by $T_p$, which we derive using the approach in \cite{HeJ-jour2022a}.  We take $\tau_{i,j,p}^\theta$ to be the soft assignment of the $p$th pixel to the $j$th group, which is specified by Student's $t$-distribution.  We modulate all of the soft-assignment values by a binary indicator variable that prohibits assigning a non-zero belongingness of a pixel $t_p$ to a superpixel $s_j$ if the $L_1$ distance between them exceeds a non-negative threshold $\epsilon_p$.  This threshold is determined automatically by limiting a pixel to only partly belong to the fifteen nearest superpixels.  After training, we would like the soft assignment to be as close as possible to some reference value.  However, we lack a target grouping.  We thus form an auxiliary version that we attempt to mimic, which is given by $u^\theta_{i,j,p} \!=\! (\tau_{i,j,p}^\theta\tau_{i,j,p}^\theta/\mu_{j,p}^\theta)/\sum_{k=1}^c \tau_{i,k,p}^\theta\tau_{i,k,p}^\theta/\mu_{k,p}^\theta$, with $\mu_{j,p}^\theta \!=\! \sum_{i=1}^n \tau_{i,j,p}^\theta$ being soft grouping frequencies \cite{XieJ-conf2016a}.\\ \noindent  The remaining variables, $\eta$ and $\pi$, are non-negative weighting coefficients. 

For this loss, the first term is a modified Kullback-Leibler divergence.  It penalizes the inferred pixel grouping from deviating too much from the auxiliary grouping.  It also encourages image pixels to not belong to far-away superpixels.  Our inclusion of this term yields a type of self-training \cite{NigamK-conf2000a} wherein we leverage the pixel assignments for the current batch to update the network and improve the pixel assignments for future batches.  We have altered the standard divergence with an extra factor that biases against returning sub-optimal groupings.  The second and third loss terms codify the quality of the image reconstruction.  The third term biases the non-spatial contrast features that are near the fringes of currently-defined superpixels.  This ensures that the superpixel boundaries coincide with stark changes in image contrast and hence the possible transition from one semantic class to another.  The interplay of each loss term tends to yield better superpixels than alternative approaches (see \Cref{fig:cren-examples}).

During training and inference, we form the superpixels in a non-iterative way.  We first fit a uniformly spaced grid to each image in the batch.  We then linearly transform the pooled features via a learned tensor that projects to a two-dimensional space.  We apply a non-linear activation function and view the response as a series of horizontal and vertical shifts for the specified grid points.  Adding these shifts to the grid points yields the superpixel seeds.  As in training, we use Student's $t$-distribution to quantify the similarity of a pixel to a superpixel seed and modulate it by a binary indicator variable to avoid assigning pixels to far-away seeds.  Since we tend to oversegment scenes, we merge neighboring superpixels at random until the entropy of the superpixel features increases beyond a specified threshold.  Superpixels that contain {\sc CAN}-derived seed cues with different labels are never merged.  This yields a compact representation that is sensitive to how scene content changes locally.

Another form of regularization that we consider comes from our information-theoretic uncertainty quantification measure \cite{SinghR-conf2020a,SinghR-jour2021a}.  This measure is used to evaluate which portions of the input imagery, at the pixel level, best match those of the training samples used to learn the {\sc CAN} parameters.  We take the lowest uncertainty pixels for each of the observed classes in a class-activation mapping to be weak constraints for the following {\sc CREN} loss function,
\begin{equation*}
\mathcal{L}_\textnormal{CREN} = \sum_{i = 1}^n\sum_{p = 1}^q \Bigg(\Omega_{i,p}\frac{\beta_{i,p}}{nq}\textnormal{log}(\psi_{i}^\theta) \Bigg) \!+ \rho\Bigg(\textnormal{log}\Bigg(\sum_{i,j=1}^n\sum_{p=1}^q \delta_{\Omega_{i,p} \neq \Omega_{j,p}}\frac{1 \!-\! H^{\sigma,\theta}_{i,j}}{nnq}\Bigg) \!- \textnormal{log}\Bigg(\sum_{i,j=1}^n\sum_{p=1}^q \delta_{\Omega_{i,p} = \Omega_{j,p}}\frac{H^{\sigma,\theta}_{i,j}}{nnq} \Bigg)\!\Bigg).
\end{equation*}
Here, the minibatch size is given by $n$, while the number of pixels per minibatch sample is given by $q$.  For the left-most expression, we have overloaded $\psi_{i}^\theta$ to be the predicted {\sc CREN} response for the $i$th batch sample.  This response naturally depends on the network parameters $\theta$.  The binary variable $\Omega_{i,p}$ denotes to which class the $p$th pixel belongs.  If a given pixel was not selected as a seed cue by our uncertainty quantification measure, then $\Omega_{i,p}$ is a $d$-dimensional zero vector for all $p$.  For a given seed cue, $\beta_{i,p}$ is inversely proportional to the class population.  Otherwise, for all non-labeled pixels, $\beta_{i,p} \!=\! 0$.  In the two remaining expressions, $H^{\sigma,\theta}_{i,j}$ represents the matrix-based cross-entropy\\ \noindent measure \cite{SledgeIJ-jour2023a} for the features of the $i$th and $j$th batch samples.  This measure depends on the positive hyperparameter $\sigma$ and, implicitly, on the network parameters $\theta$ that form the feature set.  The cross-entropy magnitudes are modulated by binary indicator variables that depend on the class labels for $\Omega_{i,p}$ and $\Omega_{j,p}$.  The non-negative variable $\rho$ is a weighting factor.

\begin{figure*}
   \includegraphics[width=6.25in]{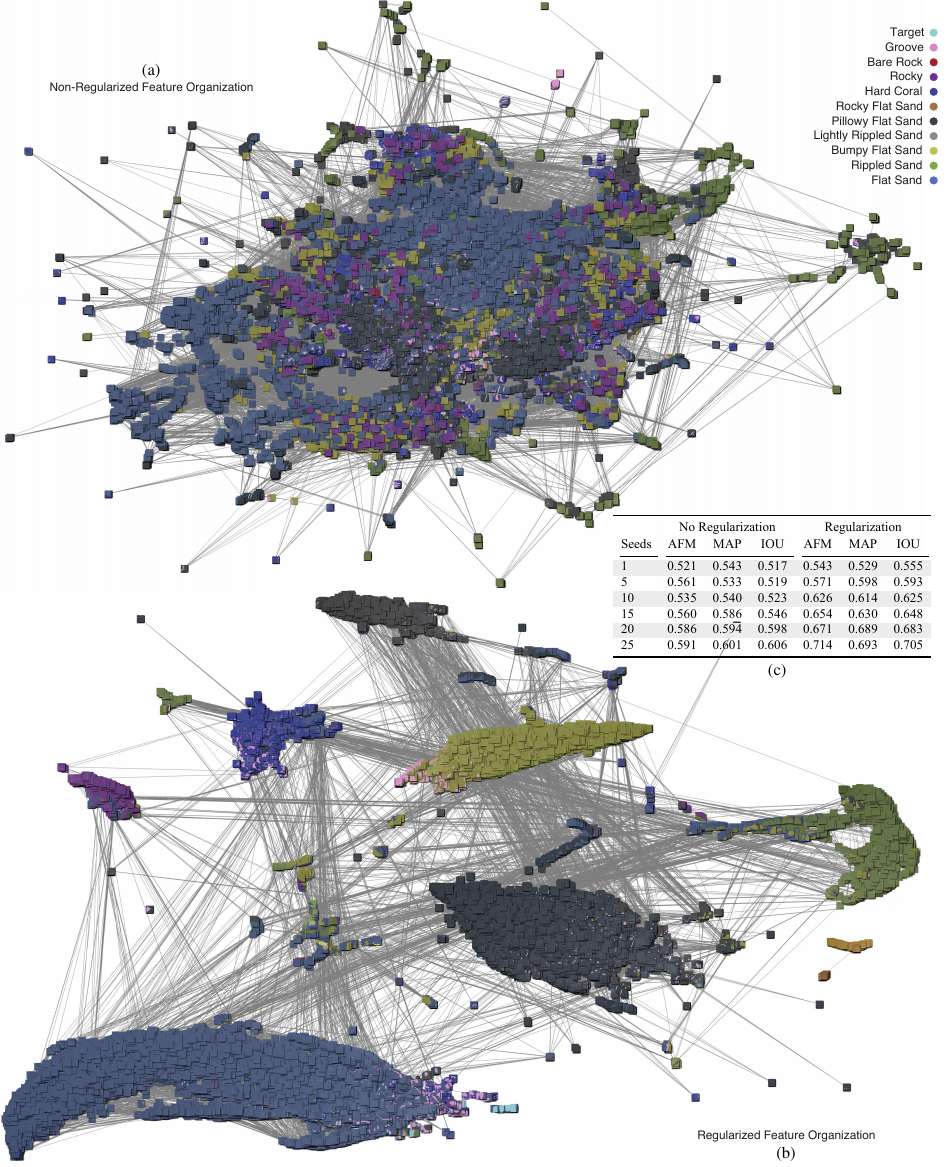}
   \caption[]{\fontdimen2\font=1.55pt\selectfont Visualization of the latent tile spaces for the dominant seafloor classes in our imaging sonar dataset.  In (a), we show a UMAP projection of the features formed after the dual encoder networks without feature regularization.  We have overlaid the $\textnormal{100}\!\times\!\!\textnormal{100}$ image tiles used to generate these features and the corresponding ground-truth segmentation maps on each point.  They highlight that, without feature regularization, the {\sc CREN} learns a highly non-separable class representation.  Accurate segmentation is still possible but is difficult to achieve without adequate supervision.  In (b), we show a similar UMAP projection for $\textnormal{100}\!\times\!\!\textnormal{100}$ image tiles.  In this case, information-theoretic feature regularization is used when training the {\sc CREN}.  The regularization yields a highly separable representation at the tile level.  It also learns one at the pixel and super-pixel level.  This trait simplifies segmentation immensely.  In both (a) and (b), the lines between tiles indicate up to five nearest neighobors in the original, non-projected feature space.  Connections are removed if the tile-to-tile distance in the feature space exceeds some threshold.  This was done to retain some of the local data structure that is lost during projection.  In (c), we supply a table of the pixel-level average $f$-measure (AFM), mean average precision (MAP, and mean intersection-over-union (IOU) segmentation scores for the dominant seafloor types as a function of the number of seed cues.  The results show that performance steadily improves for increasing seed-cue counts when regularization is used.\vspace{-0.4cm}}
   \label{fig:feature-regularization}
\end{figure*}

In this loss function, the first term constrains the {\sc CREN} to learn a pixel-level mapping that obeys the labels from the {\sc CAN}-derived seed cues.  Non-labeled pixels whose feature values are similar to those of the seed cues are implicitly encouraged to possess a similar label.  The second term imposes that the features for seed cues from different classes should be highly distinct.  The third term enforces high feature similarity for all seed cues belonging to the same class.  The interaction of the latter two terms is another form of regularization, as it leads to the formation of a large-margin feature representation (see \Cref{fig:feature-regularization}).  

Near the end of the {\sc CREN}, we obtain a preliminary segmentation map.  This map may contain a fair amount of label variability that is spurious.  We introduce the superpixels generated by the {\sc USN} to provide local label uniformity and correct for some of the errors that arise.  The inferred label shared by the most pixels within a superpixel is taken as the label for the entire superpixel.  Ties are broken randomly.  If a given superpixel contains a {\sc CAN}-based seed cue, then the label of the {\sc CAN} cue overrides the current superpixel label.

\setcounter{figure}{0}
\phantomsection\label{sec4}
\subsection*{\small{\sf{\textbf{4.$\;\;\;$Segmentation Experiments}}}}\addtocounter{section}{1}

In this section, we assess the capability of our framework for the weakly-supervised segmentation of real-world CSAS imagery.  We first demonstrate that our {\sc CAN} can reliably construct class-activation maps (see \hyperref[sec4.1]{Section 4.1}).  We then show that the {\sc CREN} yields superpixel decompositions that immensely aid in the conversion of those class-activation maps to accurate segmentation maps (see \hyperref[sec4.2]{Section 4.2}).  Lastly, we evaluate the semantic segmentation performance of the {\sc CREN} (see \hyperref[sec4.2]{Section 4.2}).  In each case, we find that our framework components outperform widely-used alternatives that we have adapted to sonar imagery.  We obtain strong evidence that the performance gap to the alternatives is statistically significant.

We also produce several supplemental results alongside the ones we report in this section.  These are given in separate appendices (see \hyperref[secC]{Appendix~C}, \hyperref[secD]{Appendix~D}, \hyperref[secE]{Appendix~E}, and \hyperref[secF]{Appendix~F}).

\phantomsection\label{sec4.1}
\subsection*{\small{\sf{\textbf{4.1.$\;\;\;$Class-Activation-Map Results and Discussions}}}}

We first illustrate the ability of the {\sc CAN} to identify class-specific regions before comparing them with alternate approaches.  We then discuss why the {\sc CAN} is able to perform well.

\vspace{0.2cm}\noindent {\small{\sf{\textbf{Training Protocols.}}}} We pre-train and fine-tune the {\sc CAN}s using ADAM-based back-propagation gradient descent with mini batches \cite{KingmaDP-conf2015a}.  We use the default parameters for ADAM.  We use a mini-batch size of 64 samples to bias against terminating in poor local minima \cite{HardtM-conf2016a}.

Pre-training of the {\sc CAN}s is done using the ImageNet dataset \cite{DengJ-conf2009a}.  We then fit the {\sc CAN}s to our CSAS dataset to specialize the networks to imaging-sonar characteristics.  The discussions in \hyperref[secA]{Appendix~A} outline how we collect, encode, and annotate the CSAS dataset.  The studies in \hyperref[secB]{Appendix~B} justify the color-by-aspect encoding that we use for the sonar imagery.  We comment on the benefits of pre-training both at the end of this section and in \hyperref[secE]{Appendix~E}.

We augment both datasets during training by performing random translations, rotations, scalings, and croppings of the images.  We add multiplicative, normally-distributed random noise to the sonar imagery to simulate sonar speckle.  Localized random haze is sometimes included in the sonar imagery to mimic the effects of thermoclines and thus speed-of-sound changes.  All of these image transformations tend to improve performance.

We conduct multiple Monte Carlo pre-training simulations and fine-tuning simulations.  When pre-training on natural imagery, we rely on the default training, testing, and validation splits and randomize the sample order in each epoch.  Pre-training terminates once the {\sc CAN} loss on the validation set monotonically increases for ten consecutive epochs.  After all of the Monte Carlo trials have concluded, we retain the network weights that yield the best test-set performance as a checkpoint.  The memory banks in the networks are then cleared.  Fine-tuning then commences on the CSAS imagery from the checkpoint and is terminated in the same fashion.  We use a similar ratio of training, testing, and validation sample sizes in this latter case.  We randomly choose the split samples for every simulation.

To evaluate activation-map quality, we consider the increase in confidence (IC) \cite{ChattopadhayA-conf2018a}, average drop (AD) \cite{ChattopadhayA-conf2018a}, and the average drop in deletion (ADD) \cite{JungH-conf2021a} metrics.  Higher values of IC and ADD indicate better performance while lower values of AD indicate better performance.  We average these scores across the Monte Carlo trials for the reported results.  We do not include score standard deviations due to the number of class super-categories that we evaluate.  We bin the scores from each class into six superclasses to provide a broad view of performance trends.

\vspace{0.2cm}\noindent {\small{\sf{\textbf{Assessment Protocols.}}}} We are interested in assessing the performance difference between our framework and the alternatives.  In particular, we want to gauge if the difference is statistically significant and thus merits attention.  We use a multi-step process for this purpose.

We first employ a non-parametric Friedman's test \cite{FriedmanM-jour1940a} to gauge the likelihood that all of the methods perform the same and that the observed differences are merely random effects.  For us to obtain strong evidence to reject the corresponding null hypothesis, our chosen threshold of $p$$<$$\textnormal{10}^\textnormal{-4}$ must be satisfied.

Our use of Friedman's test is appropriate.  If we apply Mauchly's test \cite{MauchlyJW-jour1940a} and find that the implicit condition of sphericity can be rejected with high probability, at our threshold of $p$$<$$\textnormal{10}^\textnormal{-4}$, then ANOVA-like tests \cite{FisherRA-book1925a} would not be appropriate.  The post-hoc counterparts for ANOVA-like schemes, such as either the parametric Tukey test \cite{TukeyJW-jour1949a} or the parametric Dunnett test \cite{DunnettCW-jour1955a}, would similarly not be usable to assess the statistical significance of the methods' performance differences.

To determine if our framework's results are statistically significant, we need to pair Friedman's test with a post-hoc test.  Here, we use the non-parametric Nemenyi's test \cite{DamicoJA-jour1989a,WolfeH-book1999a} if sphericity is violated.  We cannot use either a Bonferroni test \cite{SimesRJ-jour1986a} or a Bonferroni-Dunn test \cite{DunnOJ-jour1961a}, since we do not have a model to act as a control.  Related schemes, like Holm's test \cite{HolmS-jour1979a} and Hommel's test \cite{HommelG-jour1988a}, would not be applicable for the same reason.

For each approach to which we compare, the average ranks for our framework should differ by at least the critical-difference amount, for our chosen threshold of $p$$<$$\textnormal{10}^\textnormal{-4}$.  If so, then the performance of our framework is likely to be statistically significant.

Nemenyi's means-rank test cannot control the maximum type-one error.  That is, it cannot control the probability of falsely declaring that any pair of methods has significantly different performance \cite{FlignerMA-jour1984a}.  Given this issue, we can additionally compare the approaches using the Wilcoxon signed-rank test \cite{WilcoxonF-jour1945a}.  For this test, we obtain strong evidence for rejecting the null hypothesis if our threshold of $p$$<$$\textnormal{10}^\textnormal{-3}$ is met.  Since we only have a single alternate hypothesis to explain our observations, the claims of statistical significance for our framework would stand.

Lastly, we perform an upper-bound analysis of the associated Bayesian factors \cite{BergerJO-jour1999a}.  The odds in favor of alternate hypotheses relative to the null hypotheses should be better than our selected threshold of $p$$<$$\textnormal{10}^\textnormal{-3}$.

\begin{figure}[t!]
   \hspace{0.0cm}\includegraphics[width=6.6in]{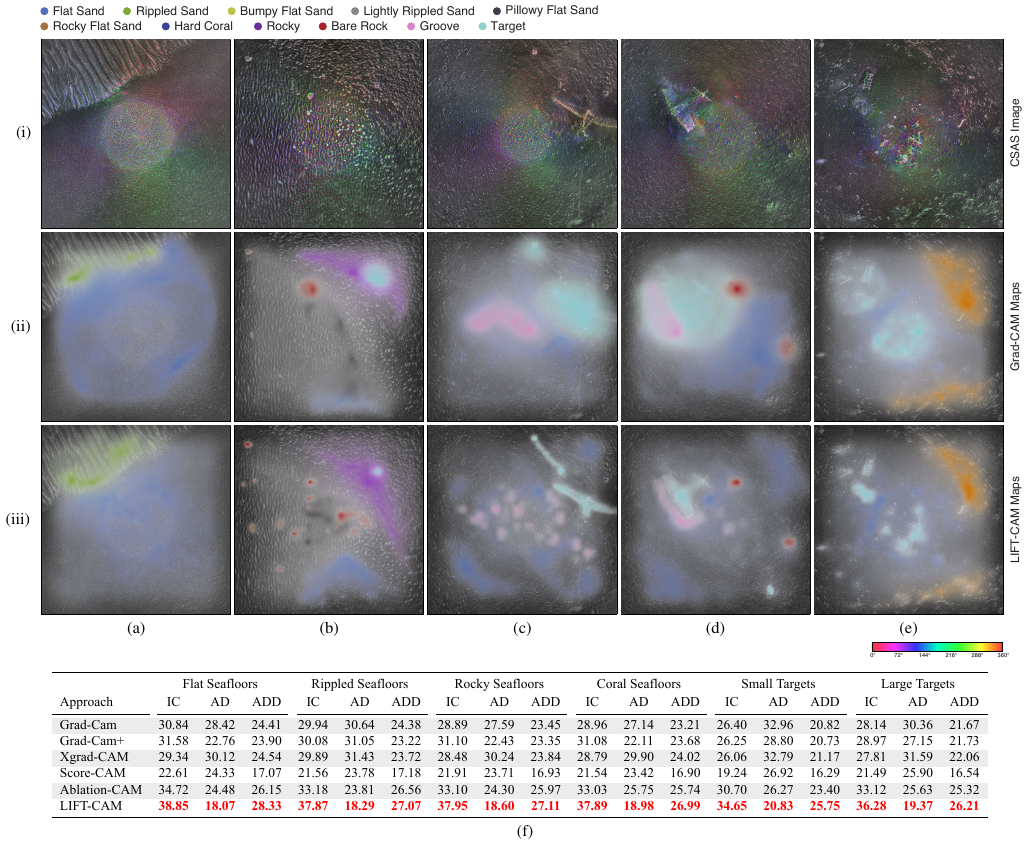}\vspace{-0.05cm}
   \caption[]{\fontdimen2\font=1.55pt\selectfont Class-activation mapping results.  In row (i), we provide examples of five underwater scenes.  The first scene, (a), highlights the transition zone between rippled sand and flat sand.  The second scene, (b), contains a field of bare rocks with pockets of flat sand and rippled sand.  A small clutter target is nestled among the rocks.  The third and fourth scenes, (c) and (d), are of crashed planes and debris on mostly-flat sand.  The fifth scene, (e), contains debris and man-made objects overlaid on flat sand.  A mixed seafloor class is also present.  In row (ii), we show corresponding Grad-CAM \cite{ZhouB-conf2016a,SelvarajuRR-conf2017a} class-activation maps obtained from our {\sc CAN}.  Grad-CAM is our chosen baseline.  In row (iii), we provide corresponding LIFT-CAM \cite{JungH-conf2021a} class-activation maps obtained from our {\sc CAN}.  Qualitatively, the maps from LIFT-CAM are superior to Grad-CAM, as they better highlight the full class extents.  A quantitative comparison of these two methods, along with Grad-CAM+ \cite{ChattopadhayA-conf2018a}, Xgrad-CAM \cite{FuR-conf2020a}, Score-CAM \cite{WangH-conf2020a}, and Ablation-CAM \cite{DesaiS-conf2020a}, are provided in the table in (f).  This table reinforces that LIFT-CAM, the approach we use in our framework, provides the best performance.  We recommend that readers consult the electronic version of the paper to see the full image details.}\vspace{-0.4cm}
   \label{fig:class-activation-results}
\end{figure}

\vspace{0.2cm}\noindent {\small{\sf{\textbf{Experimental Results.}}}} Example class-activation maps are provided in \cref{fig:class-activation-results}(ii)(a)--(e) and \cref{fig:class-activation-results}(iii)(a)--(e) for a series of underwater scenes in \cref{fig:class-activation-results}(i)(a)--(e).  In the former case, we use Grad-CAM as a baseline.  In the latter case, our preferred approach, Lift-CAM, is used.  The same {\sc CAN}s are used to infer maps for both schemes.  We do not, however, include post-processing via structured prediction for any methods except Lift-CAM.  In \cref{fig:class-activation-results}(f), we give a table of metric scores.

In all of the supplied scenes, Lift-CAM emphasizes class-specific areas well.  For the scene in \cref{fig:class-activation-results}(i)(a), it correctly identifies the boundary between the highly rippled sand and the flat sandy regions.  For the more challenging scene in \cref{fig:class-activation-results}(i)(b), it highlights all three of the present seafloor classes.  It also manages to spot a small piece of man-made debris that is nestled within the rock field.  The scenes in \cref{fig:class-activation-results}(i)(c)--(d) illustrate that our framework, in conjunction with Lift-CAM, can handle variably sized classes.  It is capable of detecting large-scale objects, like the crashed airplanes.  It can also simultaneously detect much smaller ones, like the indentations from the targets impacting the seabed and likely what are either tilefish burrows or pits caused by hydro-physical processes.  Some spatial overemphasis can be observed, though.  The scenes in \cref{fig:class-activation-results}(i)(c) and (e) show that our framework can accommodate many non-contiguous class regions per image.  In each case, our framework has to overcome a variety of visual distractors, which it does rather successfully.  

The class heatmaps shown in \cref{fig:class-activation-results}(iii)(a)--(e) align well with our ground-truth segmentation maps.  Good seed cues are likely to be returned.  The average scores in \cref{fig:class-activation-results}(f) suggest this observation holds over the entire dataset for the super-class categories that we consider.  We ran a hundred Monte Carlo trials to obtain these averages.

In comparison, the {\sc CAN} fares poorly when using Grad-CAM.  Grad-CAM tends to grossly overemphasize the spatial class extents.  For the scene in \cref{fig:class-activation-results}(i)(a), the boundary for the flat sand class bleeds significantly into that of the rippled sand despite stark visual difference between the two.  This problem is more pronounced for the complex scenes in \cref{fig:class-activation-results}(i)(b) and \cref{fig:class-activation-results}(i)(e).  For the former, the heavily-rippled-sand region almost completely displaces that for the lightly rippled sand.  For the latter, the flat sandy class is incorrectly assigned prominence over the mixed-topology class.  Grad-CAM also misses several regions that Lift-CAM correctly uncovers.  It is unable to find many of the seafloor pits in \cref{fig:class-activation-results}(i)(c).  The marker tether near the crashed aircraft in the same scene is completely missed.  Grad-CAM also fails to identify the debris at the bottom right of \cref{fig:class-activation-results}(i)(d) and \cref{fig:class-activation-results}(i)(e).  We encounter similar issues in other procedures like Score-CAM and Ablation-CAM.  This observations hints that the activation-map inference, not necessarily the {\sc CAN}, is likely to blame for the mediocre qualitative performance.

The average scores in \cref{fig:class-activation-results}(f) indicate that the poor performance of Grad-CAM on these examples extends across the entire dataset.  Related class-map inference methods like Grad-CAM+ and Xgrad-CAM suffer from similar issues.  As before, we ran a hundred Monte Carlo trials to obtain these averages.

Our specified thresholds for the statistical tests are met for all of the experimental results in this section.  The scores reported by our framework are thus likely to be statistically significant compared to the scores from the alternative inference methods.  The same claim applies to the scores presented in the appendices.  It is therefore highly unlikely that the observed performance differences between Lift-CAM and Grad-CAM, Grad-CAM+, etc. occur due to random chance.

\vspace{0.2cm}\noindent {\small{\sf{\textbf{Activation-Mapping Results Discussions.}}}} Our experimental results indicate that the {\sc CAN} is capable of accurately recognizing classes in the sonar imagery.  The Lift-CAM procedure is able to infer high-quality class-activation maps, as a consequence.  In what follows, we describe some of the traits that contributed to the framework's success.  We also discuss some of the limitations of the comparative methods.

There are many reasons why the {\sc CAN}s do well, especially when compared to {\sc FCN}-like models for target detection.  Unlike standard {\sc FCN}s, the {\sc CAN}s do not rely solely on the top-most feature map for inference.  An over-reliance on top-level features can result in poor detection performance on regions that possess weak semantic information.  As well, the {\sc CAN}s characterize features at multiple scales.  This is crucial for sonar analyses, since the spatial class extents can vary widely depending on the class type.  {\sc FCN}s largely operate on only a single scale.  {\sc FCN} activation maps are also derived from fixed-size contexts, complicating the detection of class boundaries.  {\sc CAN}, in contrast, have high spatial consistency, which stems from their ability to learn image-contrast features well.  As hinted at by our experimental results, the {\sc CAN}s can infer both global contrast attributes and edge-preserving, local contrast features, both of which are combined to exclude non-class-specific regions.  We have found that incorporating features from multiple images improves performance when one or more of these contexts are inadequate for finding low-contrast targets.  This justifies our use of flow-based matching and adjustment networks.  The insertion of content-addressable memories into the {\sc CAN}s further increases performance, as it allows for it to store and recall rarely encountered class details.  We explore the benefits of using memories in \hyperref[secF]{Appendix~F}.

There are many other reasons why the {\sc CAN}s do well.  In addition to learning robust, multi-scale contrast filters, they adeptly removed nuisances.  They do not fixate on sonar speckle, despite it being prevalent throughout the imagery.  This is partly due to the use of local-global spectral pooling operations.  Our preferred pooling method downsamples the feature representations while preserving key details and removing random-phase irregularities.  Additionally, the encoder convolutional kernels are tuned in a way so that the shift between the complete-aperture and partial-aperture regions do not greatly impact the network features.  It is therefore possible to recognize well classes with low contrast and infer their approximate locations in the class-activation maps.

The way that we train our {\sc CAN}s also aids in their performance.  A reported disadvantage of Shannon cross-entropy loss is it makes pixel-wise-independent predictions.  The independence property can sometimes cause spatial discontinuities to form in the class-activation maps, which manifest as blurry boundaries.  We largely avoid this issue in our {\sc CAN}s.  This is because we employ an edge-sensitive sub-network composed of deep-parsing layers to identify pseudo class boundaries, based on high- and low-contrast edge features.  In doing so, we adaptively constrain the per-class spatial extents to those edges as much as possible.  This provides a capability similar to superpixels but without imposing regional homogeneity.  Allowing for regional heterogeneity is important, since the initial class-activation maps may not always be locally accurate.

Pre-training the {\sc CAN}s on natural imagery also contribute to their success.  As we discuss in \hyperref[secD]{Appendix~D}, using the ImageNet dataset stabilizes the early stages of the network to evaluate a variety of contrast-oriented features.  These features are important for beginning to understand the spatial class extents and are mostly independent of the image modality being considered.  The features also provide ample discriminative cues.  The semantic information encoded by the filters becomes increasingly complex in the later stages, leading to high classification performance.  Many of these filters are modified heavily when fine-tuning on the CSAS imagery.  Some are not, though.  We have found that the ImageNet-based initialization of the receptive fields aids the network in discovering semantic concepts that it would otherwise take a long time to emerge if training solely on the CSAS imagery.  Our studies in \hyperref[secE]{Appendix~E} demonstrate that there is a conspicuous improvement in quantitative and qualitative performance that comes with pre-training, a finding that has been echoed for many years \cite{GirshickR-conf2014a}.

There are a variety of representational factors that influence the quality of the class-activation maps.  As we note in \hyperref[secA]{Appendix~A}, independently processing the sub-apertures can introduce a significant amount of ambiguity and make detecting classes in certain classes very challenging.  Most seafloor types become wholly unrecognizable, for example, which preempts detecting them in the images.  Forming a multi-aperture representation and analyzing it, as we did here, is far more effective.  Care must be taken, though, in how we encode multi-aspect information.  Our preferred encoding scheme appears to yield the best performance, as we discuss in \hyperref[secB]{Appendix~B}.  Alternate encodings can significantly trail in performance, especially ones that predominantly emphasize reflective-only details.  These claims extend to all of the class-map inference procedures that we use, not just Lift-CAM.

One the reasons why Lift-CAM does well, when combined with our {\sc CAN}s, is that it does not suffer from issues that are normally encountered by alternate class-explanation methods.  These alternate methods typically combine multiple activation maps in a linear way to yield explanations of visual content.  Since the activation maps are formed deterministically, the coefficients for this aggregation solely dictate the explanation quality.  Many existing methods rely on heuristics for choosing the coefficients, though, which can yield subpar performance.  This can often occur because the assumptions that underlie the heuristics have no clear theoretical motivation.  The heuristics are also often specified in a way that do not account for any properties that an investigator would expect to see in a good explanation model.  Lift-CAM, in contrast, relies on a principled, unified measure of model-agnostic feature importance \cite{LundbergSM-coll2017a}.  This measure leverages three attributes, missingness, consistency, and local accuracy, for specifying an additive feature value and attribution process.  Strong theoretical guarantees for the formation of unique solutions are also available for this measure, which ensures that good linear coefficients will often be returned.

There are other issues with alternate class-activation methods.  Classical approaches rely extensively on gradient-derived information, which can be unreliable.  Gradients in deep networks can often diminish due to saturation \cite{GlorotX-conf2010a,MiglaniV-conf2020a}.  Using non-modified gradients can induce localization failures for class regions.  This is particularly problematic for both Grad-CAM \cite{ZhouB-conf2016a,SelvarajuRR-conf2017a} and Grad-CAM+ \cite{ChattopadhayA-conf2018a}.  The former solely determines the coefficients for the activation maps via the gradients over all activation neurons in a given map.  Grad-CAM+ \cite{ChattopadhayA-conf2018a} uses a similar process, albeit incorporating higher-order gradients that may be more sensitive to diminished gradients.  Due to the above concerns, gradient-free class-activation methods have risen to prominence.  Methods like Score-CAM \cite{WangH-conf2020a} and Ablation-CAM \cite{DesaiS-conf2020a} discern coefficient values by looking at the responses of perturbing various maps, which does not require the evaluation of gradients.  However, there are no formal guarantees that the perturbation of class maps will emphasize meaningful details that can be further exploited for segmentation.  XGrad-CAM \cite{FuR-conf2020a} overcomes this shortcoming somewhat, as it is designed to satisfy dual attribution axioms of sensitivity and conservation \cite{SundararajanM-conf2017a}.  These axioms are likely not sufficient to ensure that unique solutions will be formed, though \cite{ShapleyLS-coll1953a,YoungHP-jour1985a}.  Without the added constraint of continuity \cite{SundararajanM-conf2017a}, XGrad-CAM may also yield vastly different class-activation mappings for similar inputs, thus stymieing the consistent formation of good seed cues.

Lift-CAM does not suffer from the shortcomings of these other methods.  As such, Lift-CAM is well suited for relating spatial content with non-spatial supervision provided by the image-level class labels.

Model-based factors, beyond the choice of class-map inference scheme, also impact the reliability of class-activation maps.  In \hyperref[secF]{Appendix~F}, we conduct a comprehensive ablation study.  We find that all of the changes from a base network design yield meaningful performance increases.  The largest sources of improvement stem from the use of multi-scale features and the subsequent transformation, storage, and recall of them.  The size of the memory and the type of memory used to handle cross-image contexts matter too.  Our chosen memory architecture achieves the best performance out of the options that we consider.  Including more memory banks generally permits forming class-activation maps that provide better seed cues.  Increasing the memory size too much for the alternate architectures often leads to noticeable declines in both class-activation-map quality and segmentation-map quality.  These claims extend to all of the class-map inference procedures that we consider.

\phantomsection\label{sec4.2}
\subsection*{\small{\sf{\textbf{4.2.$\;\;\;$Semantic Segmentation Results and Discussions}}}}

We now demonstrate that the {\sc CREN} can effectively leverage the seed cues from the {\sc CAN} to semantic segment underwater imagery.  To provide context for our results, we compare against deep, supervised and deep, weakly-supervised networks that have been modified for our imaging-sonar modality.  We also show that our chosen superpixel regularization helps to guide the formation of good segmentation boundaries.

We provide substantial discussions to not only explain why our superpixel network does well, but also why our semantic segmentation network can return good class mappings.  We then discuss why the existing weakly-supervised segmentation networks fail to perform as well as our network. 

\vspace{0.2cm}\noindent {\small{\sf{\textbf{Training Protocols.}}}} We pre-train and fine-tune the {\sc CREN}s using ADAM-based back-propagation gradient descent with mini batches \cite{KingmaDP-conf2015a}.  We rely on the default parameters for ADAM.  We use a mini-batch size of 32 samples to bias against terminating in poor local minima \cite{HardtM-conf2016a}.

Pre-training of the {\sc CREN}s is done using the PASCAL VOC dataset \cite{LiY-conf2014a,EveringhamM-jour2010a}.  We then fit the {\sc CREN}s to our CSAS dataset to specialize the networks to imaging-sonar characteristics.  We comment on the benefits of pre-training in \hyperref[secE]{Appendix~E}.  Our results in this appendix show that our framework obtains state-of-the-art performance on the PASCAL VOC and the MS COCO Stuff \cite{CaesarH-conf2018a} datasets when compared to the best weakly-supervised semantic segmentation networks in the literature.  Our framework also yields results that are competitive against fully-supervised methods developed in the last few years.

We conduct multiple Monte Carlo pre-training simulations and fine-tuning simulations in the same way as we did for the {\sc CAN}s.  This includes how we determine when to stop training.  We additionally augment the batches via the same image transformations.

We compare our framework against many supervised and weakly-supervised semantic segmentation networks.  All of these have been designed to support natural imagery, not sonar imagery.  However, they can be easily adapted for the latter modality after retraining.  We do not augment these networks with the sonar-specific enhancements that we included in our framework to promote a fair comparison.

The alternate networks rely on distinct training methodologies.  We adopt the author-recommended protocols and hyperparameter values.  In cases where little to no information is provided about the training process, we use ADAM-based gradient descent with mini-batches.  We determine reasonable hyperparameter values using grid searches.  We limit the grid searches to a total of a thousand Monte Carlo runs per approach.  As with our framework, we pre-train the alternatives on the PASCAL VOC dataset and then fit to the CSAS dataset.  The same training process as our framework is used.  When reporting simulation results, we consider the average performance taken across multiple Monte Carlo trials.

We evaluate the quality of the inferred superpixels and the segmentation maps.  For the former, we use achievable segmentation accuracy (ASA), boundary recall (BR), and $f$ score (FS) \cite{StutzD-jour2018a}.  We consider the same number of superpixels for all of the methods to which we compare.  For the latter, we rely on the average $f$-measure (AFM) \cite{AchantaR-conf2009a}, mean-average precision (MAP) \cite{LiuN-conf2016a}, and the average intersection-over-union (IOU) \cite{ZhangD-conf2017a} metrics.  These metrics produce values in the range of zero to one, with higher values indicating better performance.  We average the scores across the Monte Carlo trials for the reported results.  We do not include score standard deviations due to the number of class super-categories that we evaluate.  We bin the scores from each class into six superclasses to provide a broad view of performance.  Reporting individual scores for each of the twenty classes that we consider complicates seeing the broader trends and where the difficult approaches struggle.

\vspace{0.2cm}\noindent {\small{\sf{\textbf{Assessment Protocols.}}}} We use the same protocols outlined in the previous section to assess the segmentation performance difference between our framework and the alternatives.

\vspace{0.2cm}\noindent {\small{\sf{\textbf{Experimental Results.}}}} Our {\sc CREN}s produce two intermediate responses, sets of inferred superpixel boundaries and non-regularized segmentation maps.  We first analyze the former.

\begin{figure}[t!]
   \hspace{-0.05cm}\includegraphics[width=6.6in]{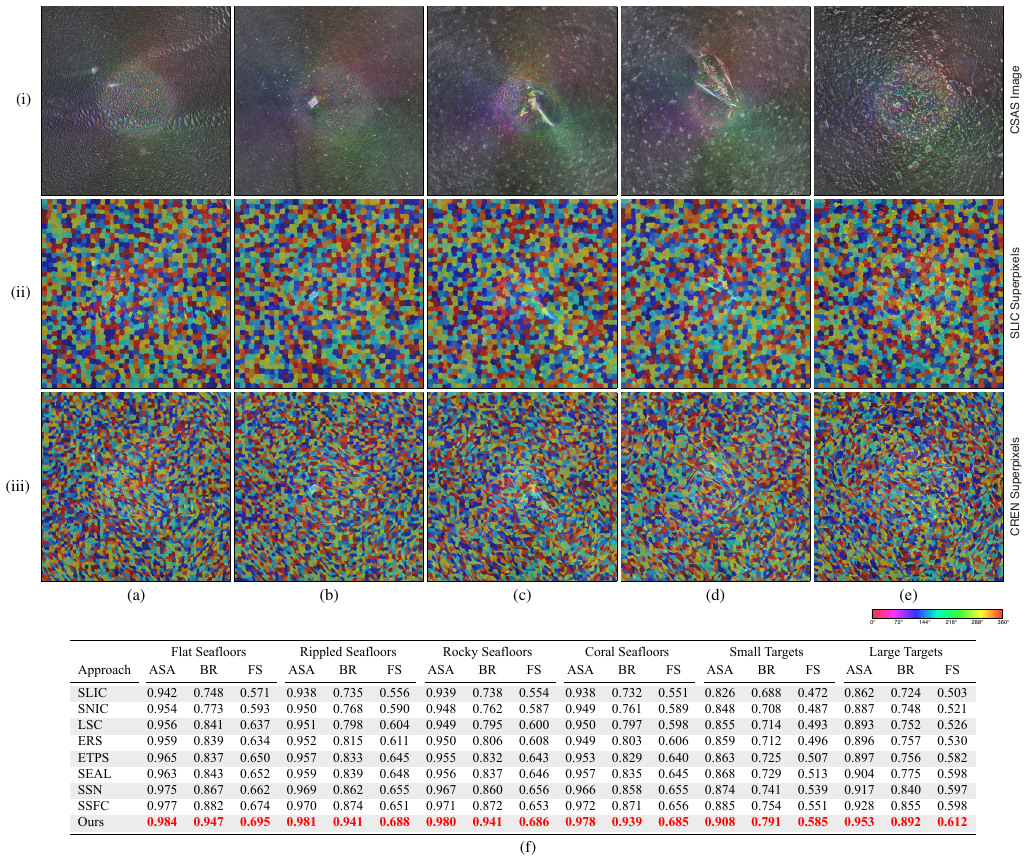}\vspace{-0.05cm}
   \caption[]{\fontdimen2\font=1.55pt\selectfont Superpixel results.  In row (i), we provide examples of five underwater scenes.  The scenes in columns (a) and (e) have mixed bottom types.  The scenes in columns (b), (c), and (d) have small-, moderate-, and large-scale targets, respectively.  In rows (ii) and (iii), we show the corresponding SLIC superpixels \cite{AchantaR-jour2012a} and CREN-based superpixels, respectively.  The latter are far more effective at identifying class transitions.  (f) A table of achievable segmentation accuracy (ASA), boundary recall (BR), and $f$ score (FS) \cite{StutzD-jour2018a}, all of which indicate that our adaptive approach offers either near-equal or better performance than competing methods.  Here, we compare against SLIC \cite{AchantaR-jour2012a}, SNIC \cite{AchantaR-conf2017a}, LSC \cite{LiZ-conf2015a}, ERS~\cite{LiuMY-conf2011a}, ETPS \cite{YaoJ-conf2015a}, SEAL \cite{TuWC-conf2018a}, SSN \cite{JampaniV-conf2018a}, and SSFC \cite{YangF-conf2020a}.  We recommend that readers consult the electronic version of the paper to see the full image details.}\vspace{-0.4cm}
   \label{fig:superpixel-results}
\end{figure}

Example superpixel maps are provided in \cref{fig:superpixel-results}(ii)(a)--(e) and \cref{fig:superpixel-results}(iii)(a)--(e) for the scenes in\\ \noindent \cref{fig:superpixel-results}(i)(a)--(e).  The first set of results are for SLIC superpixels while the second are for the deep, adaptive superpixels from the {\sc CREN}s.  We provide a table of metric scores in \cref{fig:superpixel-results}(f).

Our network initially returns a large number of superpixels that are subsequently merged by analyzing feature-histogram uncertainty.  The results that we present in \cref{fig:superpixel-results}(iii)(a)--(e) are before merging occurs.  Once merged, the final number of superpixels is nearly equivalent to the amount that we used for SLIC.  The scores that we list in \cref{fig:superpixel-results}(f) are obtained after merging.

For all of these underwater scenes, our unsupervised-trained superpixel network yields contrast-conscious superpixels that also adhere to class boundaries.  Our network, for \cref{fig:superpixel-results}(i)(a), correctly groups pixels belonging to the lightly-rippled sand and separates them from both the more heavily-ripped sand and flat sand.  It also identifies and isolates the scene haze, near the middle left of the scene, that arises due to speed-of-sound changes.  The scenes in \cref{fig:superpixel-results}(i)(b)--(d) suggest that our network can readily address small- to large-scale targets that may have highly irregular shapes.  It accurately delineates the man-made debris in \cref{fig:superpixel-results}(i)(c), including a majority of the smaller fragments.  For \cref{fig:superpixel-results}(i)(d), it correctly traces the shipwreck hull, mask, and other remains.  It even is sensitive to the small mounds of sand that are scattered throughout \cref{fig:superpixel-results}(i)(c)--\cref{fig:superpixel-results}(i)(d).  The final scene that we consider, \cref{fig:superpixel-results}(i)(e), is the most challenging, which is due to the high rock density.  Nevertheless, our network distinguishes between nearly every rock and the flat sandy seafloor.  Only a few rocks, whose dimensions are below our chosen superpixel size threshold, are missed.  This behavior was also encountered in \cref{fig:superpixel-results}(i)(b) where some of the small, white circular patterns, which are likely marine organisms, were not captured by individual superpixels.  Reducing the superpixel-size threshold would allow our network to assign such regions to independent superpixels without adversely impacting the remainder of the quantized representation.

Alternate methods, especially those that are non-deep, tend to do poorly in comparison.  Many of them rely on regular-grid initializations that do not yield promising class-sensitive pixel groupings except when the grid spacing is tight.  For instance, in scene \cref{fig:superpixel-results}(i)(a), none of the present classes are segmented well by the SLIC superpixels.  Only a few superpixels are substantially modified to characterize the sand ripples.  Even then, only the ridges are emphasized well, which makes defining the corresponding class boundaries extremely difficult.  SLIC fares better for the scene in \cref{fig:superpixel-results}(i)(b), largely due to the overall simplicity of the environment and the well-defined target contours.  It correctly identifies the square, man-made target and assigns it to a single superpixel.  More complex target types stymie these types of shallow methods, though, even when encountering high-contrast edges.  Much of the debris in \cref{fig:superpixel-results}(i)(c) would be incorrectly assigned to a non-target class, though, which would complicate inference within our segmentation network.  The intricate, low-contrast edges present in \cref{fig:superpixel-results}(i)(d) are even more of a challenge.  SLIC, along with other approaches, fail to to adapt the pixel affinities to account for them.  Only small parts of the hull would be conclusively labeled as target-like regions when relying on this regularization representation.  For \cref{fig:superpixel-results}(i)(e), SLIC is unable to separate most of the rocks from the seafloor.  It is only when the number of superpixels increases substantially that SLIC returns nearly-class-sensitive pixel groupings.  However, larger amounts of superpixels do not yield large qualitative improvements for the groupings in the other scenes.

The averaged scores presented in \cref{fig:superpixel-results}(f) indicate that the behavior observed for our network yields high quantitative performance for the full imaging-sonar dataset.  Our network is highly competitive against supervised-trained deep superpixel networks and far outpaces unsupervised, non-deep methods.  It thus should yield a promising regularization for segmentation.  We ran a hundred Monte Carlo trials to obtain these averages.

\begin{figure}[t!]
   \hspace{0.0cm}\includegraphics[width=6.6in]{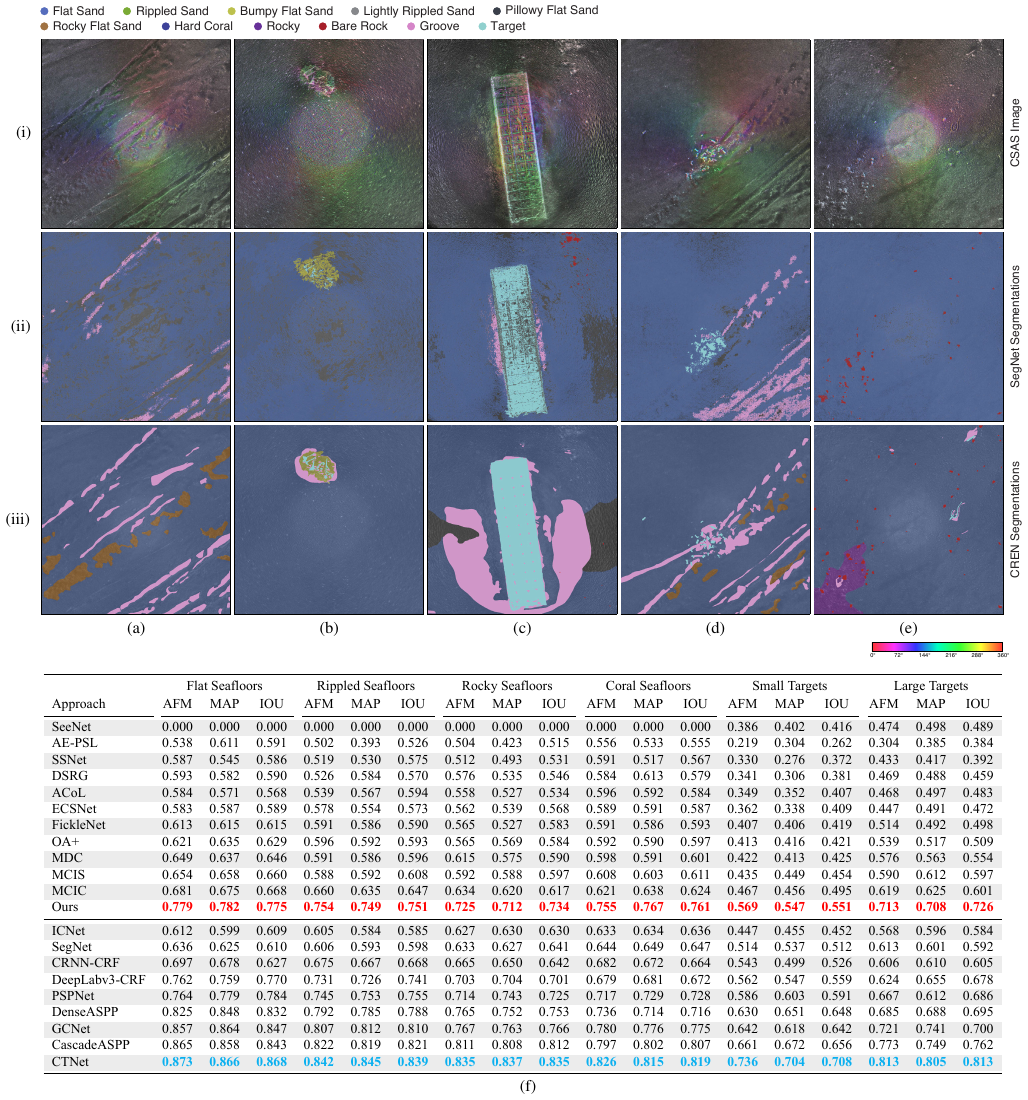}\vspace{-0.05cm}
   \caption[]{\fontdimen2\font=1.55pt\selectfont Semantic segmentation results.  In row (i), we provide examples of five underwater scenes.  The scene in column (a) has no targets.  Those in columns (b) through (e) have targets of varying size and complexity.  Each scene contains at least two bottom types.  In rows (ii) and (iii), we show the maps returned by {\sc SegNet} and our {\sc CREN}, respectively.  We use the class colormap provided at the top of the figure.  We use the same color for all target classes so as to reduce the color palette.  However, our network distinguishes between ten unique target types.  (f) A table of average $f$-measure (AFM), mean average precision (MAP) and average intersection-over-union (IOU) scores.  Higher values are better.  The table is divided into two parts.  The top part covers weakly-supervised methods.  The bottom reports results for supervised methods.  For the former type of methods, we compare against {\sc SeeNet} \cite{HouQ-coll2018a}, {\sc AE-PSL} \cite{WeiY-conf2017a}, {\sc SSNet} \cite{YuZ-conf2019a}, {\sc DSRG} \cite{HuangZ-conf2018a}, {\sc ACoL} \cite{ZhangX-conf2018a}, {\sc ECSNet} \cite{SunK-conf2021a}, {\sc FickleNet} \cite{LeeJ-conf2019a}, {\sc OA+} \cite{JiangP-T-conf2019a}, {\sc MDC}~\cite{WeiY-conf2018a}, {\sc MCIS} \cite{SunG-conf2020a}, and {\sc MCIC} \cite{FanJ-jour2023a}.  For the latter, we compare against {\sc ICNet} \cite{ZhaoH-conf2018a}, {\sc CRNN-CRF} \cite{ZhengS-conf2015a}, {\sc SegNet} \cite{BadrinarayananV-jour2017a}, {\sc DeepLab3-CRF}~\cite{ChenLC-conf2018a}, {\sc PSPNet} \cite{ZhaoH-conf2017a}, {\sc DenseASPP} \cite{YangM-conf2018a}, {\sc GCNet} \cite{CaoY-conf2019a}, and {\sc CTNet} \cite{LiZ-jour2022a}.  Based upon a cursory inspection, our network returns qualitatively better results that are less susceptible to changes in aspect coverage than early, fully-supervised methods like {\sc SegNet}.  All of metric scores indicate that our {\sc CREN} outperforms other weakly- supervised deep networks.  It is also competitive against fully-supervised deep networks.  We recommend that readers consult the electronic version of the paper to see the full image details.}\vspace{-0.4cm}
   \label{fig:semantic-segmentation-results}
\end{figure}

The ability to form superpixels that adhere to observed class boundaries enables our {\sc CREN}s to produce segmentation maps with few errors.  We offer several examples in \cref{fig:semantic-segmentation-results}(iii)(a)--(e) to demonstrate this behavior for the scenes given in \cref{fig:semantic-segmentation-results}(i)(a)--(e).  In \cref{fig:semantic-segmentation-results}(ii)(a)--(e) we provide corresponding responses from {\sc SegNet}.

In each of these scenes, our {\sc CREN}s return promising segmentation maps after transforming the {\sc CAN}-supplied class-activation maps.  Our network detects and outlines many of the anchor drag marks for \cref{fig:semantic-segmentation-results}(i)(a), including those that are rather faint.  In \cref{fig:semantic-segmentation-results}(i)(b), it spots the partly-buried debris.  Our network correctly isolates the full boundaries for the sunken barge in \cref{fig:semantic-segmentation-results}(i)(c).  It also, for that scene, correctly distinguishes between the flat and rippled seafloors from the large-scale impression that the barge made when hitting the seafloor.  All of the spent and unexploded ordnance in \cref{fig:semantic-segmentation-results}(i)(d) has been found by our network, despite their incredibly small size and occasional loss of complete aspect coverage.  As with the second scene, all of the grooves made by either hydrodynamic processes or high-powered anchors are highlighted by our network for this scene.  Scenes like \cref{fig:semantic-segmentation-results}(i)(e) are among the most difficult for any segmentation network.  Here, however, our {\sc CREN}s consistently locate the man-made objects despite them not possessing prominent acoustic returns.  It correctly discriminates between some of the larger basalt rocks and the ones that we would consider to be the basis of a rocky seafloor.  For the examples presented in \cref{fig:semantic-segmentation-results}(i)(a)--(e), the correct bottom types are reported.  Moreover, the class boundaries returned by our network match the ground truth well in all cases.

Our {\sc CREN}s do not have the benefit of direct supervision to guide their training.  Nevertheless, they qualitatively outperform many types of supervised networks, such as {\sc SegNet}.  As shown in \cref{fig:semantic-segmentation-results}(ii)(a), {\sc SegNet} misses many of the well-defined and less shallow grooves.  The lack of a principled regularizer impacts its ability to obey the boundaries for these seafloor features.  Its inability to remove speckle noise does too.  The network also consistently mislabels the flat sand not only in this scene, but also in \cref{fig:semantic-segmentation-results}(ii)(b)--(e).  For \cref{fig:semantic-segmentation-results}(ii)(b)--(c), the network has difficulties in specifying the proper extents for the present targets.  It also does not extract strong semantic signatures for labeling the indentations in these two scenes.  {\sc SegNet} ignores many of the small-scale targets in \cref{fig:semantic-segmentation-results}(ii)(d)--(e) despite their prominent anisotropy.  It erroneously infers that many of the strongly isotropic acoustic returns that belong to the seafloor, in \cref{fig:semantic-segmentation-results}(ii)(d), are spent ordnance.  The network additionally has difficulties in spotting many of the classes with weak discriminative cues, like the small mounds of rocky sand in \cref{fig:semantic-segmentation-results}(i)(a) and \cref{fig:semantic-segmentation-results}(ii)(d).  The strong amount of intra-class variability observed for other underwater topologies, which are not shown in these examples, adversely impacts the performance of {\sc SegNet}.

We provide a tabular summary of performance metrics in \cref{fig:semantic-segmentation-results}(f).  We ran a hundred Monte Carlo trials to obtain these averages.  The average scores quantitatively demonstrate that our network outperforms {\sc SegNet} and many other supervised-trained networks.  Likewise, our network outperforms all of the weakly-supervised networks to which we compare.  Many of these alternates are the current state of the art in the literature for natural imagery.

For each of the experimental results in this section, our specified thresholds for the statistical tests are met.  The scores reported by our framework are thus likely to be statistically significant compared to the scores from the alternative inference methods.  The same claim applies to the scores presented in the appendices.  It is therefore highly unlikely that the observed performance differences between our framework and the alternatives occur solely due to random chance.

\vspace{0.2cm}\noindent {\small{\sf{\textbf{Superpixel Results Discussions.}}}} One of the factors that contributed to the success of the {\sc CREN}s is that our {\sc USN}s offer the current-best superpixel regularization performance.  

Below, we cover some of the shortcomings of existing superpixel methods.  We then discuss how our {\sc USN}s avoids them.

A variety of superpixel approaches have been proposed over the years \cite{StutzD-jour2018a}.  Some of the most popular approaches, like {\sc SLIC} \cite{AchantaR-jour2012a}, {\sc SNIC} \cite{AchantaR-conf2017a}, {\sc LSC} \cite{LiZ-conf2015a}, and others, leverage unsupervised mechanisms for discerning how to group spatially contiguous pixels based on these features.  Alternatives, such as {\sc NC} and {\sc ERS} \cite{LiuMY-conf2011a}, impose image-agnostic initial pixel-affinity assignments and iteratively adjust those assignments to adapt to local image content.  The commonality amongst them is that they process fixed, hand-crafted features, which may not be suitable for non-standard imaging modalities.  As we show in our experiments, this trait, along with how the superpixel regions are adjusted, can often impede segmentation.  Many of the superpixel boundaries do not align well with the observed class boundaries, especially when only a few superpixels are used.  This occurs despite high-contrast edges being present, especially for the target classes.  It is only when a higher number are used, and the average superpixel size drops, that the boundaries start to be obeyed well and the relevant metrics improve.  In this situation, though, the superpixels are almost the same size as an individual pixel, so the regularization benefits offered by the former are minimal.  Segmentation performance thus would likely begin to decline.

Deep-superpixel networks, like {\sc SSN} \cite{JampaniV-conf2018a} and {\sc SSFC} \cite{YangF-conf2020a}, can partly overcome these flaws.  The former relaxes the nearest-neighbor grouping constraints found in {\sc SLIC} to yield a differentiable assignment process that can be inserted into end-to-end-trainable deep networks.  However, {\sc SSN}-type methods typically introduce additional issues.  The assignment process it uses is iterative.  No formal convergence guarantees are provided.  It may hence not sufficiently converge to class-sensitive superpixels within a reasonable period.  Additionally, the number of superpixels must be manually specified for {\sc SSN}-like methods.  An improper amount can cause either under- or over-segmentation that degrades the quality of the final semantic maps.  We encountered this behavior in our simulations, which partly explains the performance gap between {\sc SSN} and our method.  Both of these concerns are partially addressed by {\sc SSFC}.  {\sc SSFC} combines the feature-extraction and superpixel-segmentation assignments into a single step, which is a byproduct of using one-hot encoding processes.  Depending on the choice of loss function, investigators may not need to a priori define the superpixel count.  Rather, it can be determined automatically using the extracted features.  Strong supervision, in the form of labeled affinity maps, is needed for {\sc SSN} and {\sc SSFC}, along with other methods like {\sc SEAL} \cite{TuWC-conf2018a}, to work, though.  

Our {\sc USN}s have multiple traits that are beneficial for forming superpixels.  Unlike {\sc SSN}-style methods, our {\sc USN}s implement a one-hot group encoding that yields responses which are incredibly close to the ones found by using iterative optimization.  This encoding additionally takes advantage of the local-global spectral features.  Disparate spatial information can hence be efficiently aggregated, allowing for contrast-sensitive groupings over large regions.  We also adaptively merge superpixels, according to an information-theoretic criterion.  This process has the benefit of not only reducing the number of superpixels for subsequent segmentation inference, but also sidesteps the need to a priori specify the number of superpixels.  Since the merging process utilizes local feature-histogram uncertainty, class boundaries are typically very well respected by the superpixels.  Perhaps one of the biggest advantages of our superpixel network, though, is that the {\sc USN}-derived features are obtained directly from the imagery.  Hand-crafted features, like those used in {\sc SLIC} and {\sc SNIC}, may not be appropriate for all imaging modalities.

As suggested by its name, our {\sc USN}s eschew labeled maps to guide the superpixel formation.  Despite this, they perform either on par or better than {\sc SSFC}, depending on the chosen comparison metric.  Our {\sc USN}s are successful because they first learn an adaptive clustering feature space from which a sufficient number of seed cues are positioned to align with observed image content.  The network's pixel-assignment loss then exploits the shifted cues to assign pixels to nearby seeds.  This loss also biases against creating large superpixels, which may not be class-boundary sensitive.  Alternate seed-generation processes that are unsupervised, like selecting the centers of a uniformly-spaced grid, may not be sufficient to offer comparable superpixel generations.  Our {\sc USN}s additionally reshape the back-propagated gradients during learning.  This weighting process ensures that the network does not overfit to a particular set of images and thus preempts the formation of a poor clustering feature space.

\vspace{0.2cm}\noindent {\small{\sf{\textbf{Segmentation Results Discussions.}}}} The ability for our {\sc CREN}s to exploit high-quality superpixels is just one facet of why it does well.  In \hyperref[secF]{Appendix~F}, we conduct a comprehensive ablative study to evaluate the other factors.

We find that one of the largest contributors to performance is the ability to learn a regularized intermediate feature representation.  Since our {\sc CAN} typically supplies accurate seed cues, it becomes possible to re-weight the {\sc CREN} features so that those from distinct classes are significantly disjoint.  This behavior reduces many types of segmentation errors that are encountered in the baseline version of the network.  Another major contributor is the use of multi-scale, local-global convolution.  These layers provide a parameter-efficient way of considering both large and small receptive fields.  The large receptive fields help the {\sc CREN} aggregate far-away features, which aids when considering classes with complicated patterns that span much of the imagery.  The expansive views offered by these receptive fields are biased by the localized information from the smaller receptive fields.  The latter type ensure that neighboring semantic details are not completely ignored.  They help in situations where we have many class transitions over a small region.  Without the interaction of both, performance tends to suffer.  The spectral average pooling layers that we incorporate offer another way to increase the receptive field size.  However, unlike traditional pooling operations, they preserve informative feature content well, which improves segmentation performance.  This occurs because the pooling process truncates higher-frequency content, which tends to encode noise \cite{TorralbaA-jour2003a}.

A hallmark of our {\sc CREN}s is that they store, recall, and transform contextual details from multiple images.  This allows the network to remember rare content without the need for a great many convolutional filters.  It also augments the feature richness for frequently-observed classes in related sets of images.  As we note in \hyperref[secF]{Appendix~F}, the type of content-addressable memory that we use in the {\sc CREN} plays a major role in the effectiveness of multi-image context integration.  Our universal recurrent memories offer an efficient way to resolve problem of memory depth versus memory resolution, since the product of the memory depth and resolution is constant \cite{deVriesB-coll1991a,deVriesB-coll1991b}.  The memories within the {\sc CREN} can hence be grown to store arbitrarily large amounts of features while facilitating a nearly monotonic improvement in segmentation performance.  Existing memory architectures are unable to match this capability.  Increasing the number of memory cells in these alternatives can lead to severe overfitting, which reduces segmentation performance.  Additionally, alternate, non-content-addressable memories may be unable to effectively recall content that it observed earlier during training.  This is because information decays by a certain rate when propagating through time in these architectures.  While small decays can be used to potentially counteract this phenomenon, using them causes smoothing across time and leads to a heavy mixing of past content.  This too impacts segmentation performance, since informative features become jumbled with those that are not.  Those alternatives that are content addressable have similar depth-versus-resolution issues, since the read content and the query content often share related information.  

Another decisive performance factor is the ability for the {\sc CREN} to simultaneously segment multiple related images.  Often, multiple images of nearby regions are captured, and, in each of them, a slightly different portion of the scene can be resolved better.  By co-segmenting all of the images, our {\sc CREN}s can reduce the segmentation uncertainty.  This functionality is made possible by incorporating our flow-map network.  This network can infer dense and accurate pixel-level correspondence maps for pairs of images, which facilitates the transfer of labels between them.  Such label transfer and integration is highly effective at dealing with a loss of complete-aperture and partial-aperture coverage in our imaging-sonar modality.  As we show in \hyperref[secE]{Appendix~C}, the segmentation improvement is not negligible.  Including more images in the co-segmentation process tends to raise the average {\sc CREN} performance.

Several of the benefits enjoyed by the {\sc CAN} also extend to the {\sc CREN}.  The combined use of spectral features and spectral pooling mitigates the impact of sonar speckle on segmentation.  Likewise, pre-training on either the PASCAL VOC or the MS COCO Stuff datasets improves both the convergence rate and the segmentation quality for sonar imagery.  We explore this latter topic further in \hyperref[secE]{Appendix~E} and \hyperref[secF]{Appendix~F}.

The combination of the above factors allowed our framework to outperform all of the other weakly-supervised networks that we considered.  The most competitive alternative, {\sc MCIC} \cite{FanJ-jour2023a}, significantly trails in performance, despite also leveraging multi-image contexts.  The observed difference partly stems from its use of first-in, first-out queues as external memories.  While such queues are extended by the authors to be class sensitive, they are highly insensitive to the content of the input imagery.  They can hence mix highly disparate features that do little to enhance segmentation quality.  The authors of \cite{FanJ-jour2023a} allude to this in their discussions, as they report that performance saturates quickly and eventually drops for large queue sizes.  Based on our analyses, the performance gap also occurs because the {\sc MCIC} network relies on a fairly shallow process for integrating the multi-image contexts.  Only a weighted linear aggregation is used, which may not be sufficient for handling rare class details.  It may also have difficulties in handling commonly observed classes.  Our network, in contrast, leverages an adaptive, non-linear feature combination.  It is thus has a better potential to iteratively extract and transform the intermediate representations in a way that benefits segmentation.  All of the above components that we mentioned, which are not present in {\sc MCIC}, further widen the gap to our {\sc CAN}-{\sc CREN} framework.  Including some of these within {\sc MCIC}, like the spectral features, yields improvements.  Without an effective memory module, though, this enhanced {\sc MCIC} inevitably lags.

The other multi-context network for semantic segmentation, {\sc MCIS} \cite{SunG-conf2020a}, also has detractions.  Its predecessors do too \cite{ShenT-conf2017a}.  Foremost, {\sc MCIS} only considers the problem of dual-image co-attention for segmentation.  Without significant extensions, their network could not handle comparing arbitrary sets of images as in our network.  This severely constrains the amount of information that can be shared.  Even if the network topology was changed to address this issue, then the corresponding loss function would not scale well.  This is because it views features between unmatched classes as noise versus as discriminative details.  Either an image or a set of images would need to overwhelmingly belong to a single class for the loss to effectively transfer segmentation information.  It is difficult to adequately sample from a large enough set of images to gather enough related cross-image contexts.  Performance can even fall if poor matches are consistently returned.  Secondly, approaches like {\sc MCIS} utilize a two-stage training procedure.  As a part of this training process, a portion of the network is trained to generate class-specific pseudo-masks while the remainder converts those masks into segmentation maps.  The lack of an end-to-end trainable network limits the transfer of supervisory cues from one part of the network to the other.  For supervised-trained networks, the separated training regimes typically does not yield much, if any, adverse changes in generalization capability.  For weakly-supervised networks it does, though.  The processes of feature extraction and transformation are decoupled in a non-end-to-end-trained case, which limits the semantic content in later network stages.

There are other issues with the remaining weakly-supervised networks.  Works like {\sc DSRG} \cite{HuangZ-conf2018a}, {\sc MDC} \cite{WeiY-conf2018a}, and {\sc FickleNet} \cite{LeeJ-conf2019a} leverage expansion-based processes to iteratively grow class-affinity maps obtained from a classifier network.  {\sc DSRG} uses a shallow approach to do this, though.  There is also no estimation of classifier uncertainty to assess the quality of the activation mappings, which can yield poor seed cues.  The processes of classification and segmentation are also completely independent.  The supervision, albeit weak, used in the second half of the network has no influence on the mappings that are formed within the first part, which can impact performance.  {\sc MDC} shares these latter two shortcomings, as does {\sc OAA}+~\cite{JiangP-T-conf2019a}.  {\sc MDC} attempts to partially overcome the first issue by estimating multiple activation mappings, each with different receptive field sizes.  {\sc OAA}+, instead, looks at how the affinities change across the maps.  {\sc FickleNet} \cite{LeeJ-conf2019a} uses random feature selection in an attempt to understand the coherence of each location in the feature maps.  Promising regions that are discriminative can hence be identified.  These three approaches have the risk of over-estimating the pixel affinities, though.  They are hence using a poor substitute for our uncertainty-quantification measure.  Our measure assesses exactly what the {\sc CAN} either does or does not know and biases the ensuing seed cues accordingly.  Additionally, all of these networks leverage Grad-CAM to infer the class-activation mappings.  As our experiments indicate, Grad-CAM tends to provide boundary-insensitive mappings that complicate segmentation.

Erasing-style methods, such as {\sc AE-PSL} \cite{WeiY-conf2017a}, {\sc ACoL} \cite{ZhangX-conf2018a}, {\sc ECSNet} \cite{SunK-conf2021a}, and {\sc SeeNet} \cite{HouQ-coll2018a} can offer a more appealing way of expanding activation maps than some region-growing networks.  They mainly operate by identifying an initial set of class-specific regions, inhibiting the class network from activating in those areas, and then finding alternate regions that could belong to a given class.  This forces the classifier network to learn alternate cues in an attempt to uncover the full spatial extents.  In early methods, though, this may not occur.  For instance, {\sc AE-PSL} \cite{WeiY-conf2017a} uses the original CAM approach \cite{ZhouB-conf2016a}, not Grad-CAM \cite{SelvarajuRR-conf2017a}, to infer activation maps.  CAM constrains the network topology, which can adversely effect performance.  CAM also has difficulties in detecting large-scale objects.  Several rounds of erasing would be required to ensure that undetected regions of such objects are not prematurely judged as belonging to the background.  If not enough rounds are considered, then the localization quality of the attention generator can be degraded.  {\sc SeeNet} \cite{HouQ-coll2018a} partly corrects this latter concern, as it utilizes dual self-erasing strategies for learning object and background cues.  However, {\sc SeeNet} fails to scale to our semantic segmentation dataset since the background class can change.  There is no easy way to extend their complementary attention generator to handle arbitrary numbers of background classes.  {\sc ACoL} \cite{ZhangX-conf2018a} attempts to address the former issue of {\sc AE-PSL}, namely, the use of CAM.  The authors in \cite{ZhangX-conf2018a} propose a one-hot activation-mapping procedure that they combine with erasing.  They show that it better captures integral object regions than CAM.  Without knowledge of complementary cues, though, it is often possible for the {\sc ACoL} object activations to spread to undesirable regions.  In our framework, the combination of uncertainty quantification and feature regularization often preempts this from occurring.

\phantomsection\label{sec5}
\subsection*{\small{\sf{\textbf{5.$\;\;\;$Conclusions}}}}\addtocounter{section}{1}

In this paper, we propose a deep-network framework for weakly-supervised semantic segmentation of CSAS imagery.  Our work represents the first semantic segmentation approach for this imaging-sonar modality.

Our framework relies on convolutional, memory-based sub-networks.  The first sub-network, the {\sc CAN}, extracts multi-scale, local-global features to identify seafloor and target classes in a given image.  From its classifier responses, class-activation mappings are formed.  We use several heuristics to improve the quality of the mappings before cascading them into another network, the {\sc CREN}.  Like the {\sc CAN}, the {\sc CREN} extracts multi-scale local-global features.  These features are used to guide the selection of low-uncertainty spatial seed cues for each of the observed classes.  Pixels with similar features to the seed cues are assigned the same labels, yielding initial segmentation maps.  The {\sc CREN} imposes a large-margin separability constraint on the features, which often reduces errors during the initial map inference and limits error propagation in later stages.  The maps are then progressively upsampled and refined by the {\sc CREN}.  A third sub-network, the {\sc USN}, generates adaptive super-pixels that help the segmentation maps obey local and global changes in texture, which typically correspond with class boundaries.  

All of these sub-networks can be connected and trained end-to-end, which we do in this paper.  In doing so, they can leverage as much information as possible from the limited supervision that is provided.

One of the most important aspects of our framework is that it extensively leverages memories to improve segmentation quality.  More specifically, convolutional, content-addressable memories are located in each of the sub-networks to store class-specific details mined during both training and testing.  For new images presented to the framework, the memories are queried to find relevant entries.  Returned matches are integrated with the features found in the main network pathways.  This has the effect of incorporating supplementary information from previous images, which can aid in identifying seafloor and target classes that have highly variable visual characteristics.  Such details can also mitigate the selection of spurious seed cues in the early stages of our framework and better identify the true extents of rarely-encountered classes in the later stages.  The type of memories we employ facilitate these behaviors.  Our universal recurrent memories are nearly-linear models that are efficient to populate and update.  They possess simple reading and writing mechanisms that remain stable across time.  They are also easily extensible to handle any number of entries without a loss of precision.  Alternatives, like neural Turing machines, are more cumbersome to use and yield little to no benefits, in contrast, for our study.  Other types of memory modules similarly impede multi-image context mining for segmentation.

Our experimental results showcase the capabilities of our framework.  When pre-training on natural-image benchmark datasets, our framework obtains better semantic segmentation performance than over eighteen weakly-supervised deep networks.  The performance gap of more than ten percent to the second-best network is statistically significant.  This is due to the number of tests that we conduct.  Our framework also closes the performance gap to the best fully-supervised methods when pre-training on natural imagery.  When fine-tuning to sonar imagery, our network outperforms all of the twenty-five weakly-supervised methods to which we compare.  It performs nearly on par with over ten fully-supervised methods, including those that are currently the state of the art.  As before, these results are statistically significant.  We provide detailed ablation studies to illustrate the impact of each component on the overall segmentation quality.  For some components, we also provide comparative studies to demonstrate that our choices yield the best results.  For instance, we show that our deep superpixel process yields state-of-the-art performance.  

\setcounter{figure}{0}
\begin{wrapfigure}{r}{0.7\textwidth}
   \vspace{-0.2cm}\hspace{0.1cm}\includegraphics[width=4.5in]{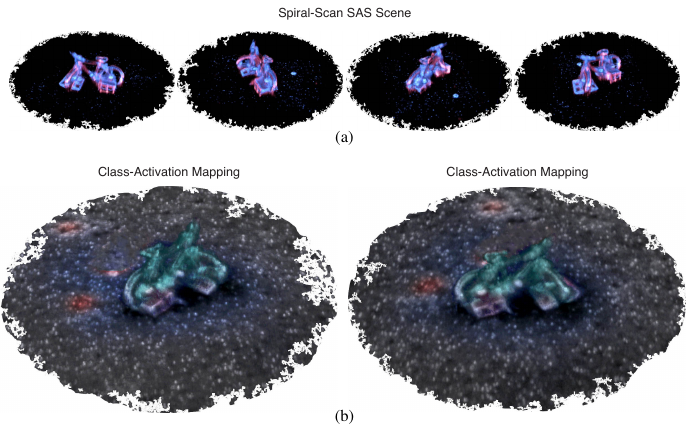}
   \caption[]{\fontdimen2\font=1.55pt\selectfont A transfer-learning application of our semantic segmentation scheme.  Here, we have taken a {\sc CAN} trained on circular-scan SAS imagery, a spatial modality, and used it to infer class-activation mappings for spiral-scan SAS (SSAS) imagery, a volumetric modality \cite{MarstonTM-jour2016a}.  The class-activation mapping highlights well the relevant parts of the scene.  It correctly delineates the voxels corresponding to the bicycle, concrete blocks, and flat seafloor.  (a) A turn-table view of the dominant high- and low-frequency content in the volumetric scene. (b) The class-activation mapping returned by our {\sc CAN}.  Pink hues correspond to debris, cyan hues to targets, blue hues to flat, sandy seafloors, and red hues to rocky elements.  In (b), we show two views of the sonar volume.\vspace{-0.1cm}}
   \label{fig:volumetric-semantic-segmentation-results}
\end{wrapfigure}

Although our weakly-supervised framework lags in performance compared to fully-supervised methods, it makes up for this gap in a crucial way.  That is, our approach only requires only global, image-level labels for training.  Such labels are quick and mostly easy for human annotators to specify.  Only a few seconds is typically needed to handle a single image.  Local, pixel-level labels are significantly more time intensive to obtain, especially for the high-resolution imagery that we consider.  Anywhere from three to ten minutes may be needed for labeling an image depending on the number of classes present and the scene complexity.

Our emphasis in this paper has been on segmentation for spatial sonar imagery.  Our framework can, with no modifications, be suitable for volumetric sonar modalities.  In many cases, our framework can leverage image-based and bathymetric-based details to produce relatively accurate segmentations without any additional fine-tuning.  We provide an example of this in \cref{fig:volumetric-semantic-segmentation-results}.  We will show, though, in our future work that fine-tuning significantly improves performance.  In our future work, we will also demonstrate that training on the sonar-image sub-apertures from the circular-scan case facilitates transfer learning to the side-scan case.  Little to no re-training may therefore be necessary to achieve good detection and segmentation performance for the latter modality, regardless of the target pose.  This assumes that acoustic shadows and other environmental phenomena do not dominate the imagery, however.

A downside to our framework is that a limited number of seed cues may not provide a sufficient amount of guidance to make segmentation performance truly competitive against fully-supervised approaches.  Other weakly-supervised segmentation methods also appear to share this trait.  In our future work, we will investigate a two-part approach to, hopefully, overcome this issue.  First, we will collect and utilize a much larger dataset.  Many more samples will likely allow our framework to better describe intra-class variability and hence improve segmentation performance.  We will also further tune our memory layers to take advantage of a larger sample set.  Second, we will investigate additional uses of our information-theoretic uncertainty measure to help guide the selection of meaningful training samples and ensure that they are weighted accordingly.  It is plausible that samples with useful characteristics are overwhelmed by those that provide little to no performance benefit.  Our framework may thus not be using the full gamut of supervision that is available.  Alongside these amendments, we will include explicit attention-focusing mechanisms.  

\setstretch{0.95}\fontsize{9.75}{10}\selectfont
\putbib
\end{bibunit}

\clearpage\newpage
\begin{bibunit}
\bstctlcite{IEEEexample:BSTcontrol}
\setstretch{1.15}\fontsize{10}{10}\selectfont

\setstretch{1.15}\fontsize{10}{10}\selectfont

\phantomsection\label{secA}
\subsection*{\small{\sf{\textbf{Appendix A}}}}
\renewcommand{\thefigure}{A.\arabic{figure}}
\setcounter{figure}{0}

The data we use for model training and testing were collected using high-resolution, multi-element synthetic-aperture-sonar (SAS) sensors with multiple frequency bands.  These sensors have an upper-end center frequency in the hundreds of kilohertz.  The spatial resolution of this band is in the centimeter range.  A low-frequency band is\\ 

\begin{wrapfigure}{r}{0.5\textwidth}
\vspace{-0.85cm}
\hspace{0.075cm}\includegraphics[width=3.2in]{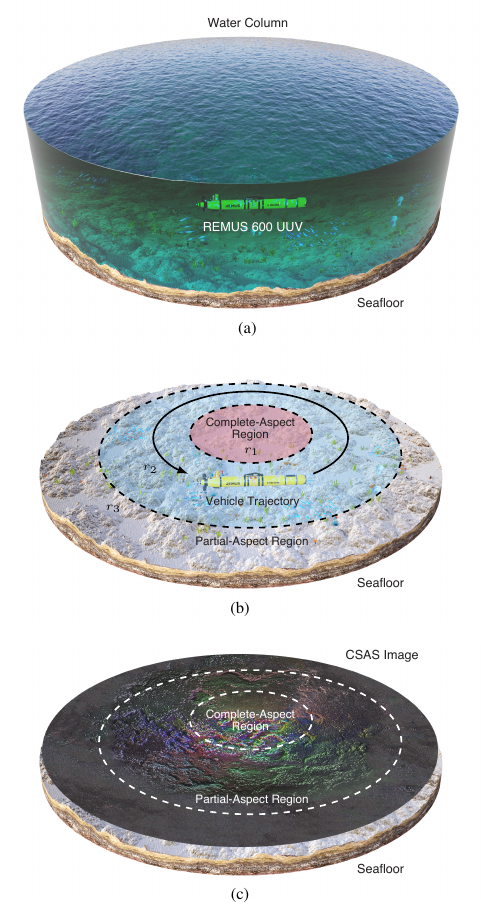}\vspace{-0.025in}\\
\caption[]{\fontdimen2\font=1.55pt\selectfont An overview of the data collection process for CSAS signatures.  (a) A diorama of a REMUS 600 UUV operating in a shallow-water environment near some hard coral.  (b) Once a pivot point has been identified, the UUV moves in an approximately circular trajectory around that point, with a typical trajectory diameter of tens of meters ($r_2$).  The UUV transmits acoustic pulses from a sonar array on the port side as it moves along this trajectory.  These pulses acoustically illuminate the seafloor and any targets in the water column, up to a certain distance.  A small region around the pivot point, referred to as the complete-aspect region (red ellipse), is ensonified from all aspect angles.  Shape resolvability is the highest in this region.  The typical diameter of this region is tens of meters in our studies ($r_1 \!\ll\! r_2$).  Outside of the complete-aspect region, the seafloor and target shape are only partly resolved.  A given spatial location may only be illuminated by a few pulses from different aspect angles.  We refer to this area as the partial-aspect region (blue ellipse).  It typically has a diameter of tens to hundreds of meters ($r_3 \!\gg\! r_2$).  The size of each region is a function of the sonar sensor characteristics, the diameter of the circular trajectory taken by the vehicle, and the vehicle glide height.  (c) An overlaid portion of a CSAS image with the complete-aspect and partial-aspect regions highlighted (white ellipses).  We use the colormap described below.  Note that the seafloor characteristics in the diorama and those in the sonar imagery are not the same.  The former were randomly generated while the latter were captured at a real-world location.\vspace{-0.6cm}}
\label{fig:csas-overview}
\end{wrapfigure}

\noindent also available, but data from it are not used in this paper.  In some cases, interferometric data can be collected and aligned with the sonar data products.  We did not use such characteristics, though, since not all of the scenes had corresponding bathymetry maps.

These SAS sensors were mounted on multiple Hydroid REMUS 600 underwater vehicles.  The vehicles operated in a variety of littoral and oceanic environments throughout the world.

In this paper, we consider several thousand unique underwater scenes.  For many of the scenes, the vehicle maneuvered in a strip-map-search mode so that either human operators or target-detection models could identify potential targets and specify a pivot location.  In others, the pivot locations were manually specified before data collection began.

After a pivot location is determined, the vehicle uses a circular search pattern to ensonify potential targets from many aspect angles.  We provide a visual overview in \cref{fig:csas-overview}.  Up to three circular passes are made for each scene, with an average diameter of 60 m.  This provides views of the underwater scene up to a circular diameter of around 150 m.  Each pass usually has slightly different center points.  Discrepancies are sometimes due to positioning-system-estimation errors.  They also stem from vehicle shift caused by strong water currents.  These center-point deviations offer differing views of the scene, which change the target aspect coverage.  The vehicle depth varies for each scene.  Typically, it is a few meters above a target.

\begin{figure*}[t!]
   \hspace{0.0cm}\includegraphics[width=6.65in]{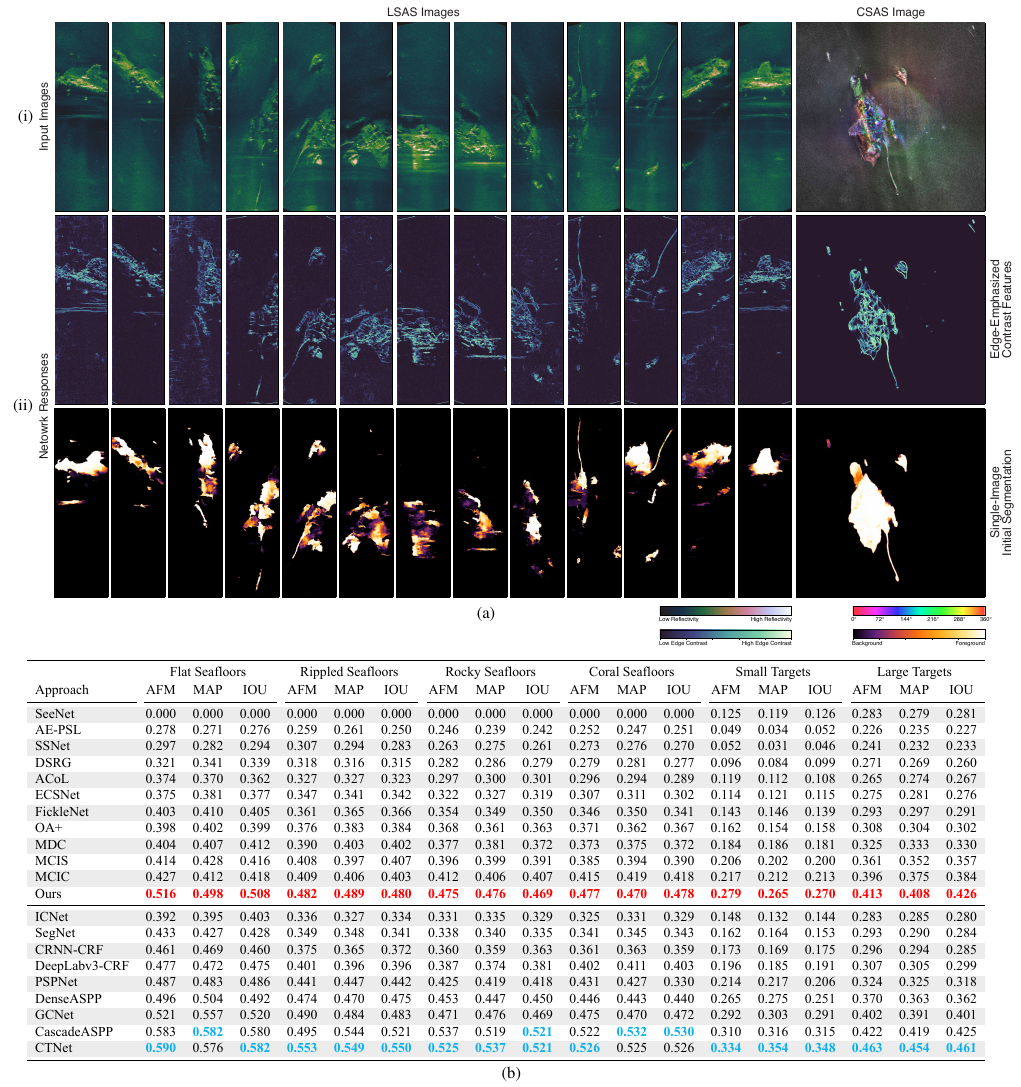}\vspace{-0.125cm}
   \caption[]{\fontdimen2\font=1.55pt\selectfont An overview of the benefits of CSAS versus LSAS for semantic segmentation.  In (a), row (i), we provide thirteen LSAS images of the same crashed fighter jet target acquired from different aspect angles.  Lighter colors indicate higher acoustic reflectivities.  The image in the last column is the CSAS image formed from a hundred sub-apertures.  A hue-based color map is used to denote ensonification direction, with brightness denoting the return strength.  In (a), row (ii), we provide plots of {\sc CREN}-inferred edge details used to form initial segmentation maps.  Lighter colors indicate more prominent edges.  As shown in these images, there is weak edge information present for many aspect angles, which is predominantly caused by acoustic shadows.  This property makes the delineation of target boundaries difficult for our multi-network approach and leads to poor masks in LSAS imagery.  The poor geometric resolvability also complicates distinguishing between target and non-target classes, which we show at the bottom of row (ii).  Here, we trained our {\sc CREN} to differentiate between foreground and background classes, a degenerate case of the semantic segmentation problem.  This was done so that we could plot a single-image class heatmap of the foreground class strength.  Darker colors denote the scene background and lighter colors correspond to the foreground elements.  For CSAS imagery, class-specific edge information is more abundant than for LSAS imagery due to the better aspect coverage.  Far better masks are typically uncovered.  While the initial segmentation mask shown has errors, it is quite close to the ground truth.  In (b), we show segmentation quality statistics for networks trained only on LSAS imagery.  When compared to the results in \cref{fig:semantic-segmentation-results}(f), the performance of all networks is much poorer than for CSAS imagery.  We recommend that readers consult the electronic version of the paper to see the full image details.\vspace{-0.5cm}}
   \label{fig:ssas-vs-csas-segmentation}
\end{figure*}

Using redundant circular trajectories leads to survey times that are often much higher than that of either LSAS or LRAS with strip-map patterns.  It is common to see a time increase of a factor of two to four.  This higher survey time is offset by the improvement in shape resolvability and class distinguisability.  Using circular trajectories hence usually yields a noticeable increase in semantic segmentation performance and a significant reduction in segmentation uncertainty.  We offer an example in \cref{fig:ssas-vs-csas-segmentation}, along with supporting statistics, to corroborate the former claim.  The multiple passes made around a center point often can be exploited to improve segmentation performance (see \hyperref[secC]{Appendix C}).

\renewcommand{\dblfloatpagefraction}{0.999}
\renewcommand{\dbltopfraction}{0.999}

As the vehicle moves in a circular trajectory, it transmits, receives, and records acoustic signatures.  We coherently sum the backscattered sound waves that are collected by the vehicle's SAS array.  Vehicle motion compensation and correction, beamforming, and image formation are conducted in manner similar to \cite{MarstonT-conf2011a}  Aperture localization is completed without the aid of either calibration scatterers or consistent global-positioning information about the platform.  Targets in the acoustic signatures are brought into focus using a correlation-based scheme \cite{MarstonT-conf2012a,MarstonTM-jour2021a}.  Multi-look processing is used to reduce sonar-image speckle \cite{ChenL-jour2020a}.

Since multi-aspect information about the underwater scene is available, a choice must be made concerning aspect-angle width and hence spatial resolution.  Narrow sub-apertures permit a high localization in aspect.  Each sub-aperture image will possess poor spatial resolution, though.  Conversely, large sub-apertures will yield imagery with high spatial resolution at the expense of poor localization in aspect.   We struck a balance between the two characteristics by considering 100 sub-apertures that are spaced uniformly by 3.6$^\circ$.  This number of sub-apertures facilitates variety of applications. 

We applied a variety of transforms to the resulting data products.  We logarithmically scaled the CSAS images to enhance scene visibility.  The images shown throughout were both contrast and brightness scaled beyond what is normally done for sonar operators.  This additional scaling was performed to help with figure legibility.  Only logarithmic scaling was employed for the imagery shown to the deep networks considered in this paper.

We also color-coded the sub-apertures to improve both scene interpretability and target distinguishability.  Sub-apertures were mapped to a continuous hue color wheel.  The colors were determined by the direction of ensonifiction, with 0$^\circ$ corresponding to red, 120$^\circ$ to blue, 240$^\circ$ to green, and so on.  The lightness of the color was specified by a reflectivity power mean.  Saturation was determined by a reflectivity power mean weighted by the sub-aperture center angle.  This representation links the physical characteristics of scattering direction, scattering intensity, and angular anisotropy with hue, variance, and saturation \cite{PlotnickDS-jour2018a}.  Example images and analyses are provided in an appendix (see \hyperref[secB]{Appendix B}).  While alternate color mappings can be employed, this one has been studied and utilized extensively for several years.  It has been repeatedly shown to be conducive for the analysis of targets and benthic habitats.  It also, as we show, yields the best segmentation performance out of the alternate colormaps that we considered.

For each CSAS image, we asked multiple experts to separately provide labeled segmentation maps.  A consensus was then reached as to the final segmentation boundaries.  Alternate data modalities, such as optical imagery, were sometimes used to correct the maps.

Given that some of our CSAS surveys relied on processed LSAS imagery, it may seem as though semantic segmentation in the former modality is redundant.  This is not true in practice, though.  Due to vehicle drift, we do not necessarily know where a potential target may lie in a CSAS image.  It can also sometimes be difficult to correctly align LSAS imagery from arbitrary passes in a region with CSAS imagery.  This is because image characteristics may be vastly different across multiple looks.  Certain target facets may additionally become better illuminated in the CSAS imagery versus the LSAS imagery.  This can lead to significantly different class segmentation boundaries compared to those found in the LSAS image.  Lastly, many of our recent surveys do not have corresponding LSAS imagery.  The vehicles simply collect CSAS data products over a given region.  In view of these facts, it is clear that there is a need for automated schemes that can analyze CSAS imagery.

\setstretch{0.95}\fontsize{9.75}{10}\selectfont
\putbib
\end{bibunit}

\clearpage\newpage
\setstretch{1.15}\fontsize{10}{10}\selectfont
\phantomsection\label{secB}

\subsection*{\small{\sf{\textbf{Appendix B}}}}
\renewcommand{\thefigure}{B.\arabic{figure}}
\setcounter{figure}{0}

In this appendix, we investigate the qualitative and quantitative effectiveness of color-by-aspect mappings for multi-aspect sonar imagery.  We illustrate that the color scheme used for the CSAS imagery is effective for human interpretation and automated analyses.

\begin{figure}[t!]
   \hspace{-0.05cm}\includegraphics[width=6.4in]{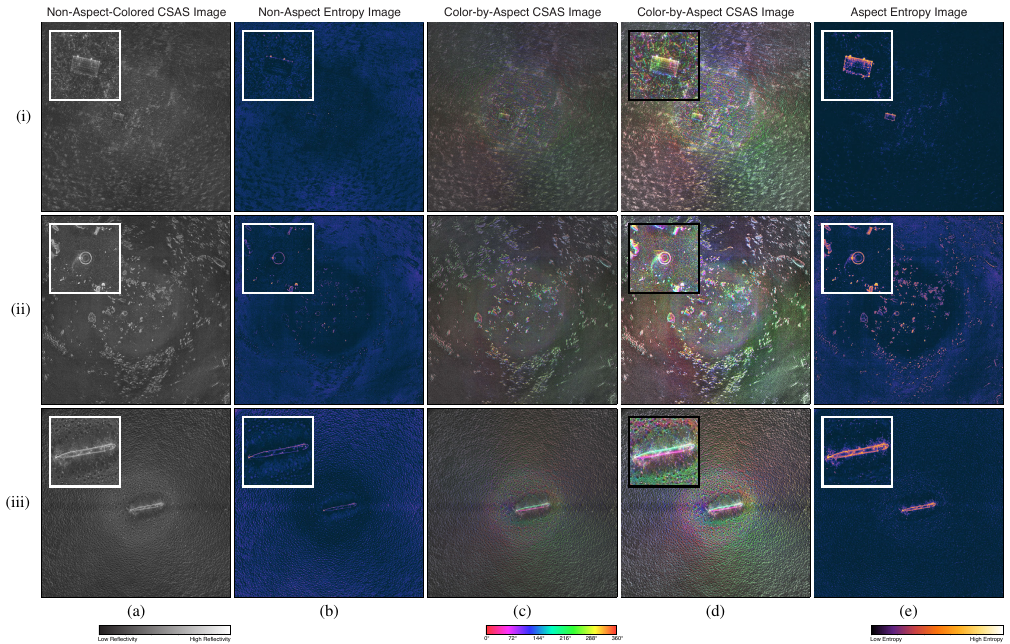}\vspace{-0.05cm}
   \caption[]{\fontdimen2\font=1.55pt\selectfont Examples of two underwater scenes and how a color-by-aspect encoding naturally enhances scene content for automated analyses.  Row (i) shows a mostly flat, sandy seafloor with a flat-screen television that is laying flat.  Row (ii) shows a flat, sandy seafloor interspersed with multiple rock formations and a rubber tire.  Row (iii) is of a pitted seafloor with a munition, likely an aircraft-launched torpedo.  (a) A non-aspect-colored CSAS image.  Here, lightness is indicative of the reflectivity strength.  The color scheme, shown in the bottom-left corner, thus is independent of ensonification angle. (b) A zoomed-in, normalized entropy image of the reflectivity map.  The call-outs in the top-left corner further zooms in on the targets.  (c) An aspect-colored CSAS image, where the angular color scheme is given in the bottom-left corner.  (d) A version of (c) with color, brightness, and contrast transformations to better highlight aspect dependence compared to (c).  The call-out in (d) focuses on the target.  (e) A normalized-entropy depiction of (d).  In both of these example scenes, relying on a color-by-contrast encoding permits better delineating class boundaries due to the encoding emphasizing anisotropy details.  This naturally permits defining more accurate segmentation maps compared to a reflectivity-only encoding.  We recommend that readers consult the electronic version of the paper to see the full image details.}\vspace{-0.4cm}
   \label{fig:csas-colormap}
\end{figure}

To motivate the use of color-by-aspect, we consider multi-aspect imagery that only contain sub-aperture reflectivity.  Examples are shown in \cref{fig:csas-colormap}(i)(a)--\labelcref{fig:csas-colormap}(iii)(a).  The color scheme is such that dark gray shades, approaching black, correspond to low acoustic returns while light gray shades, approaching white, correspond to high returns.

Reflectivity-encoded imagery contains textural content that can be used for scene analyses.  The lack of directional scattering information can significantly hinders this objective, though.  Without an aspect-dependent encoding, target facets can be challenging to distinguish from the seafloor.  For instance, the flat-screen television in \cref{fig:csas-colormap}(i)(a) has a somewhat low total reflectivity near the top of the screen and the sides.  The corresponding non-aspect-based entropy image, in figure \cref{fig:csas-colormap}(i)(b), indicates that there is little change in broadside glint from the surrounding regions.  Only the base of the television would likely be reliably delineated.  While the situation is improved somewhat for the tire in \cref{fig:csas-colormap}(ii)(a), the non-aspect-based entropy image in \cref{fig:csas-colormap}(ii)(b) suggests that the interior tire well may be ignored.  Both examples allude that automated segmentation approaches would not isolate much of the target well.  Additionally, bathymetric details cannot be reliably assessed for non-aspect-dependent color schemes, which complicates understanding scene topography.  It is difficult, for both \hyperref[fig:csas-colormap]{figures A.1}(i)(a) and \labelcref{fig:csas-colormap}(ii)(a), to discern if there are strong seafloor height changes, for example.  This behavior further impedes semantic segmentation for certain seafloor types.

The use of aperture color-coding, as illustrated in \hyperref[fig:csas-colormap]{figures B.1}(i)(c)--\labelcref{fig:csas-colormap}(iii)(c), allows for better interpretation of the scene content.  This helps with our aim of automated segmentation and labeling.  For instance, an aspect-sensitive encoding highlights bathymetry far more effectively than in \cref{fig:csas-colormap}(i)(a) and \labelcref{fig:csas-colormap}(ii)(a).  It can be discerned, for example, that both environments contain mostly flat, sandy seabeds.  Some depth change is present, though.  In \labelcref{fig:csas-colormap}(i)(c)--(d), there are mounds of sand that become conspicuous.  These small-scale seafloor features were corroborated by optical-camera footage of the area obtained by divers.  In \labelcref{fig:csas-colormap}(ii)(c)--(d), the presence of small-scale rock groups becomes far more pronounced.  For \cref{fig:csas-colormap}(iii)(c)--(d), seafloor height differences are apparent around the munition target.  It is likely that the ordnance recently impacted the seafloor with a high amount of force, which pushed the sediment and filled in some of the nearby holes.  It appears either that the debris slowly glided toward the seafloor, impacted with its broadside perpendicular to the seafloor and is partly buried, or that sediment has built up around the target over time.  Additionally, the ridges of the pits for this scene are not heavily dominated by directional scattering.  This leads us to believe that they are not rough surfaces and have a graceful gradient.  If they possessed rough edges, like the rocks in \cref{fig:csas-colormap}(ii), then we would expect to see stronger scattering at lower angles of incidence.

\begin{figure}[t!]
   $\;$\hspace{1.5cm}\includegraphics[width=5.1in]{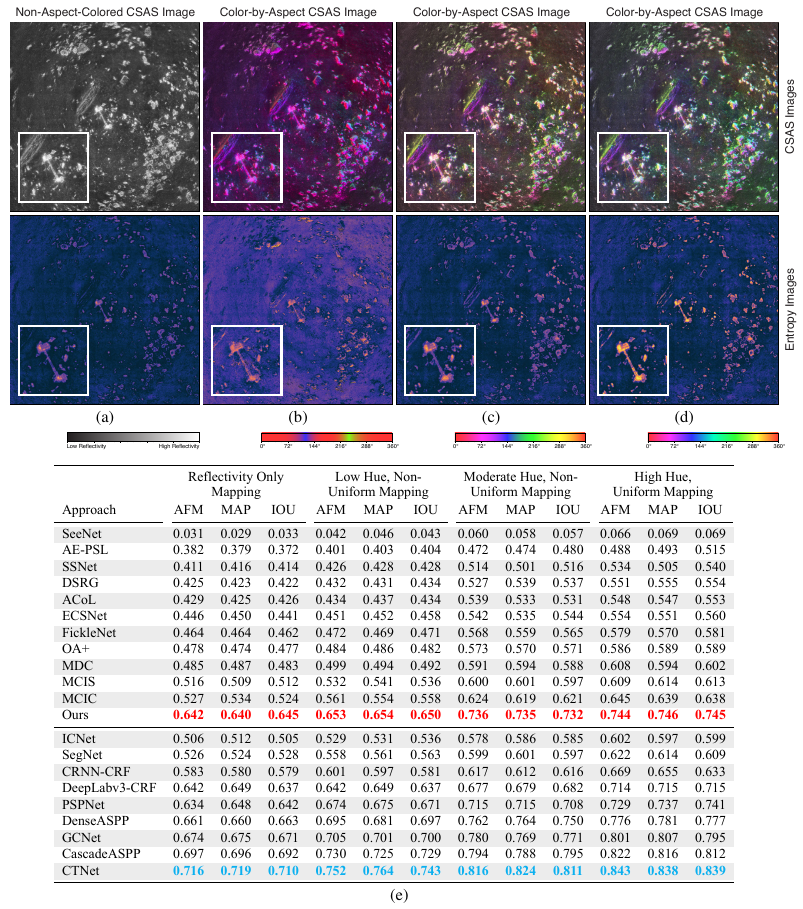}\vspace{-0.1cm}
   \caption[]{\fontdimen2\font=1.55pt\selectfont Illustration of how alternate angular color mappings can reduce segmentation performance.  In (a)--(d), we show CSAS imagery of a construction worksite lamp resting on a dredged, rocky seafloor with limestone deposits.  Non-aspect and aspect entropy maps are also provided.  For (a), only reflectivity is encoded.  For (b), many of the sub-apertures are mapped to red, with only a few to blue.  The mapping becomes more uniform in (c) but it is still compressed compared to the default in (d).  Call-out images in the bottom-left corners highlight that segmentation quality improves for our CREN as the angular color maps have a wider dynamic range and are increasingly uniform.  Target anisotropy also increases as indicated by the entropy plots.  In (e), we provide segmentation statistics for the non-angular and angular color schemes considered in (a)--(d).  We retrained the CRENs, and supporting networks, for each mapping.  These statistics are averaged across all target classes.  Higher values are better, and the best values for the weakly-supervised and fully-supervised methods are denoted using red and blue, respectively.  We recommend that readers consult the electronic version of the paper to see the full image details.\vspace{-0.4cm}}
   \label{fig:csas-altcolormap}
\end{figure}

An important insight from \hyperref[fig:csas-colormap]{figures B.1}(i)--(iii)(c)--(d) is that targets can be distinguished from seafloor via two factors.  The first is their brightness.  The second is their color contrast.  The brightness of a target region is dictated by its acoustic reflectivity.  Color contrast emerges from having facets with strong scattering directions that are different from those of surrounding regions.  Emphasizing anisotropy was a motivating factor for developing the aperture color-coding scheme that we employ.

An additional benefit of using this circular color mapping is that image rotations can be performed by simply circularly shifting the hue wheel.  We take advantage of this trait for data augmentation during network training.

In \hyperref[fig:csas-colormap]{figures B.1}(i)--(iii)(b), we plot normalized entropy images of the sub-aperture reflectivity maps from \hyperref[fig:csas-colormap]{figures B.1}(i)--(iii)(a).  Pixels which are highly anisotropic will have entropy values near one.  That is, there is complete uncertainty as to the preferred direction, since sound reflected off of that point for every ensonification angle.  High reflectivity values can also lead to high entropies.  Pixels corresponding to isotopic point scatters will have an entropy near zero.  Lower entropies are mapped to progressively cooler and darker colors in these plots, while warmer and brighter colors correspond to increasingly high entropies.

The entropy maps indicate that portions of the target boundaries can be isolated from the seafloor due to their brightness.  While considering only acoustic reflectivity can be sufficient for defining target bounding boxes, it would not be for contours of complex targets and seafloor classes that have varying surface reflectivities.  These variations could be due to material properties or simply the orientation of the targets and material on the seafloor.  

Often, it is necessary to consider directionally sensitive characteristics to produce accurate segmentations whenever variations in acoustic reflectivity are not discriminative enough.  Such characteristics can be provided by color-by-aperture encodings, and when considering the one established in an appendix (see \hyperref[secA]{Appendix A}), it becomes possible to delineate full target boundaries well.  This is corroborated when comparing the aspect entropy maps in \hyperref[fig:csas-colormap]{figures B.1}(i)--(iii)(b) to those in \hyperref[fig:csas-colormap]{figures B.1}(i)--(iii)(e).  In \hyperref[fig:csas-colormap]{figures B.1}(i)--(iii)(e), almost the full extent of the target boundaries are emphasized, even for places with low color contrast.   Some surface structure is too, like the back panels and base supports on the flat-screen television along with the propeller components and fins of the torpedo.  Even ridges of the transducer shell, along with the fractures in the torpedo casing, are highlighted well.  This is shown in \hyperref[fig:csas-colormap]{figures B.1}(i)(e) and \labelcref{fig:csas-colormap}(iii)(e).  Many of the rough-surfaced rocks in \cref{fig:csas-colormap}(ii)(d) are accented well when using a color-by-aspect encoding, which is indicated by the entropy plot in \cref{fig:csas-colormap}(ii)(e).  These components have a high entropy because the corresponding facets are either highly reflective, extremely anisotropic, or both.

Note that the aperture color scheme utilized throughout this paper is but one of many possibilities.  As well, our decision to associate certain sub-aperture center angles with specific hues was originally arbitrary.  It has, however, emerged as the dominant color scheme after several human studies.  Investigators can, however, circularly shift the hue color wheel to their preference.  More generally, they can consider any circular color scheme.

Compressing the mapping, so that it includes few hues, can significantly hamper the understanding of directional scattering, though.  It can also impact segmentation performance, as we illustrate in \cref{fig:csas-altcolormap}.  In this figure, we consider an underwater scene of a construction lamp that has been plumb-lined into a dredged basin with limestone deposits.  We consider multiple color mappings that progressively encode directional backscattering in a more perceptually distinct manner.  We start from a purely reflective CSAS image in \cref{fig:csas-altcolormap}(a).  We incorporate a small amount of directional sensitivity for the color mapping used in \cref{fig:csas-altcolormap}(b).  In both cases, the lack of anisotropy-based color cues complicates delineating many of the limestone rock faces.  It also hampers isolating the shaft of the construction lamp.  Only the lamp's flood-light shells and its base can be reliably extracted, which is due to them being metallic and hence ringing significantly.  The shaft, in comparison, is made of a plastic composite and thus does not have a high acoustic reflectivity in many directions simultaneously.  When the aspect information is emphasized more, as in \cref{fig:csas-altcolormap}(c)--(d), target anisotropy increases.  This permits using color contrast as a discriminative feature for identifying target boundaries and seafloor phenomena.  Solution quality hence improves and is corroborated by the statistics presented in \cref{fig:csas-altcolormap}(e).  These statistics were obtained by using the same training and testing protocols outlined in the experiment section (see \hyperref[sec4]{section 4}).

\clearpage\newpage
\begin{bibunit}
\bstctlcite{IEEEexample:BSTcontrol}
\setstretch{1.15}\fontsize{10}{10}\selectfont

\phantomsection\label{secC}
\subsection*{\small{\sf{\textbf{Appendix C}}}}
\renewcommand{\thefigure}{C.\arabic{figure}}
\setcounter{figure}{0}

In this appendix, we introduce the {\sc SFN}, a network for multi-image correspondence and warping.  We illustrate that this network permits forming multi-image mosaics that are suitable for the joint semantic segmentation of content from spatially proximate scenes.

\vspace{0.15cm}{\small{\sf{\textbf{Network Architecture.}}}} Our {\sc SFN} extracts features from the CSAS imagery that are used to iteratively inform a regularized, dense correspondence map between image pairs.  This is done in two separate stages.  In the first stage, a convolutional, memory-based encoder converts an image pair into pyramids of multi-scale features.  In the second stage, groups of matching, refinement, and regularization layers are employed to conduct cascaded flow inference and construct coarse-to-fine flow fields.

\begin{figure*}[b!]
   \vspace{-0.4cm}\hspace{-0.15cm}\includegraphics[]{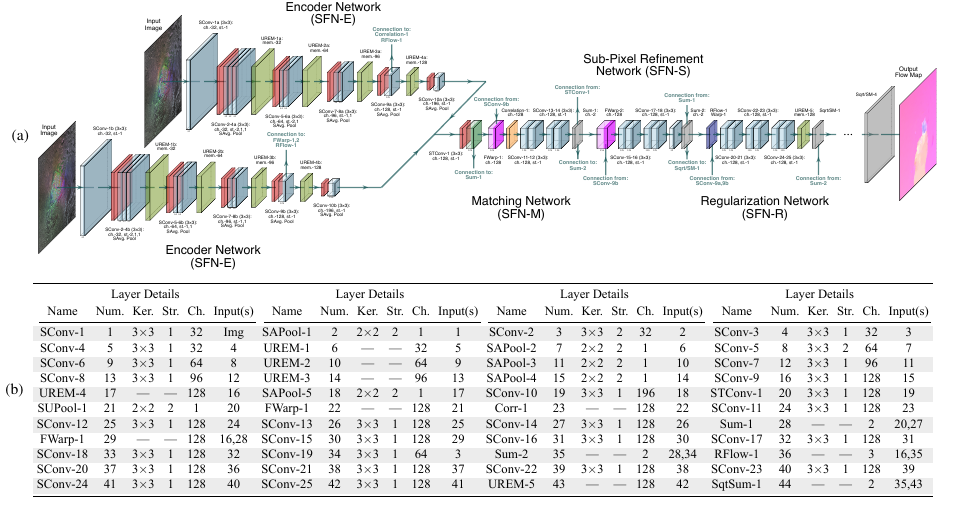}\vspace{-0.15cm}
   \caption[]{\fontdimen2\font=1.55pt\selectfont An overview of our image correspondence network.  (a) A network diagram for the small-scale flow network ({\sc SFN}) used to estimate the dense, large-displacement scene correspondence between two CSAS images with hard coral.  The {\sc SFN} relies on dual ten-layer encoders ({\sc SFN-E}s) to extract pyramidal, multi-scale features about images pairs that can be used to understand how scene content from one image changes to the next.  The filter weights for both encoders are tied together.  At each pyramidal level, a flow field is inferred from the high-level features.  This is performed by first matching descriptors according to a matching branch ({\sc SFN-M}).  The matching branch iteratively constructs a volume of region alignment costs by aggregating short-range matching costs into a three-dimensional grid.  Since the cost volume is created by measuring pixel-by-pixel correlation, the resulting optical flow estimate from the previous pyramidal level is only accurate up to that level.  Sub-pixel refinement is necessary to extend the results to a new pyramidal level, which is performed by another branch ({\sc SFN-S}).  Lastly, to remove undesired artifacts and enhance flow vectors near target boundaries, a regularization branch is used ({\sc SFN-R}).  This network relies on feature-driven local convolution to smooth the flow field in the interior of the target while preserving sharp discontinuities at the target edges.  Multiple {\sc SFN-M}, {\sc SFN-S}, and {\sc SFN-R} networks are cascaded to upsample and process the flow-field estimate; these additional networks are not shown in the above diagram.  The block coloring scheme used in \cref{fig:can-network} is reused in this diagram.  We additionally use pink boxes to denote feature warping, yellow-orange boxes to denote correlations, and purple-blue boxes to denote regularized flow warping.  (b) A tabular summary of the major network layers of the {\sc CREN}.  Since the encoder branches share the same weights, we do not distinguish between them in this table.  For conciseness, we have excluded the three additional sets of {\sc SFN-M}, {\sc SFN-S}, and {\sc SFN-R} branches from this table.  In some instances, we shorten the names of various layers.  We recommend that readers consult the electronic version of this paper to see the full image details.}
   \label{fig:sfn-network}
\end{figure*}

As shown in \cref{fig:sfn-network}, the encoder possesses a dual-stream topology that transforms a pair of sonar images into pyramids of multi-scale features.  This is done through a series of multi-stride convolutional layers with memory modules (SConv-1* through SConv-10*).  At each pyramid level, a flow field can be inferred from the high-level features of both images, versus the images themselves, using a differentiable bilinear interpolation.  This makes the {\sc SFN} robust to large-scale displacements.  As the features progress through the encoder layers, their spatial resolution is reduced so as to capture increasingly prominent and sizeable spatial changes.  We share network weights and memory contents across both streams.  This reduces the number of tunable parameters without noticeably impacting performance.

The remainder of the {\sc SFN} progressively transforms the extracted features so as to infer and correct a sub-pixel-accurate correspondence map.  This is done according to three processing blocks, which carry out descriptor matching, refinement, and regularization.  For the descriptor matching block, a spectral, transposed-convolution layer (STConv-1) is used to spatially upsample the previous flow-field estimate by a factor of two.  This is followed by feature-warping (FWarp-1) and correlation (Correlation-1) layers, which provide a point-point correspondence cost between images.  Four successive convolutional layers (SConv-11 through SConv-14) are used to construct a residual flow from the cost volume.  The upsampled flow-field estimate and residual flow are summed (Sum-1) to account for any changes at the particular current scale that could not be predicted solely through the deconvolution.  This typically yields accurate flow maps up to that scale.  The accuracy is improved further by the sub-pixel refinement network block.  In particular, a secondary residual flow field is computed via minimizing the feature-space distance between one image and an interpolated version of the second image.  Erroneous artifacts are hence prevented from being amplified when being passed to the next pyramid level.  The refinement block is composed of a feature-warping layer (FWarp-2) and four spectral convolutional layers (SConv-15 through SConv-18).  Each convolutional layer has a single stride.  Leaky rectified-linear units are inserted after every convolutional layer.  An element-wise sum layer (Sum-2) combines the residual flow field with the result from the matching block.  

Even with sub-pixel refinement, distortions and vague flow boundaries may still be present in the fields, which can disrupt the resulting warped segmentation maps.  We remove such details using feature-driven, local convolution to adaptively smooth the flow field, which is implemented by the regularization bloc.  This regularization block acts as an averaging filter if the flow variation over a given patch has few discontinuities.  It also does not over-smooth the flow field across boundaries.  To realize this behavior, we define a feature-driven distance metric that estimates local flow variation using the pyramidal features, the inferred flow field from the previous block, and an occlusion probability map.  First, we remove the mean from the inferred flow field and warp it with respect to a downsized version of the second input image (RFlow-1).  We then apply a Frobenius norm to the difference of the color intensity values of this result and the color intensity values of a downsized version of the first input image.  This result is subsequently transformed by a bank of spectral convolutional filters (SConv-20 through SConv-25).  We insert a content-addressable memory (UREM-5) to store and recall pertinent features that may improve flow-field accuracy for new images.  A normalized Boltzmann function is then applied.  Such an operation is defined using a convolutional-distance, a negative square-root, a soft-max function, and a locally-connected convolutional layer (Sqrt/SM-1).  The convolutional part of this layer adaptively constructs potentially unique soft-max-based filters for individual flow patches and convolves them with the inferred flow field to remove the undesired artifacts.  This yields a flow-field estimate at a given image resolution that is then progressively upsampled and processed by additional sub-networks.

In the latter part of the {\sc SFN}, we rely on a cost-volume-based processes \cite{XuJ-conf2017a} with a coarse-to-fine refinement strategy for establishing correspondences between multi-scale features.  That is, the {\sc SFN} quantifies and stores the cost of matching one pixel in a sonar image with a pixel in another sonar image.  Relying on such a representation simplifies the search for a minimal-disparity warping that usually enforces coherence \cite{SunD-conf2018a}.  It also facilitates the application of heuristic refinement strategies to further constrain the search process in a way that yields quantitative improvements to the warping.

Unfortunately, cost volumes can become corrupted by poorly matched features.  Such poor matches can occur when pixels from one image cannot be matched to another due to, say, occlusions.  Large-scale view-point shifts can also disrupt feature matching, as the content from one image may be entirely out of view in another.  Poor matches can additionally emerge when analyzing nearly homogeneous image regions, as there is often immense correspondence ambiguity.  It is not trivial to fix the cost volume in these cases and others.  Flow-field refinement strategies assume access to an accurate flow initialization.  Leveraging either coarser or finer-level contrast features also does not typically ameliorate the situation.  Features from any given pyramidal scale are likely to be just as susceptible as others to occlusions and textural homogeneity.  They hence would not necessarily yield a better initialization.

There are regularizers, however, that can be integrated into the cost-volume formulation to mitigate poor feature matches.  We consider two of these in the {\sc SFN}.  One involves modulating the cost volume.  Another entails using local flow consistency to deform the flow field in cases where sufficient matching uncertainty is present.  Both regularizers are borrowed from {\sc LiteFlowNetv}3 \cite{HuiTW-conf2020a}.

At the end of the {\sc SFN}, the resolution of the flow field is upsampled one last time so that is equal to that of the input sonar images.  With this flow field, we can warp image content from one image and non-linearly combine them to form a mosaic.  This process is repeated for all of the available sonar images that are captured for the same spatial region.  The final mosaic can then be passed to the {\sc CAN} and {\sc CREN} for analysis.

\begin{figure}[t!]
   \hspace{-0.075cm}\includegraphics[width=6.4in]{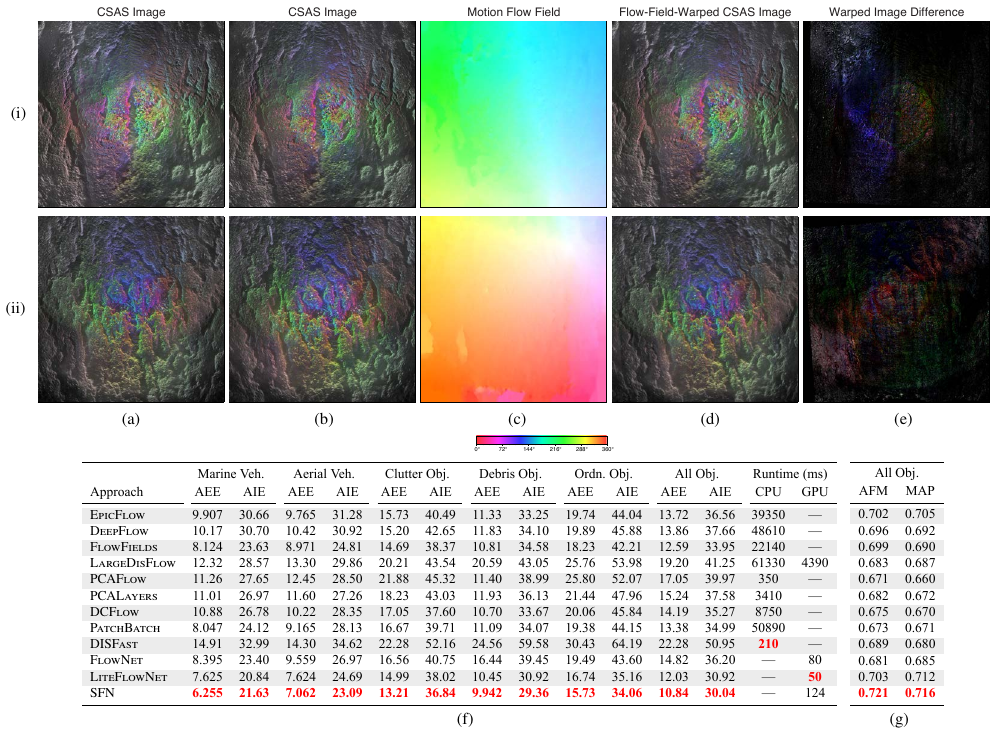}\vspace{-0.105cm}
   \caption[]{\fontdimen2\font=1.55pt\selectfont A depiction of multi-image registration.  We consider two scenes of shallow-water corals in rows (i) and (ii).  In (a) and (b), we show pairs of images with different center points.  The corresponding {\sc SFN}-inferred flow field between (a) and (b) is given in (c).  Flow vectors are color-coded according to their angle.  The flow-vector magnitude dictates both saturation and lightness.  The images in (d) show how (a) is warped according to the {\sc SFN}-derived flow field.  (d) should resemble (b) if a good flow field is uncovered.  Per-pixel image errors between (b) and (d) are provided in (e).  The errors have been enhanced for visualization purposes.  In (f), we provide a table of comparative flow-field statistics.  Here, we compare the {\sc SFN} against {\sc EpicFlow} \cite{RevaudJ-conf2015a}, {\sc DeepFlow} \cite{WeinzaepfelP-conf2013a}, {\sc FlowFields} \cite{BailerC-conf2015a}, {\sc LargeDisFlow} \cite{SundaramN-conf2010a}, {\sc PCAFlow} \cite{WulffJ-conf2015a}, {\sc PCALayers} \cite{WulffJ-conf2015a}, {\sc DCFlow} \cite{XuJ-conf2017a}, {\sc PatchBatch} \cite{GadotD-conf2016a}, {\sc DISFast} \cite{KroegerT-conf2016a}, {\sc FlowNet} \cite{DosovitskiyA-conf2015a}, {\sc LiteFlowNet} \cite{HuiTW-conf2018a}, {\sc LiteFlowNetv}2, and {\sc LiteFlowNetv}3 \cite{HuiTW-conf2020a}.  We use the average endpoint error (AEE) and average interpolation error (AIE) to quantify performance.  Runtime statistics are also listed.  Lower values are better.  In (g), we give a table of segmentation statistics for the full {CREN} network when using the comparative optical-flow approaches.  Higher values are better.  The best results are denoted using red.  These results indicate that {\sc SFN} provides the highest performance with the lowest computation time.  We recommend that readers consult the electronic version of the paper to see the full image details.\vspace{-0.5cm}}
   \label{fig:opticalflow}
\end{figure}

\begin{figure}[t!]
   \hspace{0.15cm}\includegraphics[width=6.275in]{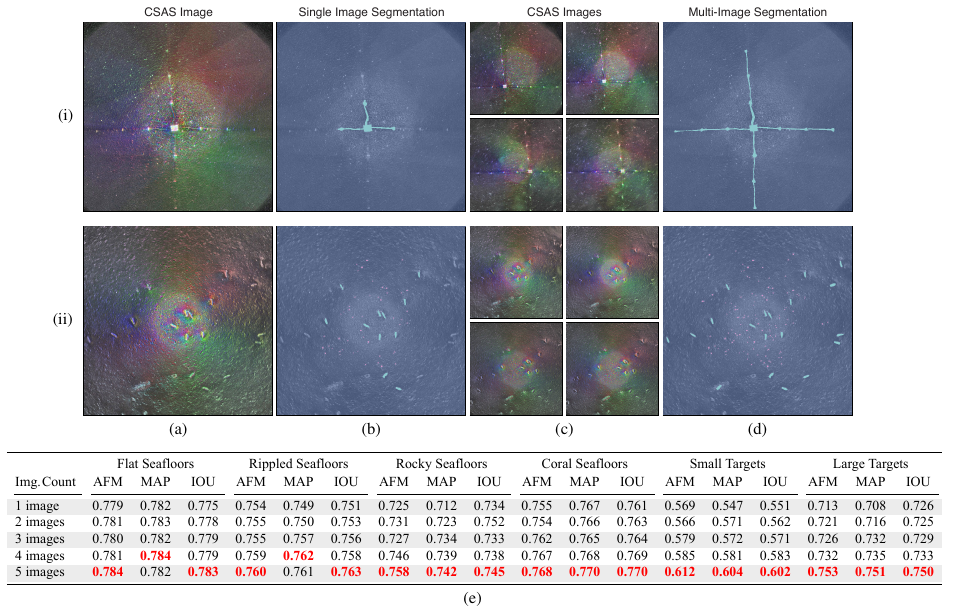}\vspace{-0.125cm}
   \caption[]{\fontdimen2\font=1.55pt\selectfont An overview of the benefits of combining segmentations from multiple images for (i) a localizing marker and (ii) a field of spent ordnance.  In (a), we show a CSAS image, along with its corresponding single-image segmentation in (b).  We utilize up to four additional surveys of the same area, shown in (c), to create an aggregate segmentation (d) that is often better than any single saliency map.  This is due to changes in aperture coverage, which increase the image contrast of certain target regions more effectively in some images.  In (e), we corroborate these claims using several segmentation statistics for our {\sc CREN}.  Higher values are better.  The best values are denoted using red.  We recommend that readers consult the electronic version of the paper to see the full image details.\vspace{-0.5cm}}
   \label{fig:multiimage}
\end{figure}

\vspace{0.15cm}{\small{\sf{\textbf{Results and Analysis.}}}} In \cref{fig:opticalflow}, we illustrate the {\sc SFN}s outperform several non-deep approaches and deep approaches.  {\sc SFN}s do better in terms of the average endpoint and interpolation error, even against {\sc LiteFlowNet}3, as indicated in \cref{fig:opticalflow}(f).  This, in turn, yields more accurate flow fields that facilitate effective multi-image segmentation, as highlighted in \cref{fig:opticalflow}(g) and is further explored in \cref{fig:multiimage}.  {\sc SFN}s have a higher throughput too.

We obtained the results in \cref{fig:opticalflow} by training, where appropriate, on the MPI Sintel \cite{ButlerDJ-conf2012a} dataset.  We then fit to the KITTI 2012 \cite{GeigerA-conf2012a} and KITTI 2015 datasets \cite{MenzeM-conf2015a}.  All three datasets are widely used for optical flow estimation.  We used the same training and convergence protocols as in the experiment section (see \hyperref[sec4]{section 4}).  We relied on author-supplied parameter values.  The results were averaged across twenty Monte Carlo trials.  The deep networks were initialized using random weights for each trial.

The improvements we observe stem from multiple mechanisms within the {\sc SFN} not found in the alternatives.  Foremost, the {\sc SFN}s extract motion-based features.  They employ pyramidal, convolutional encoders which define progressively more coarse directional features for deeper layers.  Features from these deeper layers permit handling large-scale displacements effectively, even in the presence of drastic illumination changes.  Features from earlier layers are integrated, toward the end of the flow inference pipeline, to address small-scale spatial transformations.  The interplay of both feature types is needed for multi-aspect sonar imagery, since the vehicle center position, along with its roll and pitch, can vary dramatically across circular survey trajectories.  In comparison, approaches like {\sc PCAFlow} \cite{WulffJ-conf2015a} and {\sc PCALayers} \cite{WulffJ-conf2015a}, construct flow fields directly from the images at a single spatial scale.  They can hence be ineffective whenever global, large-scale and local, small-scale transformations are simultaneously encountered and keypoint-matching errors accumulate.  These approaches also cannot always handle global acoustic illumination changes well, since the non-feature-based keypoints are sensitive to visual appearance.

The way that the flow fields are inferred from the features also aids in performance.  For each image pair, the encoder-derived features are used in a series of pixel-by-pixel matching processes at progressively larger scales.  This occurs across three processing blocks within the {\sc SFN}, which match feature descriptors, refine the flow, and regularize it.  For the descriptor-matching block, feature-warping and correlation operations are performed, which provide a point-to-point correspondence cost between images.  A residual flow is then constructed from this cost volume.  The upsampled flow-field estimate and residual flow are summed to account for any changes at the particular current scale that could not be predicted solely through convolution-based upsampling.  This yields relatively accurate flow maps up to that spatial scale.  

The flow accuracy is further improved by a sub-pixel-refinement network block.  In particular, a secondary residual flow field is computed via minimizing the feature-space distance between one image and an interpolated version of the second image.  Erroneous artifacts are hence de-emphasized when being passed to the next pyramid level.  However, even with sub-pixel refinement, distortions and vague flow boundaries may still be present in the fields.  These issues can disrupt the resulting warped saliency segmentation maps.  {\sc SFN}s remove such details using feature-driven, local convolution within the regularization block.  Such processes adaptively smooth the flow fields.  They act like averaging filter if the flow variation over a given patch has few discontinuities.  It, however, does not over-smooth the flow field across boundaries, thereby preserving well-defined object edges.  Most deep-networks neglect both of these steps \cite{WeinzaepfelP-conf2013a,GadotD-conf2016a}.  In doing so, early-stage errors propagate, disrupting the final flow solution.  The flow fields are also often blurred across boundaries, especially in the presence of multi-scale motions \cite{RevaudJ-conf2015a}. 

The {\sc LiteFlowNet}3-based regularizers that we included in the {\sc SFN} also contribute to performance.  By identifying unreliable features and modulating the cost volume, our network reduces the effects of outliers and hence constructs better initial flows.  This helps when encountering occluded seafloor components, particularly hard corals and rocky outcroppings that can obscure targets at one vehicle survey height and not another.  Additionally, the {\sc SFN}s automatically enhance the flow consistency throughout the estimation process.  They identify inaccurate flow regions using intensity edges and local flow co-occurrences near object boundaries.  Flows for these regions are replaced with more accurate ones from nearby areas using inferred displacement fields.  By systematically repairing these defects, the {\sc SFN}s handle local image homogeneity well.  The networks can therefore accurately align flat sandy seabeds with little to no texture.  Likewise, they do well large-scale targets with smooth facets, such as ship hulls and bridge pilings.  This occurs despite the {\sc SFN}s relying on predominantly directional features at each pyramid level (see \hyperref[secD]{Appendix D}).

Due to these beneficial properties, the {\sc SFN}s can uncover robust flow fields that help propagate segmentation masks from multiple images.  Performance improvements are thus observed for increasing image counts, especially when target-aspect coverage changes greatly.  This is demonstrated in \cref{fig:multiimage}.  We obtained the statistics in this figure by training the {\sc CREN} on the dataset used in the body of the paper (see section 3) and evaluating on a significantly larger CSAS dataset.  The latter dataset is composed of images where up to five circular passes were made per region.  We relied on the same training protocols as the remaining experiments (see \hyperref[sec4]{section 4}).

When using non-{\sc FlowNet} approaches, segmentation performance stagnates when considering more than two images.  The flow fields they return typically have too many issues to effectively transfer segmentation labels, except for CSAS images with slight displacements.  Most of the imagery exhibit moderate to large displacements, though.  Strong ocean currents can shift underwater sensing platforms, causing the center-position to change by several meters.  Likewise, inertial-measurement errors can disrupt vehicle state estimates and lead to the same outcome.  The utility of the non-{\sc FlowNet} approaches that we considered is thus extremely limited for our segmentation application.

If severe alignment issues are encountered, then there will be little benefit for the {\sc CREN}s to process multiple images during segmentation.  Fortunately, this does not occur.  As depicted in \hyperref[fig:opticalflow]{figures C.1}(i) and \hyperref[fig:opticalflow]{C.1}(ii), {\sc SFN}s do well for scenes with complex seafloor geometry.  They also do well for scenes with simple seabed textures, like \cref{fig:multiimage}(i) and \hyperref[fig:multiimage]{C.2}(ii).  {\sc SFN}s find suitable, non-linear interpolations that minimize the endpoint error and hence yield low per-pixel differences.  Errors that are observed are often due to changes in either reflectivity magnitudes or the aspect hue, as shown in \hyperref[fig:opticalflow]{figures C.1}(i)(e) and \hyperref[fig:opticalflow]{C.1}(ii)(e).  Neither of these discrepancies can always be mitigated well by image-interpolation methods, since the flow fields only characterize motion, not changes in visual appearance.  However, such errors do not typically influence segmentation mask aggregation within the {\sc CREN}s, so attempting to reduce them is mostly irrelevant for our application.

\setstretch{0.95}\fontsize{9.75}{10}\selectfont
\putbib
\end{bibunit}

\clearpage\newpage
\begin{bibunit}
\bstctlcite{IEEEexample:BSTcontrol}
\setstretch{1.15}\fontsize{10}{10}\selectfont

\phantomsection\label{secD}
\subsection*{\small{\sf{\textbf{Appendix D}}}}
\renewcommand{\thefigure}{D.\arabic{figure}}
\setcounter{figure}{0}

In this appendix, we augment our experimental results by describing the contrast features that emerge after pre-training and then fine-tuning the {\sc CAN}, {\sc CREN}, and {\sc SFN}.  We then discuss how the features aid in segmenting sonar imagery.

To visualize the filter receptive fields, we use an approach similar to \cite{SimonyanK-conf2014a} but in a color-decorrelated, Fourier-transformed space.  Working in such a space preconditions the receptive-field optimization procedure and aids in quick convergence.  We also use a variety of transformation robustness processes to stabilize the optimization.

\begin{figure*}
   $\;$\vspace{-0.4cm}\\
   $\;$\hspace{1.0cm}\includegraphics[width=5.5in]{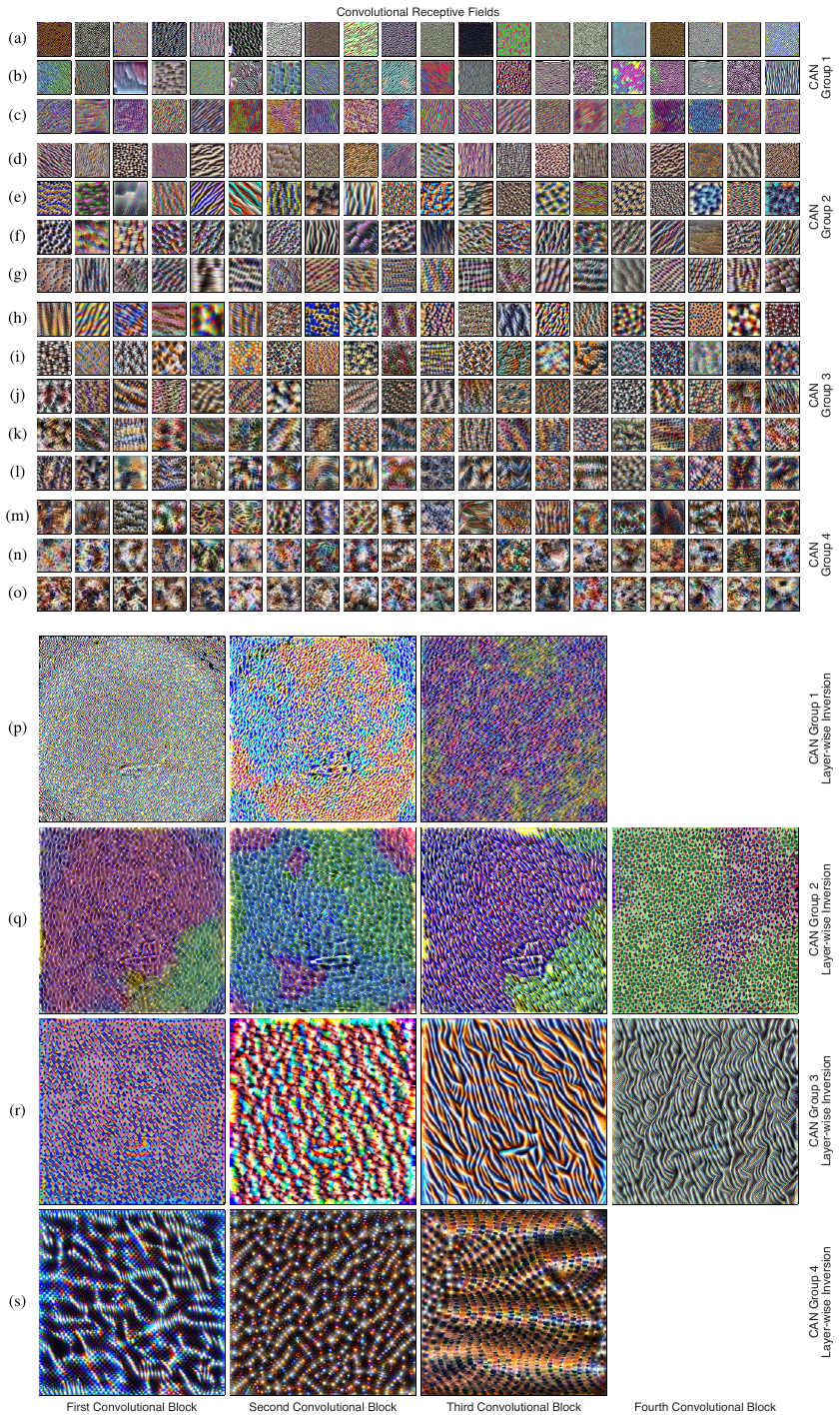}\vspace{-0.1cm}
   \caption[]{\fontdimen2\font=1.55pt\selectfont {\sc CAN} feature visualizations after training.  In rows (a)-(o), we provide group-level activation mappings.  Initial groups, like (a)--(g) focus on directional features, such as lines and simple semi-periodic textures.  Later groups, like (h)--(o), are attuned to patterns and proto-objects.  Such components emphasize relevant contrast features for classification.  In rows (p)-(s), we show the aggregate feature inversion of the first four blocks in a group.  Here, we consider a CSAS image of a crashed fighter plane with sheered wings.  The feature-inverted representations indicate that deeper groups of the {\sc CAN} retain progressively more abstract contrast features that are used to identify dominant classes.\vspace{-0.4cm}}
   \label{fig:can50-features}
\end{figure*}

\begin{figure*}
   $\;$\vspace{-0.4cm}\\
   $\;$\hspace{1.0cm}\includegraphics[width=5.5in]{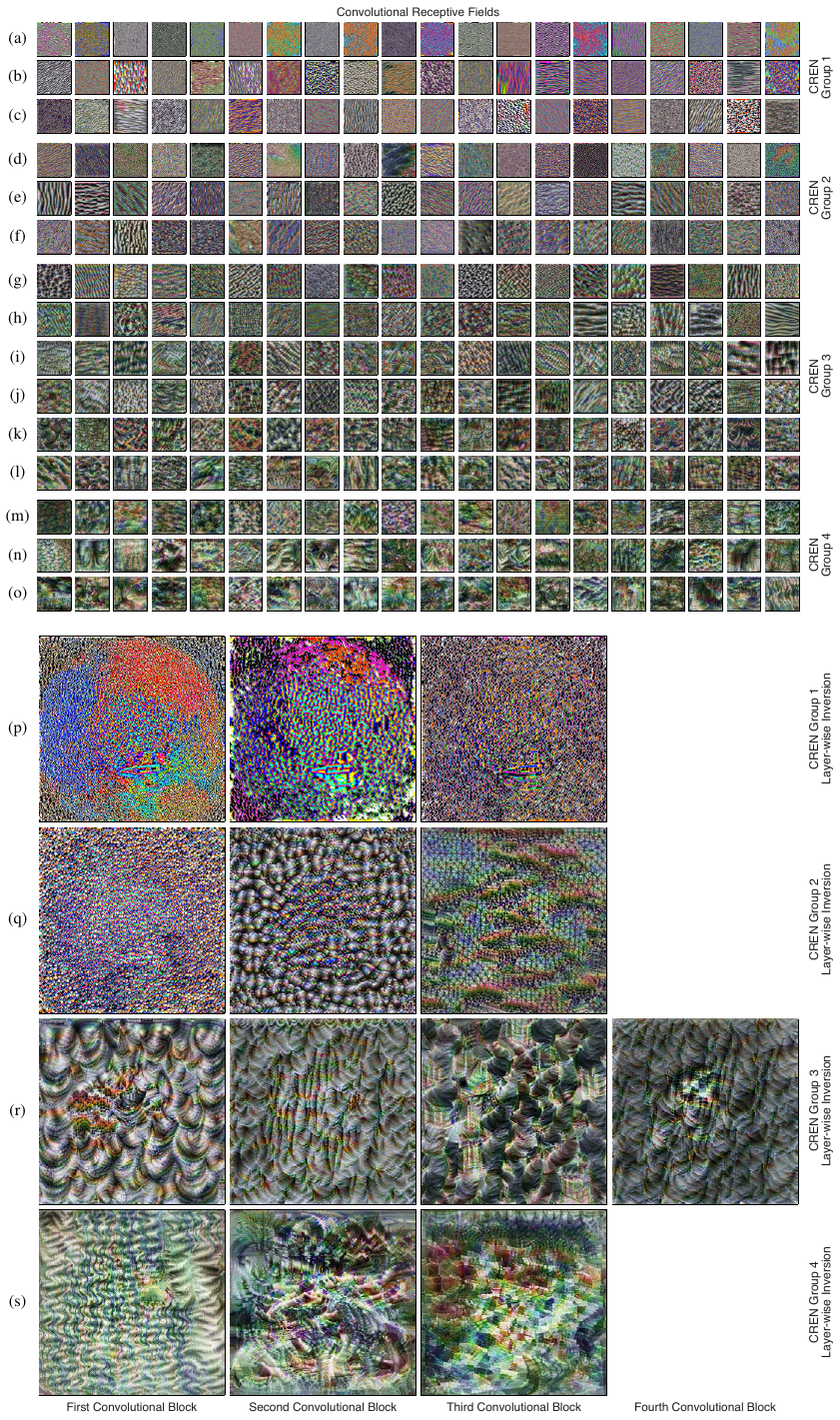}\vspace{-0.1cm}
   \caption[]{\fontdimen2\font=1.55pt\selectfont {\sc CREN} feature visualizations after training.  In rows (a)-(o), we provide group-level activation mappings.  Initial groups, like (a)--(g) focus on directional features, such as lines, edges, and simple semi-periodic textures.  Later groups, like (h)--(o), are attuned to complex patterns.  Such components emphasize relevant contrast features for segmentation.  In rows (p)-(s), we show the aggregate feature inversion of the first four blocks in a group.  Here, we consider a CSAS image of a crashed fighter plane with sheered wings.  The feature-inverted representations indicate that deeper groups of the {\sc CREN} retain progressively more pronounced directional contrast features that are used to identify class boundaries.\vspace{-0.4cm}}
   \label{fig:cren38-features}
\end{figure*}

As shown in \cref{fig:can50-features}(a) and \cref{fig:cren38-features}(a), the first two groups of both networks mainly construct low-level features.  These are color-contrast detectors and Gabor-like edge detectors, which specify a visual reconstruction basis.  Non-oriented color-contrast respond to a, mostly, uniform hue.  The oriented versions react to a particular color on one side of the receptive field and a disparate, usually complementary, color on the other side.  They emphasize localized color transitions for various orientation and shift configurations, which form the foundation of texture in sonar imagery.  Color-contrast detectors also highlight target anisotropy in multi-aspect sonar imagery.  Gabor filters mainly respond to edges at different angles.  Other instances of these filters are negative-reciprocals, which fill holes.  These two classes of Gabor filters identify possible seabed boundaries, based on the acoustic reflectances, in a coarse way.  Both the color- and edge-based features also facilitate the formation of complex filters in later layers.

The second through sixth convolutional groups extract a more diverse set of features, which can be seen in\\ \noindent \cref{fig:can50-features}(b)--(c) and \cref{fig:cren38-features}(b)--(c).  There are invariant color-contrast filters.  There are low-frequency edge detectors, which accent hazy, poorly defined edges of targets and transitions between seabed types.  Such filters are immensely useful for sonar imagery, as errors during auto-focusing can blur boundaries.  Facets of environmental features that are not acoustically illuminated from multiple directions can also appear blurred, especially if they are near the image fringes.  Color and multi-color filters are also present in these layers.  These track local brightness and hue.  Others respond to mixtures of colors, albeit without any obvious spatial preferences.  The remaining features tend to be derived from combinations of first-stage Gabor filters.  The combined-Gabors filters are fairly invariant to exact position and respond to color contrasts that align with the edges.  They appear to be tuned to reveal seabed types that are composed of local, semi-repeating patterns.  Examples include hard corals, rocky outcroppings, and rippled sediment on the seafloor.  The combined-Gabor filters additionally isolate complicated, non-locally-linear target boundaries in a lightness-invariant manner, which helps segment targets in different configurations with variable scattering intensities.

Simple shape predecessors emerge in the third through fifth convolutional groups of the {\sc CAN} and the fourth through sixth groups of {\sc CREN}.  Line, combed-line, shifted-line, curve, shifted-curve, circle, angle, corner, and divergence detectors can be found.  All of these filters react to complete and incomplete versions of their corresponding shape primitives.  Their role is to specify features that reveal class boundaries more reliably than low-level Gabors.  These filters also are the foundation of non-periodic patterns found in the {\sc CAN} and {\sc CREN}.  As shown in \cref{fig:can50-features}(d)--(e) and \cref{fig:cren38-features}(d)--(f), they are combined to yield hatched, directional detectors and multi-circle detectors in the activation space.  Rhythmic, wavy detectors are also observed, as are black-white and color-shift detectors.  These types of activations, along with the other observed patterns, predominantly focus on local changes in frequency, color alterations, and actual edges.  They emphasize some of the characteristics seen in rippled sand, rock fields, and large-scale, man-made targets and hence help to identify their boundaries. 

\begin{figure*}
   $\;$\vspace{-0.4cm}\\
   $\;$\hspace{1.0cm}\includegraphics[width=5.5in]{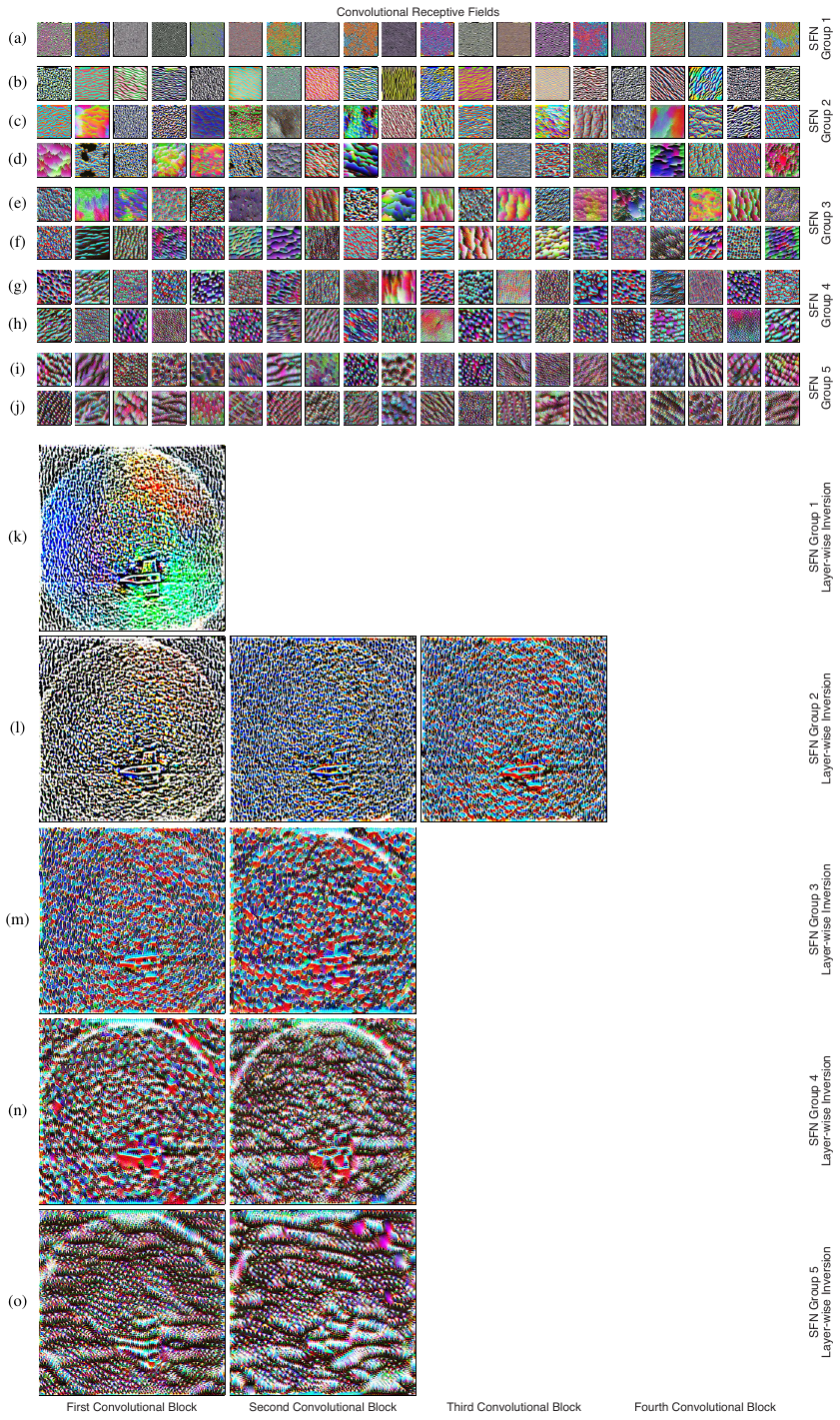}\vspace{-0.1cm}
   \caption[]{\fontdimen2\font=1.55pt\selectfont {\sc SFN} feature visualizations after training.  In rows (a)-(o), we provide group-level activation mappings.  Initial groups, like (a)--(d) focus on simple directional features.  Later groups, like (e)--(j), emphasize more complex directional features.  In rows (k)-(o), we show the aggregate feature inversion of the first four blocks in a group.  Here, we consider a CSAS image of a crashed fighter plane with sheered wings.  The feature-inverted representations indicate that deeper groups of the {\sc SFN} retain more abstract features that describe large-scale motion flow.  We recommend that readers consult the electronic version of the paper to see the full image details.\vspace{-0.4cm}}
   \label{fig:lfn90-features}
\end{figure*}

\Cref{fig:can50-features}(p)--(q) and \cref{fig:cren38-features}(p)--(q) illustrate the cumulative effect of these filters for the sonar image presented in \cref{fig:pretraining}(a)(i).  Here, we invert the features formed by all of the filters in a block of layers \cite{MahendranA-conf2015a}.  

In the second block of the first group, given in \cref{fig:can50-features}(p), the {\sc CAN} accentuates the sand-ripple crests, which is a byproduct of the curve, line, contrast, and pattern filters.  The valleys appear to be excluded entirely due to the effects of the black-white filters.  Some of the crashed fighter plane fuselage is also accentuated.  In the third block of the first group, the {\sc CAN} has begun to deduce one of the major classes in the sonar image, rippled sand.  The network further refines this reduction in the second through third blocks of the second group, which is shown in \cref{fig:can50-features}(q).  It eventually describes well the individual ripples and their orientation.  Additionally, the network relies on skip connections from preceding blocks to bring forward features that enable it to isolate the plane cockpit, fuselage, and parts of the wings that remain.  Much of the airplane is distinguished better than in earlier blocks as a consequence of brightness-gradient filters that fire due to specularities.  The improvement also stems from angle, corner, and divergence detectors.  Low-frequency edge detectors permit delineating the starboard wing well despite the acoustic pings being highly scattered for the corresponding target facets.  All of these responses enable the {\sc CAN} to reliably infer that a man-made target is present and ascertain its location.

As with the {\sc CAN}, the group in the {\sc CREN} initially highlights low-level edge details due to the oriented and complex Gabor filters.  As shown in \cref{fig:cren38-features}(p), the {\sc CREN} quickly establishes a highly non-linear representation of the sonar image content.  It traces the outlines of the crashed plane and identifies those facets that are strongly anisotropic.  As well, it leverages the class-activation maps from the {\sc CAN} to quickly identify the wavy patterns of the rippled seabed.  In the second group, the {\sc CREN} refines its understanding of the seabed texture.  The first block in \cref{fig:cren38-features}(q) indicates that the {\sc CREN} can distinguish between small-scale ripples, which are present in the complete-aspect region of the sonar image, and larger ripples, which are located near the fringes of the image.  This becomes clearer for the representation in the second block.  The larger-scale ripples are depicted by the large, repeated blob responses while the small-scale ripples are depicted using the smaller, multi-color blobs in the center of the inverted feature representation.  The {\sc CREN} uses a distinct set of filters to specify the rippled sand class near the image corners.  In these image regions, the seabed is ensonified from mainly one direction, which changes its appearance greatly compared to the remainder of the image.  Interestingly, the notion of the plane target has seemingly disappeared by the second group.  It is, however, non-linearly encoded in various layers.

Increasingly complex features are observed in the sixth through twelfth blocks.  There are center-surround detectors, which look for a particular color in the center of the receptive field and another color at the edges.  More elaborate versions, that emphasize center patterns, are also found.  The latter type of filters are sometimes combined with brightness-gradient and color-contrast ones to yield high-low frequency units that specify non-periodic textures and highlight texture transitions.  High-low-frequency changes are often an additional cue for target boundaries and help in cases where the target are juxtaposed against a high-frequency background pattern.  All of the above filters are aggregated in the sixth and seventh layers, leading to pattern and texture activations, which we present in \cref{fig:can50-features}(f)--(l) and \cref{fig:cren38-features}(f)--(l).  Based on activation-grid images, we conclude that the texture filters filters fire in the presence of rocky and hard-coral bottom types, along with loose rocks.  They also aid in the delineation of pitted sand, large sand ripples, rough troughs, and parallel grooves from multiple anchor drag marks.  Some properties of targets are additionally emphasized, but nearly not as much as the seabed characteristics. 

When pre-training the {\sc CAN} on natural imagery, the receptive fields near the ninth through twelfth blocks tend to be invariant representations of proto-objects.  Near-complete depictions of objects can also be seen.  This trait, however, is noticeably absent after fitting the {\sc CAN} to sonar imagery.  We speculate that it occurs for two reasons.  Foremost, unlike the pre-training dataset, our sonar-image dataset contains orders of magnitude fewer instances of targets.  Since the network sees them sparingly, it defaults to implementing filters that simply detect high-frequency and anisotropy transitions at local scales.  Additionally, there is a lack of supervision as to the target type in our sonar dataset, which further impedes the ability for the {\sc CAN} to encode individual classes.  This spurs the network to further implement general edge-detection processes and leverage them with local contrast features to distinguish between target and non-target regions.  We believe that the ratio of target to seafloor pixels also biases the {\sc CAN}s to implement general features that do well for seabeds.  This is not a surprising outcome, since most observed targets are relatively small and much of the sonar imagery is dominated by the seafloor content.

Filters for the remaining blocks in the {\sc CAN} and {\sc CREN} have activations corresponding to intricate, often non-repeating textures.  Most are not easy to interpret, as indicated by the examples in \cref{fig:can50-features}(m)--(o) and \cref{fig:cren38-features}(m)--(o).  They appear to be amalgams of part and texture detectors.  The filters for the final blocks of the {\sc CREN}, which are not shown in \cref{fig:cren38-features}, continue this trend.  We hypothesize that these filters serve to characterize seabed transitions for challenging scenes.  They may also be used to denote the presence of targets on certain seafloor types and represent the multitude of patterns observed within a particular seabed class.

As demonstrated in \cref{fig:can50-features}(q)--(r), filters from the later blocks of the {\sc CAN} yield increasingly abstract characterizations of the scene.  The inverted feature response for the fourth block in \cref{fig:can50-features}(q) and the first block in \cref{fig:can50-features}(r) seem to be related to combined notions of aspect-angle-dependent backscattering strength, haze caused by local speed-of-sound changes, and the presence of some semi-repeating texture.  These notions are further transformed by the second block in \cref{fig:can50-features}(r).  They eventually give way to a more easily interpretable representation in the third and fourth blocks of \cref{fig:can50-features}(r).  Here, it is obvious from the flowing, bifurcating lines that the {\sc CAN} has deduced that the rippled sand class is present.  Note that, due to the propagation of features from earlier layers, the {\sc CAN} still retains knowledge about the plane target, which is evident from the third convolutional block in \cref{fig:can50-features}(r).  For all of the blocks in \cref{fig:can50-features}(s), little discernible information about the classes is present.  Given that this is the final convolutional group in the network, it is likely that the inverted feature representation is signifying that a particular set of classes are present in the image.

Much like the {\sc CAN}, the feature inversions for the {\sc CREN} in \cref{fig:cren38-features}(r)--(s) are not readily decipherable.  Due to the presence of skip connections, multiple simple and complex notions about local image content are non-linearly mixed in the later network layers.  Nevertheless, the network leverages this information well to define initial segmentation masks that are accurate.

\setstretch{0.95}\fontsize{9.75}{10}\selectfont
\putbib
\end{bibunit}

\clearpage\newpage
\begin{bibunit}
\bstctlcite{IEEEexample:BSTcontrol}
\setstretch{1.15}\fontsize{10}{10}\selectfont

\phantomsection\label{secE}
\subsection*{\small{\sf{\textbf{Appendix E}}}}
\renewcommand{\thefigure}{E.\arabic{figure}}
\setcounter{figure}{0}

\begin{figure*}[b!]
   \vspace{-0.4cm}\hspace{-0.05in}\includegraphics[width=6.55in]{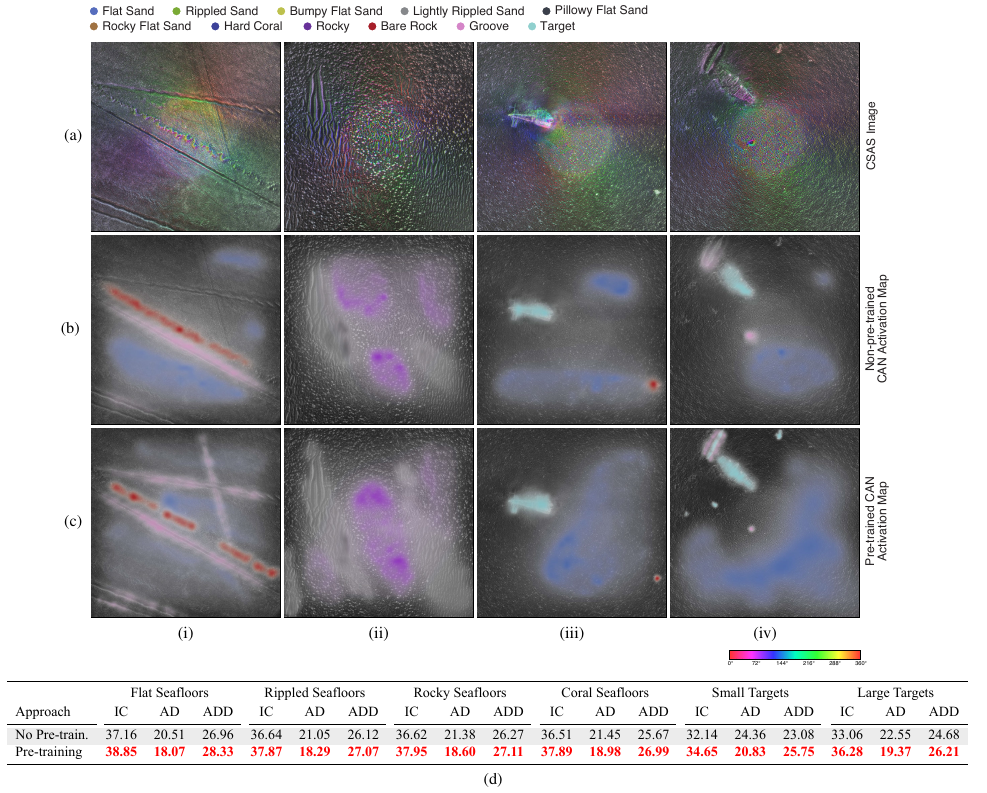}\vspace{-0.05cm}
   \caption[]{\fontdimen2\font=1.55pt\selectfont Pre-trained and fine-tuned {\sc CAN} class activation maps for four scenes, shown in columns (i)-(iv).  The scene in (i) contains flat sand, rocky sand, and anchor drag marks.  The one in (ii) is of a region with a large rock field and rippled and.  The scenes in (iii) and (iv) contain crashed aircraft on seafloor that is mostly flat sand.  In row (a), we provide CSAS imagery for each scene.  In row (b), we provide the class-activation maps for a {\sc CAN} trained solely on sonar imagery.  The maps in row (c) are from a {\sc CAN} that is pre-trained on natural imagery and subsequently fit to sonar imagery.  The results indicate that the class maps are qualitatively better in (c) than in (b) for these scenes.  The statistics presented for the table in (d) quantitatively corroborate these results across the full sonar-image dataset.  Here, we show the change in increase in confidence (IC), average drop (AD), and average drop in deletion (ADD) when pre-training versus when not pre-training.  The best values are denoted using red.  We recommend that readers consult the electronic version of the paper to see the full image details.}
   \label{fig:can-pretraining}
\end{figure*}

As we mentioned in the experimental section (see \hyperref[sec4]{section 4}), pre-training and then fine-tuning the networks tend to yields the best performance.  It is interesting to note, though, that a {\sc CREN} pre-trained only on natural imagery is often descriptive enough for segmenting sonar imagery.  In this appendix, we offer ample evidence to justify this assertion.  We also illustrate the benefits of pre-training the {\sc CAN} and the {\sc CREN} on natural imagery versus training solely on sonar imagery.

We first consider the {\sc CAN} network.  As shown in \cref{fig:can-pretraining}(b), training only on sonar imagery can yield rather poor activation maps.  In \cref{fig:can-pretraining}(b)(i), for instance, several of the anchor drag marks are missed entirely by the network.  Some of the rippled sand patches are ignored for the second scene, as indicated in \cref{fig:can-pretraining}(b)(ii).  For \cref{fig:can-pretraining}(b)(iv), the debris surrounding the crashed plane are not detected by the network.  The {\sc CREN} would be forced to compensate for these oversights, which could impede segmentation performance.

\begin{figure*}
   \hspace{-0.05in}\includegraphics[width=6.55in]{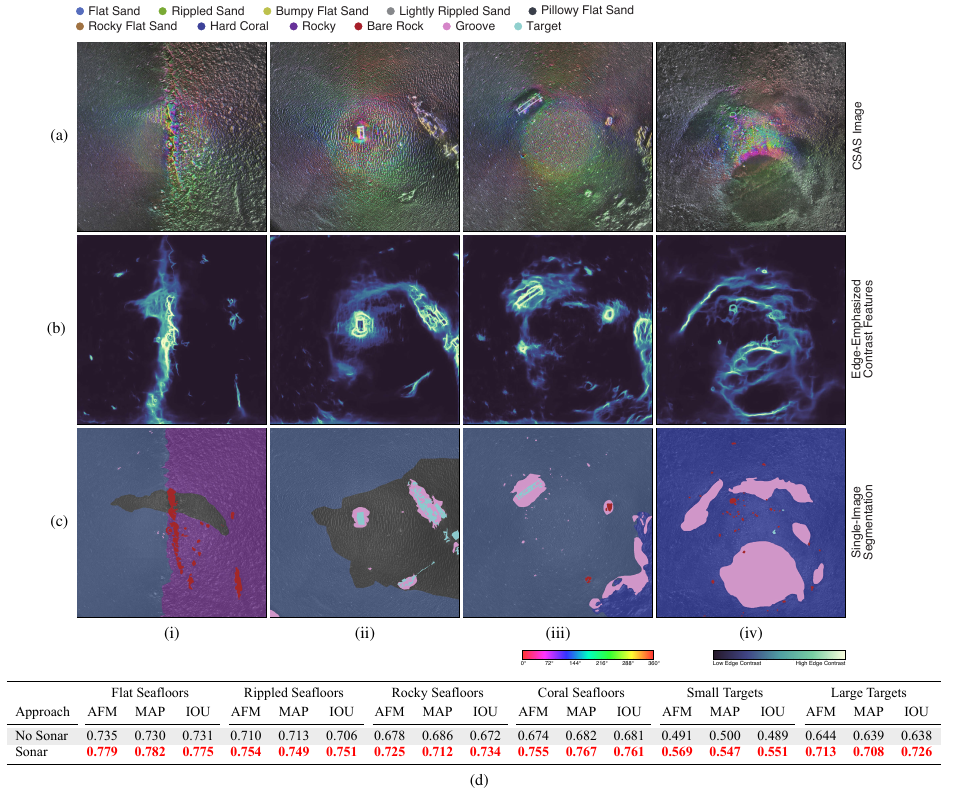}\vspace{-0.05cm}
   \caption[]{\fontdimen2\font=1.55pt\selectfont Sonar-fitted {\sc CAN} and non-sonar-fitted {\sc CREN} segmentation results for four scenes, shown in columns (i)-(iv).  Here, we trained the {\sc CREN} only on the PASCAL VOC dataset.  The scene in (i) is of a half-rocky and half-mostly-flat seafloor with some sand ripples.  The scenes in (ii) and (iii) have debris targets and large rock faces that are on either slightly-rippled or mostly-flat seabeds.  The scene in (iv) contains several flat sandy indentations that are surrounded by sparse rocky terrain.  In row (a), we provide CSAS imagery for each scene.  Aggregate edge-contrast features are provided in row (b), which highlight the edges that are emphasized by the entire backbone network.  Segmentation maps are given in row (c).  The results indicate that the PASCAL-VOC-trained {\sc CREN}s can segment CSAS imagery well despite not being fitted to them.  The statistics presented for the table in (d) quantitatively corroborate these results across the full sonar-image dataset.  Here, we show the change in average $f$-measure (AFM), mean average precision (MAP, and mean intersection-over-union (IOU) scores for segmentation when pre-training versus when not pre-training.  The best values are denoted using red.  We recommend that readers consult the electronic version of the paper to see the full image details.\vspace{-0.4cm}}
   \label{fig:cren-pretraining}
\end{figure*}

The quality of the class-activation maps is greatly improved when when pre-training the {\sc CAN} network, which can be seen in \cref{fig:can-pretraining}(c).  The grooves missed in \cref{fig:can-pretraining}(b)(i) are correctly identified in \cref{fig:can-pretraining}(c)(i), along with their near-complete extents.  The rock-laden sandy regions in \cref{fig:can-pretraining}(c)(ii) are better accented than in\\ \noindent \cref{fig:can-pretraining}(b)(ii).  The heavily rippled sand is also delineated well.  In \cref{fig:can-pretraining}(c)(iv), all of the scattered debris is highlighted.  For \cref{fig:can-pretraining}(c)(iii), the results are mostly consistent with those in  \cref{fig:can-pretraining}(b)(iii), though some of the class boundaries are better respected.

Such improvements, and those presented in \cref{fig:can-pretraining}(d), likely stem from several factors.  Foremost, pre-training the {\sc CAN} can yield good initializations for the weights in the early network layers.  Such weights often remain stable, at least from what we observed in our experiments.  This permits the gradient updates to predominantly change the class-specific filters in the later network stages during fine tuning.  That is, the gradient updates focus more on adjusting high-level concepts implemented by the {\sc CAN} versus lower-level ones.  Secondly, pre-training tends to reduce the chance of overfitting \cite{RiceL-conf2020a}, thereby allowing the networks to generalize better to unseen samples.  Pre-training also improves robustness to class imbalance \cite{HendrycksD-conf2019a}, which is crucial here, since some of the classes are significantly underrepresented.  All of these behaviors provide better seed cues.  In doing so, the {\sc CREN} can more easily adjust the segmentation boundaries derived from the class-activation maps, especially for non-highlighted class regions that are isolated from the seed cues.

For these results, we pre-trained and fine-tuned the {\sc CAN} using the same protocols outlined in the experiment section (see \hyperref[sec4]{section 4}).  This process was necessary to derive reasonable seed points from the class-activation maps.  Training on only natural imagery would yield few to no seed cues, since sonar imagery exhibits vastly different class statistics than natural imagery.

We now validate our second claim, which is that {\sc CREN}s trained on only natural imagery can semantically segment scenes well.  Here, we used the same pre-training protocols employed throughout the paper (see \hyperref[sec4]{section 4}) and then fixed the weights.  The {\sc SFN} was not used in these cases to avoid biasing the results.

Results for this experiment are shown in \cref{fig:cren-pretraining}.  For the scenes presented in \cref{fig:cren-pretraining}(i)--(iv)(a), the non-fine-tuned {\sc CREN} uncovered mostly accurate class boundaries.  It separates well the flat, sandy seafloor from the rocky terrain present in \cref{fig:cren-pretraining}(i)(a).  It also isolates many of the larger rocks.  The boundaries are not perfect, though, since the {\sc CREN} has not learned to use acoustic-shadow cues outside of the complete-aperture region to refine the predictions.  In \hyperref[fig:cren-pretraining]{figures D.4}(ii)--(iii)(c), the network correctly identifies the debris targets, the rocky outcropping, and, for \cref{fig:cren-pretraining}(ii), the slightly wavy sand.  The pitted flat sand poses some difficulties for the segmentation in \cref{fig:cren-pretraining}(iv)(c), since they lie outside of the complete-aperture region and thus exhibit low contrast.

\begin{wrapfigure}{r}{0.5\textwidth}
\vspace{-0.3cm}
\hspace{-0.175cm}
\begin{tabular}{c c}
\includegraphics[width=1.57in]{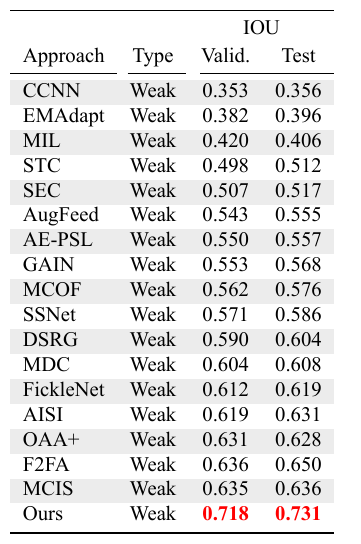} & \hspace{-0.5cm}\includegraphics[width=1.57in]{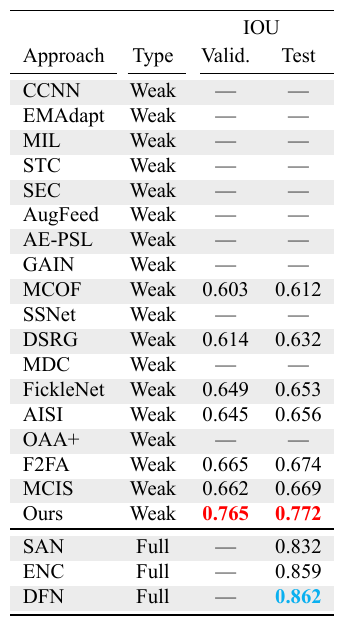}\vspace{-0.07in}\\
{\footnotesize (a)} & \hspace{-0.5cm}{\footnotesize (b)}\end{tabular}\vspace{-0.05in}\\
\caption[]{\fontdimen2\font=1.55pt\selectfont Pre-training performance of our framework on the PASCAL VOC 2012 dataset for: (a) A {\sc VGG}-16-like backbone network and (b) a {\sc ResNet}-101-like backbone network.  For both feature-extraction backbones, we include the enhancements considered in the paper.  Here, we compare against the following weakly-supervised methods: {\sc CCNN}~\cite{PathakD-conf2015a}, {\sc EMAdapt}~\cite{PanpandreouG-conf2015a}, {\sc MIL}~\cite{PinheiroPO-conf2015a}, {\sc SEC}~\cite{KolesnikovA-conf2016a}, {\sc AugFeed}~\cite{QiX-conf2016a}, {\sc STC}~\cite{WeiY-jour2017a}, {\sc AE-PSL}~\cite{WeiY-conf2017a}, {\sc GAIN}~\cite{LiK-conf2018a}, {\sc MCOF}~\cite{HuangZ-conf2018a}, {\sc DSRG}~\cite{HuangZ-conf2018a}, {\sc MDC}~\cite{WeiY-conf2018a}, {\sc AISI}~\cite{FanR-conf2018a}, {\sc FickleNet}~\cite{LeeJ-conf2019a}, {\sc SSNet}~\cite{YuZ-conf2019a}, {\sc OAA}+~\cite{JiangP-T-conf2019a}, {\sc F2FA}~\cite{LeeJ-conf2019b}, and {\sc MCIS}~\cite{SunG-conf2020a}.  We additionally compare against the following fully supervised methods: {\sc SAN} \cite{ZhongZ-conf2020a}, {\sc ECN} \cite{ZhangH-conf2018a}, and {\sc DFN} \cite{YuC-conf2018a}.  We report mean IOU scores for both the standard test and validation set splits.\vspace{-0.3cm}}
\label{fig:pascalvoc-results}
\end{wrapfigure}

In each of these cases, the pre-trained {\sc CREN} isolates different classes well, provided that the {\sc CAN}s yield high-quality seed cues.  Likewise, the network does well over the entire imaging sonar dataset.  We believe this occurred because the {\sc CREN}'s features learn to emphasize edges and respect them when growing the segmentation boundaries regardless of the imaging modality.  To show this, we plotted, in \cref{fig:cren-pretraining}(i)--(iv)(b), instances of the edge-sensitive features derived from the {\sc CREN}s.  These edge maps indicate that the {\sc CREN} can reliably detect transitions between high-contrast regions.  They also accent differently textured areas of the scenes, such as for the rocky terrain and the flat sand in \cref{fig:cren-pretraining}(i)(b) and the pitted and flat sand in \cref{fig:cren-pretraining}(iv)(b).  Some amount of anisotropic sensitivity is also present, which is obvious from \cref{fig:cren-pretraining}(ii)(b) and \cref{fig:cren-pretraining}(iii)(b).  We find that specializing the {\sc CREN} to imaging sonar data helps to emphasize anisotropy-based cues more and utilize them effectively. 

\begin{wrapfigure}{r}{0.3\textwidth}
\vspace{-0.3cm}
\hspace{0.175cm}
\begin{tabular}{c}
\includegraphics[width=1.57in]{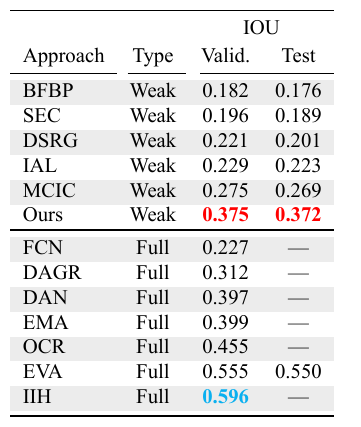}
\end{tabular}\vspace{-0.05in}\\
\caption[]{\fontdimen2\font=1.55pt\selectfont Pre-training performance of our framework on the MS COCO Stuff dataset.  Here, we compare against the following weakly-supervised methods with their default backbones: {\sc BFBP}~\cite{SalehF-conf2016a}, {\sc SEC}~\cite{KolesnikovA-conf2016a}, {\sc DSRG}~\cite{HuangZ-conf2018a}, {\sc IAL}~\cite{WangX-jour2020a}, and {\sc MCIC}~\cite{FanJ-jour2023a}.  We additionally compare against the following fully supervised methods: {\sc FCN}~\cite{LongJ-conf2015a}, {\sc DAGR}~\cite{ShuaiB-conf2016a}, {\sc DAN}~\cite{FuJ-conf2019a}, {\sc EMA}~\cite{LiX-conf2019a}, {\sc OCR}~\cite{YuhuiY-conf2020a}, {\sc EVA}~\cite{FangY-conf2023a}, and {\sc IIH}~\cite{WangW-conf2023a}.  We report mean IOU scores for both the standard test and validation set splits.\vspace{-0.2cm}}
\label{fig:mscoco-results}
\end{wrapfigure}

Despite the {\sc CREN} not being fit to sonar imagery, it is able to supply mostly correct labels for segmentation maps obtained from sonar imagery.  This occurs because the labels used for the seed cues are derived from the {\sc CAN}.  Some discrepancy in performance is observed, though, compared to the case where natural-image pre-training and sonar-image fine-tuning is used.  This is a byproduct of two factors.  Foremost, the superpixels generated by the {\sc USN} are not necessarily attuned well to the imaging-sonar characteristics.  Secondly, the {\sc CREN} acts as if there is no content-addressable memory available.  The memories are wiped after natural-image pre-training and no relevant sonar-derived features are present in the memory banks.

It may be somewhat surprising that pre-training on natural imagery confers benefits for both networks.  However, pre-training and then fine-tuning a deep network is known to significantly improve performance for the target task \cite{GirshickR-conf2014a}.  Natural-image to sonar-image transfer learning is known to be viable \cite{HuoG-jour2020a,SledgeIJ-jour2022a}.  We have also successfully employed natural-image pre-training for tasks involving other imaging-sonar modalities, including volumetric sonar \cite{MarstonTM-jour2016a}.  We hypothesize that pre-training works because the visual characteristics are similar enough between natural imagery and many imaging-sonar modalities.  Strong edges and hence changes in contrast are typically present between class boundaries, for instance, which serve as discriminative cues.

Our framework must at least achieve reasonable performance when pre-training for this transfer learning to be effective.  In \cref{fig:pascalvoc-results}, we demonstrate that this occurs for the PASCAL VOC 2012 dataset \cite{LiY-conf2014a,EveringhamM-jour2010a}.  We additionally report mean-intersection-over-union (IOU) scores that are state of the art for weakly-supervised approaches.  These scores are also rather compelling against some of the best-performing, fully-supervised approaches that are currently available.

The PASCAL VOC 2012 dataset roughly mimics traits seen in our CSAS dataset.  Both have about twenty classes.  Many classes can appear in an image.

In the future, we may collect sonar data that possess many times more target types and seafloor types.  A larger number of classes may appear in an image.  It is therefore useful to see how well our framework can scale to these cases.  We simulate this case by considering pre-training performance on the MS COCO Stuff dataset \cite{CaesarH-conf2018a}.

In \cref{fig:mscoco-results}, we report mean IOU scores for the MS COCO Stuff dataset.  We illustrate that state-of-the-art weakly-supervised segmentation performance is achieved for our framework.  Our framework is also competitive against older, fully-supervised models.  This is despite our framework having access to significantly fewer supervisory signals.  A performance gap remains, though, between our framework newer models for at least two reasons.  Foremost, it is likely that a limited number of seed cues cannot encompass all of the visual characteristics for a given class.  Additionally, our framework possesses a mere fraction of the number of parameters as present-day, fully-supervised models.  Even if reliable seed cues could be discovered, the overall modeling capacity of our framework is limited.

For the results in \cref{fig:pascalvoc-results}, we swapped the backbones of both the {\sc CAN} and {\sc CREN} encoders with either {\sc VGG}-16 or {\sc ResNet}-101 networks to align with what many authors normally consider.  We did, however, include all of the other enhancements that we used in the paper.  These are core parts of our framework and should not be excluded, as they provide noticeable benefits (see \hyperref[secF]{Appendix F}).  For the results in \cref{fig:mscoco-results}, we used the default feature backbones listed by the various authors.  We also employed the same training protocols used throughout the paper (see \hyperref[sec4]{Section 4}).  While other authors sometimes utilize different training strategies, ours is fairly standard and should also promote a fair comparison.  When relying on model compression and the backbones specified in both the {\sc CAN} and {\sc CREN} encoders, the mean IOU performance of our framework is on par with that of the {\sc ResNet}-101 version in \cref{fig:pascalvoc-results}.  We also did not utilize the {\sc SFN} network for any of these tests.  The multi-image registration and alignment capabilities offered by this network are mainly suited for our applications where significant overlap in scene content is expected.  The PASCAL VOC 2012 and MS COCO Stuff datasets do not possess this trait.

For both natural-image datasets, we improve upon the state of the art for weakly-supervised networks by about ten percent.  Our specified thresholds for the statistical tests are met for these comparisons.  Our findings are thus likely statistically significant.

\setstretch{0.95}\fontsize{9.75}{10}\selectfont
\putbib
\end{bibunit}

\clearpage\newpage
\begin{bibunit}
\bstctlcite{IEEEexample:BSTcontrol}
\setstretch{1.15}\fontsize{10}{10}\selectfont

\phantomsection\label{secF}
\subsection*{\small{\sf{\textbf{Appendix F}}}}
\renewcommand{\thefigure}{F.\arabic{figure}}
\setcounter{figure}{0}

\begin{figure}[t!]
   \hspace{-0.075cm}\includegraphics[width=6.75in]{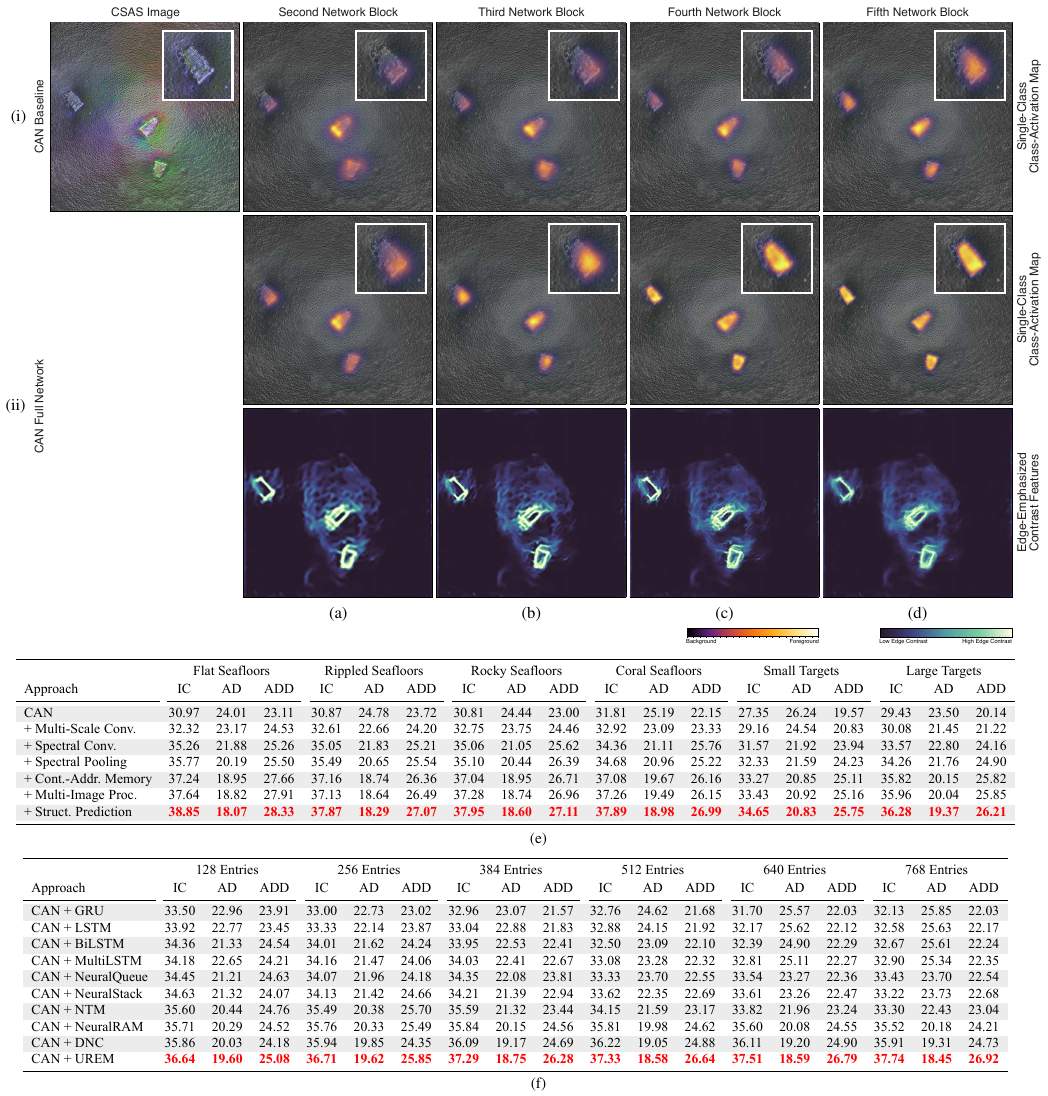}\vspace{-0.105cm}
   \caption[]{\fontdimen2\font=1.55pt\selectfont A study of a convolution-only {\sc CAN} architecture against the version considered in this paper.  Here, we consider a scene with man-made debris.  In (a)-(d), we provide class-activation maps, for a single class, produced for the final four blocks of the encoder backbone of the {\sc CAN}s.  We connected small-scale decoder networks, whose responses were not back-propagated during training, to produce classification responses and hence yield class-activation maps.  The maps for each convolutional block have been bilinearly re-sampled to be the same size as the CSAS image.  In (a)-(d), we also provide aggregate edge-contrast features.  The color scheme is such that light yellow corresponds to a dominant edge while dark blue corresponds to no edge.  The results for the full {\sc CAN}, given in row (ii), are much higher quality than the baseline, given in row (i).  The class-activation mappings for the full {\sc CAN} often obey object boundaries, even in the earliest stages of the network.  This property helps in the selection of high-quality seed cues for segmentation.  We supply a comprehensive ablation study in the table shown in (e).  From these results, we see that the performance of the baseline {\sc CAN} is significantly lower than the full {\sc CAN} considered throughout this paper.  In all cases, Lift-CAM was employed to infer the class-activation maps.  A table outlining the impact of the memory type and size is given in the table in (f).  Here, we consider gated-residual units (GRUs) \cite{ChoK-jour2014a}, long short-term memories (LSTMs) \cite{HochreiterS-coll1996a}, bi-directional LSTMs (BiLSTMs) \cite{GravesA-jour2005a}, multiplicative LSTMs (MultLSTMs) \cite{KrauseB-conf2017a}, neural queues and stacks \cite{GrefenstetteE-coll2015a}, neural Turing machines (NTMs) \cite{GravesA-jour2014a}, neural random-access memories (NeuralRAMs) \cite{WestonJ-conf2015a,GuicehreC-jour2018a}, differential neural computers (DNCs) \cite{GravesA-jour2016a}, and the default cause of our universal recurrent memories (UREMs).  We recommend that readers consult the electronic version of the paper to see the full image details.\vspace{-0.5cm}}
   \label{fig:can-ablation-study}
\end{figure}

We have introduced a number of changes to standard convolutional backbones to promote better segmentation performance.  In this appendix, we conduct an ablation study to quantify the contribution of each change.

We first consider the {\sc CAN} network.  Here, we modify the network slightly.  After each convolution-memory block, we insert a small number of additional convolution layers and append a soft-max layer.  This is done to create a series of classifiers so that we can assess what the network has learned at a given stage.  We do not back-propagate the error from each of these classifier branches onto the main pathway of the {\sc CAN} to avoid biasing the network's performance.  The network training protocols are the same as in the experiment section (see \hyperref[sec4]{section 4}).

As shown in \cref{fig:can-ablation-study}(i), the baseline network, without any of our customizations, does poorly when attempting to identify target and non-target classes.  It fails to extract target-specific features in the earliest stages of the network.  The network has substantial difficulties with targets outside the complete-aperture radius, likely due to a combination of the poor contrast outside that area and the limited expressiveness and transformation power of the early-stage filters.  To compound matters, the feature quality does not change greatly in later network stages.  The true target extents are not defined well, even in the deepest part of the network.  The class-activation confidence is also low throughout the inferred maps for many of the targets.  This occurs even within the complete-aperture region where a great amount of contrast and anisotropy cues should be available to the network.  The inability for the network to not only delineate class boundaries well, but also highlight class-relevant spatial regions can lead to the formation of poor seed cues.  Few locations are available for sampling and even fewer may be deemed to have a low enough classification uncertainty to be selected.

The full network, in comparison, fares much better.  As illustrated in \cref{fig:can-ablation-study}(ii), even in the earliest stages, the network yields class-activation maps that are qualitatively equivalent to those near the end of the baseline network considered in \cref{fig:can-ablation-study}(i).  The map quality also drastically improves with each additional stage.  The target-class maps are nearly homogeneous for each piece of metallic debris.  The extents of the class map also align well with the edge-emphasized contrast features uncovered by the network at each stage.

\begin{figure}[t!]
   \hspace{-0.075cm}\includegraphics[width=6.75in]{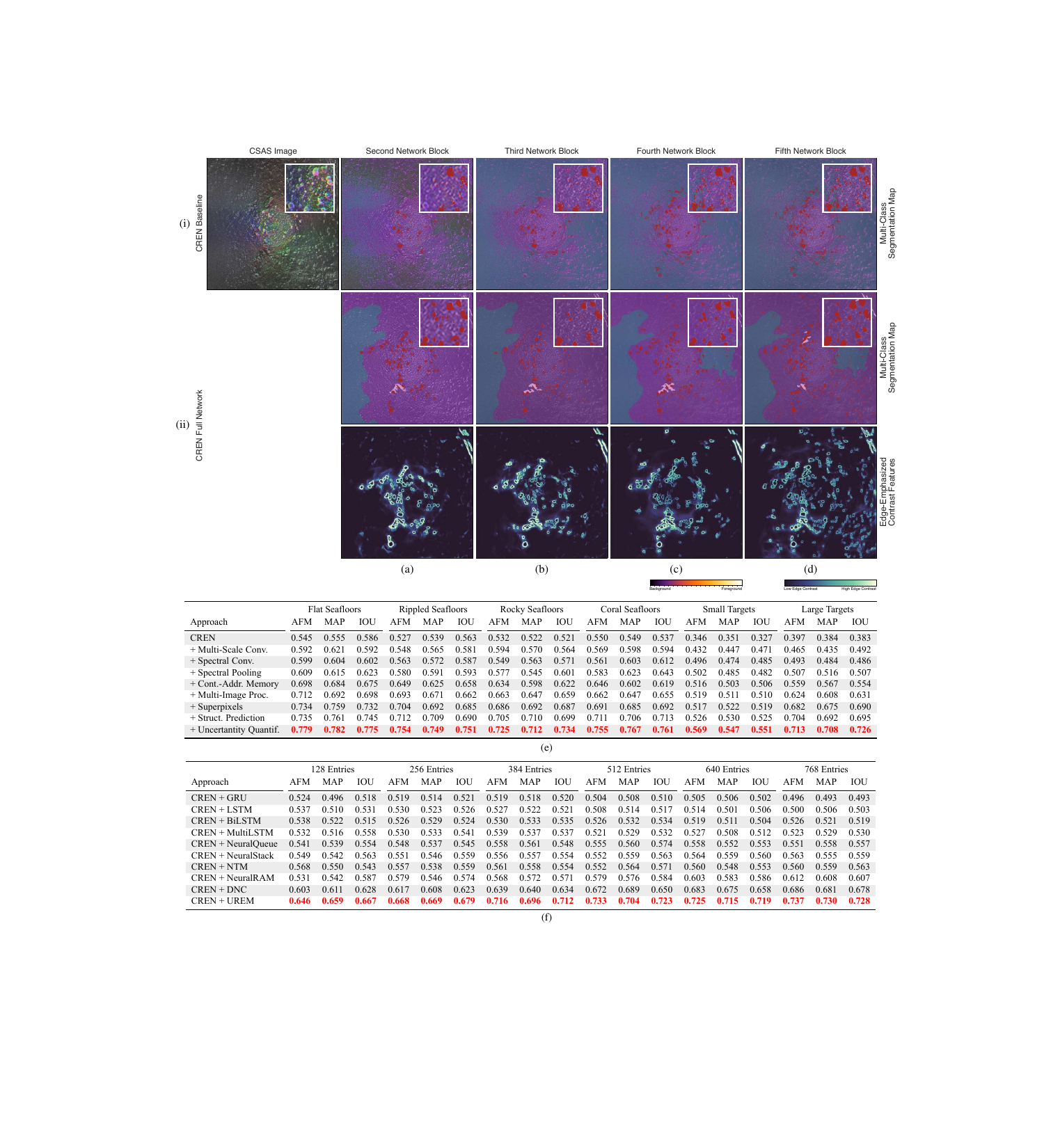}\vspace{-0.105cm}
   \caption[]{\fontdimen2\font=1.55pt\selectfont A study of a convolution-only {\sc CREN} architecture against the version considered in this paper.  Here, we consider a scene with a mixture of flat sand, rocky sand, and large rocks.  In (a)-(d), we provide segmentation maps produced for the final four blocks of the decoder backbone of the {\sc CREN}s.  We connected small-scale networks, whose responses were not back-propagated during training, to produce a segmentation response.  In (a)-(d), we also provide aggregate edge-contrast features.  The color scheme is such that light yellow corresponds to a dominant edge while dark blue corresponds to no edge.  The results for the full {\sc CREN}, given in row (ii), are much higher quality than the baseline, given in row (i).  The segmentation maps for the full {\sc CAN} often obey object boundaries.  They also delineate well the changes between seafloor types.  We supply a comprehensive ablation study in the table shown in (e).  From these results, we see that the performance of the baseline {\sc CREN} is significantly lower than the full {\sc CREN} considered throughout this paper.  A table outlining the impact of the memory type and size is given in the table in (f).  Here, we consider GRUs \cite{ChoK-jour2014a}, LSTMs \cite{HochreiterS-coll1996a}, BiLSTMs \cite{GravesA-jour2005a}, MultLSTMs \cite{KrauseB-conf2017a}, NeuralQueues and NeuralStacks \cite{GrefenstetteE-coll2015a}, NTMs \cite{GravesA-jour2014a}, NeuralRAMs \cite{WestonJ-conf2015a,GuicehreC-jour2018a}, DNCs \cite{GravesA-jour2016a}, and the default cause of our UREMs.  We recommend that readers consult the electronic version of the paper to see the full image details.\vspace{-0.5cm}}
   \label{fig:cren-ablation-study}
\end{figure}

We primarily attribute this performance gain to two factors.  The first is that our use of global spectral filters permits efficiently defining large receptive fields that can aggregate information over a wide area \cite{LuoW-coll2016a} without a reduction in resolution.  Multi-scale contexts have demonstrated promise for making robust classification decisions \cite{YuF-conf2016a} and this is likely the case here too.  Our simultaneous integration of localized receptive fields, along with regularizers like drop-out \cite{SrivastavaN-jour2014a}, preempt overfitting.  With only local, non-regularized filters, significantly more network layers than what we use would be needed to realize large receptive fields.  There is also a chance that learning can stall in this case, leading to classifiers, like those in the baseline network, that tend to underfit \cite{LiuZ-conf2023a}.  The second factor is the extraction and storage of multi-image contexts.  Doing so enables leveraging vital details that aid in classification and would otherwise be difficult to capture solely with convolution.  The table in \cref{fig:can-ablation-study}(e) quantifies the individual contribution of these two changes from the baseline network.  The remaining changes are included too.  However, their impact on network performance is less pronounced.

We now consider the {\sc CREN}.  We modify the network by appending branches after the transposed-convolution blocks, except the last block, in the decoder.  Each branch implements a one-step upsampling and thresholding so that a full-resolution segmentation map can be constructed.  We do not back-propagate the error from each of these branches onto the main pathway of the {\sc CREN} to avoid biasing the network's performance.  The network training protocols are the same as in the experiment section (see \hyperref[sec4]{section 4}).

The results in \cref{fig:cren-ablation-study}(i)--(ii) demonstrate that the baseline {\sc CREN} returns poor segmentation maps compared to the full network.  In early stages of the decoder, the baseline network is unable to resolve the flat, sandy seafloor class.  It also has difficulties in recognizing the rocky terrain that is present.  It instead labels those areas as belonging to an unknown class.  It is only when semantic features from the encoder are incorporated in the later stages that the ratio of known to unknown regions dwindles and the map quality improves slightly.  The full {\sc CREN}, in comparison, produces qualitatively and quantitatively better segmentation maps.  As with the baseline network, some improvement is made in later stages.  This behavior again occurs due to the integration of encoder-extracted features, which tend to emphasize low-anisotropic edges well, as shown by the contrast plots in \cref{fig:cren-ablation-study}(ii).

One of the reasons why the baseline network does poorly is that there is little to no adherence of the segmentation maps with observable class boundaries.  This behavior stems from the strong lack of supervision.  Alongside this issue is that there are no localized structured-prediction constraints to encourage neighboring regions to possess similar labels.  Both of these drawbacks are corrected with the inclusion of deep-superpixel regularization.  Our superpixel approach has the tendency to obey class boundaries, even for complex transitions.  It also identifies and iteratively merges perceptually homogeneous regions to ensure spatial label consistency.  Another reason why the baseline network underperforms is that there is no feature regularization.  The features used for making the initial segmentation masks are highly mixed.  Without ancillary supervision, it would be incredibly difficult to learn a robust feature-to-mask mapping.  The full network, in contrast, typically forms a compact, mostly-separable representation that is linearly separable.  Each of the feature clusters is strongly correlated with a particular class, which yields initial segmentation masks with few mistakes.  This assumes, however, that good seed cues are chosen.  Simply selecting high-rated regions in the class-activation maps and taking those as seeds will not always be appropriate.  Our uncertainty-quantification measure is therefore necessary.  The table in \cref{fig:cren-ablation-study}(e) quantifies the individual contribution of these three changes from the baseline network.  The remaining changes are included too.

Memory plays an important role in both networks.  The type of memory that is used for storing multi-image contexts also immensely influences performance.  In \cref{fig:can-ablation-study} and \cref{fig:cren-ablation-study}, we provide a secondary set of ablation studies to illustrate this claim.  We insert a range of memory modules, in lieu of our universal, recurrent event memories, into the {\sc CAN} and {\sc CREN}.  We adopt the best-known practices from the literature to aid in tuning these modules and their various parameters.  The network training protocols are the same as in the experiment section (see \hyperref[sec4]{section 4}).

Gated-type networks \cite{JozefowiczR-conf2015a}, like long short-term memories \cite{HochreiterS-coll1996a,HochreiterS-jour1997a}, are among the simplest architectures that we consider for these experiments.  Unlike classical recurrent networks, they posses explicit external memory and are thus able to link current content to the past with an arbitrary amount of time lag.  However, they possess one crucial flaw, which is that the stored information is forgotten whenever the external representation is updated.  One way to overcome this shortcoming is to consider banks of such single-entry memories, like we do in these experiments.  This can lead to incredibly long and difficult training rounds, though, which impacts performance.  Another possibility is to store the external information separately in multiple memory locations.

In this latter case, we consider a wide range of multi-entry memories.  These, unfortunately, also have practical issues.  Queue-based memories are largely ineffective when the memory banks are increased beyond a certain size.  Depending on the queue priority, image contexts mined either recently or far into the past that may be useful for inference will likely not be used.  Stack-based memories \cite{GrefenstetteE-coll2015a} have similar issues.  Content-addressable memories are typically better suited for our purposes.  Not all of them are effective, though.  Modules like neural Turing machines \cite{GravesA-jour2014a} can be difficult to train without diverging \cite{WestonJ-conf2015a,GuicehreC-jour2018a}.  Even when they can be reliably trained, they may overwrite useful data in memory.  Memory decay, due to the use of non-ideal reading operations, is also possible over time.  This limits the achievable temporal resolution in practical settings.  Differential neural computers \cite{GravesA-jour2016a} correct some of the problems with neural Turing machines.  However, they, and also and dynamic memory networks \cite{KumarA-conf2016a}, utilize reading and writing operations that can still yield poor time resolution. 

The content-addressable memories that we use do not possess these shortcomings.  They use a key-value querying mechanism that mimics transformer networks \cite{VaswaniA-coll2017a}.  This yields a decomposition of the memory event is free from the memory-depth and resolution trade-off.  We are thus free to use banks of arbitrary size with few concerns.  Another advantage of our chosen memory module is that they avoid the precision efficiency bottleneck created by the need for continuous read and write operations.  As such, they facilitate incredibly quick and stable training regimes.  This property also contributes to the graceful improvement in performance that we observe in \cref{fig:can-ablation-study} and \cref{fig:cren-ablation-study} when increasing the storage size.  We find that there is a nearly linear relationship between the context storage amount and the performance improvement for both classification and segmentation.  Such behavior contrasts with existing multi-context segmentation schemes where segmentation quality can quickly saturate for small memory sizes and decrease when the memory size grows.  Without the use of efficient, content-addressable memories, these alternate models may scale poorly.

\setstretch{0.95}\fontsize{9.75}{10}\selectfont
\putbib
\end{bibunit}

\end{document}